\documentclass[a4paper,fleqn]{cas-sc}

\usepackage[square,numbers,sort,compress]{natbib}
\usepackage{graphicx}
\usepackage[normalem]{ulem}
\useunder{\uline}{\ul}{}
\usepackage{breakurl}

\usepackage{booktabs}
\usepackage{eurosym}
\usepackage{breakurl}
\usepackage{todonotes}
\usepackage[normalem]{ulem}
\usepackage{multirow}
\usepackage[nolist]{acronym} 
\usepackage{tabulary}
\usepackage{hyperref}
\usepackage{color}
\usepackage{amssymb}
\usepackage{rotating}
\usepackage[T1]{fontenc}
\usepackage{lscape}
\usepackage[autostyle]{csquotes}
\usepackage[inline]{enumitem}
\usepackage[ruled,vlined]{algorithm2e}

\usepackage[utf8]{inputenc}
\usepackage{longtable}
\usepackage{pdflscape}
\usepackage{graphicx}
\usepackage{tabularx}
\usepackage{multirow}
\usepackage{textcomp}
\usepackage{csquotes}
\usepackage{booktabs}
\usepackage{graphics}
\usepackage{xr-hyper}
\usepackage{hyperref}
\usepackage{amsmath}
\usepackage{amssymb}
\usepackage{natbib}
\usepackage{epsfig}
\usepackage{lscape}
\usepackage{lineno}
\usepackage{array}
\usepackage{float}
\usepackage{todonotes}
\usepackage{tabu}
\usepackage{siunitx}

\usepackage{pdflscape}
\usepackage{setspace}
\usepackage{textcomp}
\definecolor{bblue}{HTML}{4285F4}
\definecolor{rred}{HTML}{DB4437}
\definecolor{ggreen}{HTML}{0F9D58}
\definecolor{yyellow}{HTML}{F4B400}
\usepackage{pgfplots}
\usepgfplotslibrary{units}
\pgfplotsset{compat=1.14}

\def\tsc#1{\csdef{#1}{\textsc{\lowercase{#1}}\xspace}}
\tsc{WGM}
\tsc{QE}
\tsc{EP}
\tsc{PMS}
\tsc{BEC}
\tsc{DE}

\begin{document}
\let\WriteBookmarks\relax
\def\floatpagepagefraction{1}
\def\textpagefraction{.001}

\shorttitle{Advancing 3D Point Cloud Understanding through Deep Transfer Learning}

\shortauthors{S. S. Sohail et~al.}

\title [mode = title]{Advancing 3D Point Cloud Understanding through Deep Transfer Learning: A Comprehensive Survey}                      

\author[1]{Shahab~Saquib~Sohail}
%
\author[2]{Yassine Himeur}\cormark[1]
\ead{yhimeur@ud.ac.ae}

\author[3]{Hamza Kheddar}
%
\author[4,5]{Abbes Amira}[]
%
\author[6]{Fodil Fadli}
%
\author[2]{Shadi Atalla}
%
\author[2]{Abigail Copiaco}[]
%
\author[2]{Wathiq Mansoor}

\address[1]{School of Computing Science and Engineering, VIT Bhopal University, Sehore, MP 466114, India}
\address[2]{College of Engineering and Information Technology, University of Dubai, Dubai, UAE}
\address[3]{LSEA Laboratory, Department of Electrical Engineering, University of Medea, 26000, Algeria}
\address[4]{Department of Computer Science, University of Sharjah, UAE}
\address[5]{Institute of Artificial Intelligence, De Montfort University, Leicester, United Kingdom}
\address[6]{Department of Architecture \& Urban Planning, Qatar University, Doha, 2713, Qatar}

\begin{abstract}
The 3D point cloud (3DPC) has significantly evolved and benefited from the advance of deep learning (DL). However, the latter faces various issues, including the lack of data or annotated data, the existence of a significant gap between training data and test data, and the requirement for high computational resources. To that end, deep transfer learning (DTL), which decreases dependency and costs by utilizing knowledge gained from a source data/task in training a target data/task, has been widely investigated. Numerous DTL frameworks have been suggested for aligning point clouds obtained from several scans of the same scene. Additionally, DA, which is a subset of DTL, has been modified to enhance the point cloud data's quality by dealing with noise and missing points. Ultimately, fine-tuning and DA approaches have demonstrated their effectiveness in addressing the distinct difficulties inherent in point cloud data. This paper presents the first review shedding light on this aspect. it provides a comprehensive overview of the latest techniques for understanding 3DPC  using DTL and domain adaptation (DA). Accordingly, DTL's background is first presented along with the datasets and evaluation metrics. A well-defined taxonomy is introduced, and detailed comparisons are presented, considering different aspects such as different knowledge transfer strategies, and performance. The paper covers various applications, such as 3DPC object detection, semantic labeling, segmentation, classification, registration, downsampling/upsampling, and denoising.  Furthermore, the article discusses the advantages and limitations of the presented frameworks, identifies open challenges, and suggests potential research directions.
\end{abstract}




\maketitle


\begin{acronym}
\acro{3DPC}{3D point cloud}
\acro{GAN}{generative adversarial network}
\acro{CV}{computer vision}
\acro{DL}{deep learning}
\acro{ML}{machine learning}
\acro{DTL}{deep transfer learning}
\acro{DA}{domain adaptation}
\acro{UDA}{unsupervised domain adaptation}
\acro{TD}{target domain}
\acro{SD}{source omain}
\acro{CNN}{convolutional neural network}
\acro{MAE}{multi-scale masked autoencoder}
\acro{IoU}{intersection over union}
\acro{SSL}{self-supervised learning}
\acro{SVCN}{sparse voxel completion network}
\acro{OA}{overall accuracy}
\acro{CD}{Chamfer distance}
\acro{LiDAR}{light detection and ranging}
\acro{BiLSTM}{bidirectional LSTM}
\acro{ITL}{inductive transfer learning}
\acro{SFUDA}{source-free unsupervised DA}
\acro{PCM}{point cloud mixup}
\acro{MLP}{multi-layer perceptron}
\acro{CLIP}{contrastive vision-language pretraining}
\acro{TTL}{transductive transfer learning}
\acro{CLDA}{cross-modal learning DA}
\acro{AE}{auto-encoder}
\acro{RNN}{recurrent neural network }
\acro{CRF}{conditional random field}
\end{acronym}

\section{Introduction} \label{sec1}
\subsection{Preliminary}
\Ac{CV} continues to attract significant interest as a growing branch of \ac{ML}, which targets different problems in smart cities, medicine, autonomous driving, medicine, video surveillance, scene understanding, safety, and security \cite{himeur2023video,copiaco2023innovative}. With the rapid development of sensor technology, 3D sensors have recently been widely adopted, which increased the interest of the \ac{CV} research community in developing 3D sensor data processing methodologies \cite{zhou2023attention,habchi2023ai}. Additionally, with the use of augmented reality and virtual reality (AR/VR), 3D vision problems become more important since they provide much richer information than 2D \cite{ji2023semi,kerdjidj2024exploiting}.
Typically, numerous 3D sensors for acquiring 3D data have been used, including depth-sensing cameras (such as Apple Depth, RealSense, and Kinect cameras), \ac{LiDAR}, which is used for mobile mapping terrestrial laser scanning, aerial \ac{LiDAR}  \cite{fotsing2022volumetric}. 
Moreover, \acp{GAN} can be used to augment data when data scarcity problem occurs. In this context, 3D data can provide rich scale, shape, and geometric information to be complemented with 2D images for better representing the surrounding environment \cite{lai2022stratified,bechar2024federated}.

While there are different approaches to representing 3D data, such as volumetric grids, meshes, and depth images, the \acp{3DPC} are the most used. Typically, a \ac{3DPC} representation conserves the original geometric information in 3D space \cite{abbasi2022lidar}. Besides, a \ac{3DPC} is an ensemble of data points representing an object's surfaces in 3D coordinates. In this regard, spatial coordinates are used to represent data points, surface normals, and color information and format ( e.g., RGB, HSV, and others) \cite{liu2022imperceptible}. 

Additionally, although \ac{3DPC} can be considered as non-Euclidean geometric data,  in practice, delineated as small Euclidean subgroups with a standard coordinates system and global parametrization \cite{yu2022part}. 
The success of this representation is due to its invariability to transformations and attacks, including rotation, scaling, translation, etc., which makes it robust to extracting objects' features. Moreover, using \acp{DL} techniques to extract \ac{3DPC} data and perform complex tasks such as detection, classification, recognition, and retrieval has expanded their adoption in many research and development areas \cite{li2022simipu}. Typically, \acp{3DPC} are widely adopted to \ac{CV} tasks, such as object recognition, semantic segmentation, and scene understanding \cite{elharrouss2021panoptic,cao2023semantic}. Additionally, they are used to create detailed models of environments for navigation and localization, detailed maps and models of buildings, landscapes, and other structures, as well as detailed models of buildings and other structures for design and planning \cite{liu20233d}. Moreover, they can be utilized to inspect and maintain industrial equipment, infrastructure, and other assets \cite{yang2022automated,liu2022whale}. Besides, the \ac{3DPC} technology helps implement realistic and immersive experiences in AR and VR applications.

The traditional \ac{ML} approaches and, recently, \ac{DL} methods witnessed rapid growth and have attracted researchers because of their applicability in many real-life applications, including smart healthcare \cite{ahmed2021deep}, disease diagnosis and medical image classifications \cite{habchi2023ai}, business and marketing \cite{mehdiyev2020novel}, recommender systems \cite{kiran2020dnnrec}, energy \cite{himeur2021artificial}, agriculture \cite{kashyap2021towards},  robotics \cite{de2019semiautomatic},  and more. The primary advantage of these learning algorithms is that they train a model that learns the hidden pattern and can be exploited for any specific purpose with high accuracy. However, with time, some issues have been raised by the research communities. First, these learning algorithms require extensive training datasets, especially \ac{DL} algorithms \cite{himeur2023video}.
Second, most \ac{DL} models are based on supervised learning and then require huge amounts of ground-truth data. Third, it is a common assumption that the data that is trained and the future data to be processed must have the same distribution and be in the same feature space \cite{sayed2022deep}. 
It becomes difficult and sometimes impossible to maintain  the aforementioned assumption in several real-life scenarios. For example, when performing specific tasks, such as  classification in a particular domain; however, we may have the requisite trained data in another domain, where both datasets may not have the same distribution or same feature space.

\begin{figure}[t!]
\begin{center}
\includegraphics[width=1\textwidth]{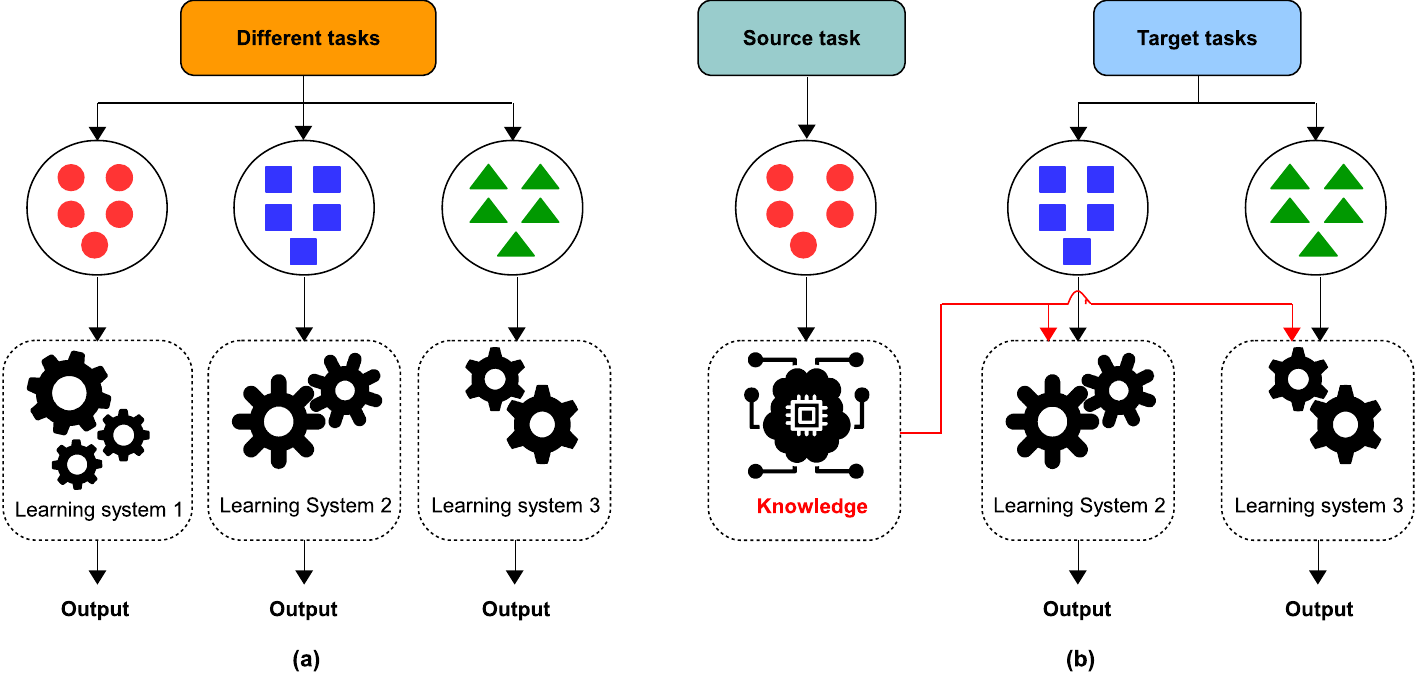}\\
\end{center}
\caption{Difference between conventional ML and DTL techniques for multiple tasks: (a) conventional ML and (b) DTL.}
\label{domain_adapt}
\end{figure}

Furthermore, an essential factor for ensuring accurate performance of \ac{ML} algorithms is consistency in the distribution and feature space of the datasets used for training and testing. If the distribution of data changes, it becomes necessary to rebuild the model from scratch by collecting new training data \cite{KheddarASR2023}. However, this process is not only costly but also often impractical due to the challenges associated with re-collecting training data. Therefore, there is a need for a mechanism that can reduce the expenses associated with re-collecting training data, minimize the cost of data labeling, and still achieve high performance without requiring extensive amounts of training data.
To that end, knowledge transfer has been suggested, which would fit the above constraints and can significantly improve performance. This knowledge transfer mechanism is termed \textit{\ac{DTL}} \cite{himeur2023video,KheddarASR2023,kheddar2023deep}. \textcolor{black}{Fig. \ref{domain_adapt} explains briefly the difference between conventional \ac{ML} and \ac{DTL} techniques.}

\begin{figure*}[t!]
\begin{center}
\includegraphics[width=1\textwidth]{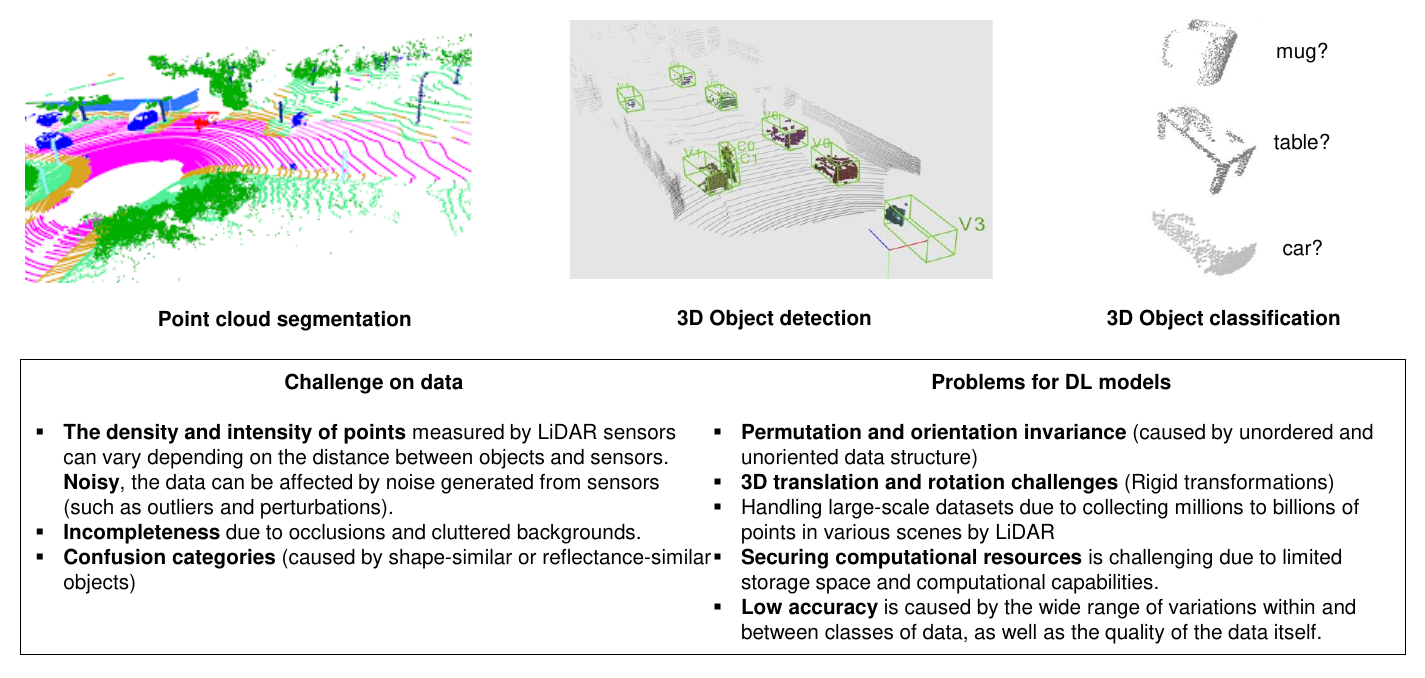}\\
\end{center}
\caption{Tasks and challenges related to data and \ac{DL}/\ac{DTL}-based applications on \acp{3DPC}.
}
\label{PC-tasks}
\end{figure*}

On the one hand, employing \ac{DL} for \ac{3DPC} poses a rather complex challenges because of: (i) the variability in point density and reflective intensity, influenced by the distance between objects and \ac{LiDAR} sensors, (ii) existence of noise originated from sensors ( e.g., perturbations and outliers), (iii) data incompleteness caused by occlusion between objects and cluttered
background, and (iv) confusion categories caused by shape-similar or reflectance-similar objects. \textcolor{black}{On the other hand, challenges of \ac{DL}/\ac{DTL} models include (i) permutation and orientation invariance, (ii) 3D translation and rotation challenges, (iii) difficulty of handling large-scale datasets, (iv) securing computational resources, and (v) low performance}. Fig. \ref{PC-tasks} summarizes the tasks and challenges related to data and \ac{DL}/\ac{DTL}-based applications on \acp{3DPC}.

\subsection{Our contributions}
Arguably, one of the top success stories of \ac{DL} is \ac{DTL}. Accordingly, the finding that pretraining a \ac{DL} network on a rich source set (e.g., ImageNet) can help boost performance once fine-tuned on a usually much smaller target set has been instrumental to many applications in language and \ac{CV}. 
Likewise, several studies have explored the potential of \ac{DTL} in \ac{3DPC} applications to address the aforementioned challenges. This offers an opportunity to present the first review article that comprehensively examines the contributions of \ac{DTL}-based approaches in advancing our understanding of 3D scenes, such as \ac{3DPC} segmentation, 3D object detection, 3D object classification, \ac{3DPC} registration, and more.
For instance, \ac{DTL} has been shown to be effective in various ways, including (i) leveraging knowledge learned from synthetic data to improve semantic segmentation in real \ac{LiDAR} \ac{3DPC}, (ii) enabling accurate classification of \acp{3DPC} even with limited training data, (iii) mitigating the issue of overfitting in \ac{3DPC} classification, and (iv) reducing the labor-intensive process of annotating \ac{3DPC} datasets, among other benefits.
In this respect, this paper first presents a background of \ac{DTL}, where a well-defined taxonomy is conducted. Moving on, datasets and evaluation metrics used to evaluate existing \ac{DTL}-based \acp{3DPC} techniques are discussed. Next, existing studies are overviewed based on different aspects, and their pros and cons are identified. 
After that, the challenges encountered when using \ac{DTL} for \acp{3DPC} are identified before deriving future research directions. To summarize, the main contributions of this review can be stated as follows: 
\begin{itemize}
\item Presenting, to the best of the authors' knowledge, the first review article on using \ac{DTL} and \ac{DA} for \ac{3DPC} applications; 
\item Discussing existing datasets and evaluation metrics used for assessing the performance of \ac{3DPC} frameworks based in \ac{DTL}; 
\item Introducing a well-defined taxonomy to overview existing \ac{DTL}-based \ac{3DPC} studies; 
\item Identifying the current challenges encountered when using \ac{DTL} for \ac{3DPC} understanding tasks; and 
\item Deriving future research directions that can attract significant research interest in the near future. 
\end{itemize}

\subsection{Methodology of the survey}
The present review study is based on the protocols and procedure suggested in \cite{kitchenham2004procedures}, which has recently been adopted widely in many studies such as \cite{himeur2021survey}. This study is motivated by the recent developments in \ac{3DPC} technology. There are some reviews available on the theme which has exploited \ac{ML} \cite{bello2020deep}, \ac{DL} \cite{guo2020deep}, and  reinforcement learning \cite{li2017deep}; however, to the best of our search and efforts, we could not find a review exclusively exploring \ac{DTL} for different \ac{3DPC} tasks. The proliferation of \ac{DTL} techniques has influenced researchers by virtue of which the span of related techniques has been expanding and has covered much recent AI-driven research. Our study gives the readers an insight into how \ac{DTL} is being implemented for performing many \ac{3DPC} tasks. This review aims at finding the answer to research questions presented in Table \ref{tab:RQs}.

\begin{table*}[t!]
\caption{Research questions covered in this review.}
\label{tab:RQs}
\scriptsize
\color{black}
\begin{tabular}{
m{8mm}
m{55mm}
m{90mm}
}
\hline

No. & Question	& Objective   \\

\hline
RQ1 & What are the key principles and theoretical foundations of  \ac{DTL} and \ac{3DPC}?	& Provide a foundational understanding of the key principles and theories behind \ac{DTL} and \ac{3DPC}. This will aid readers in grasifying the basic tenets upon which the field of study is built and support comprehension of more complex concepts discussed later in the review.  \\ \hline

RQ2 & Why the application of \ac{DTL} in \ac{3DPC} is receiving increasing attention?	& Examine and explain why \ac{DTL} is increasingly being used for \ac{3DPC}. This will help highlight the unique benefits and opportunities offered by \ac{DTL} in handling 3D data, indicating its growing importance in the field.   \\ \hline

RQ3 & How the different \ac{3DPC} tasks can be implemented using \ac{DTL}? 	& Present a comprehensive overview of how \ac{DTL} can be utilized in various \ac{3DPC} tasks. This objective aims to showcase the versatility and wide-ranging applicability of \ac{DTL} in \ac{3DPC} tasks, providing practical insights for researchers and practitioners.   \\ \hline

RQ4 & What are the \ac{DTL} mechanisms and settings for performing the \ac{3DPC} tasks and optimizing their performance? 	& Delve into the specific mechanisms and settings of \ac{DTL} that optimize performance in \ac{3DPC} tasks. The objective here is to provide a clear understanding of the operational details of \ac{DTL}, which can serve as a guide for those intending to implement \ac{DTL} in their own \ac{3DPC} tasks.   \\ \hline

RQ5 & Which \ac{DTL} models are better appropriate for \ac{3DPC} tasks?	& Evaluate and suggest \ac{DTL} models that are most suitable for \ac{3DPC} tasks. By doing this, the review will provide practical advice for readers looking to apply \ac{DTL} in \ac{3DPC}, assisting them in model selection.   \\ \hline

RQ6 & What are the issues in implementing \ac{DTL} for performing the above tasks, and how to tackle these challenges?	& Identify the key challenges associated with implementing \ac{DTL} for \ac{3DPC} tasks and to propose potential solutions to these challenges. This will help improve the application of \ac{DTL} in \ac{3DPC} by addressing problems head-on, promoting a robust and reliable usage of these techniques.   \\ \hline

RQ7 & What are the future research directions for improving \ac{DTL} in \ac{3DPC}? & Forecast future research directions for \ac{DTL} in \ac{3DPC}. This objective aims to stimulate new research initiatives by identifying promising avenues for further exploration, driving innovation and development in the field.  \\ \hline

\hline

\end{tabular}
\end{table*}

\begin{figure}
\centering
\includegraphics[scale=0.7]{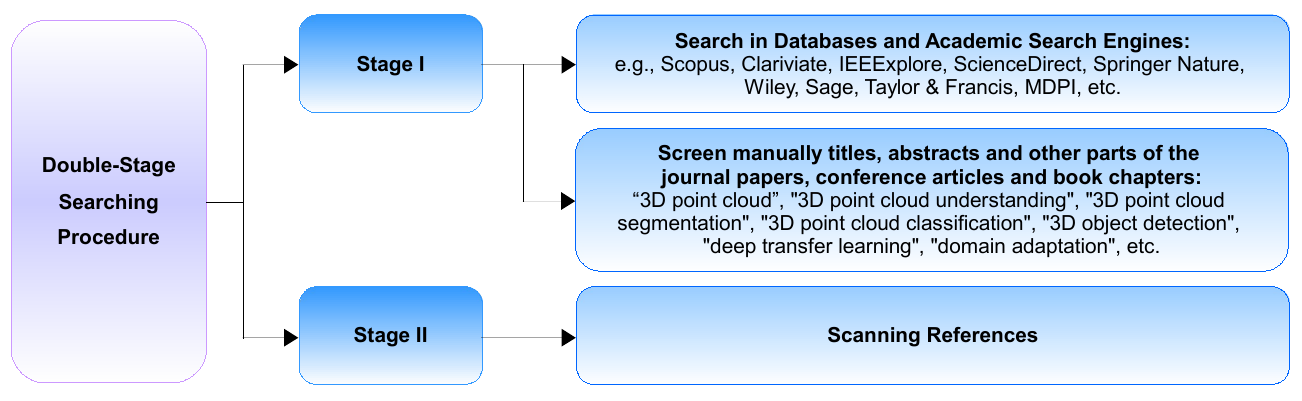}\\
(a)\\
\includegraphics[width=1\textwidth]{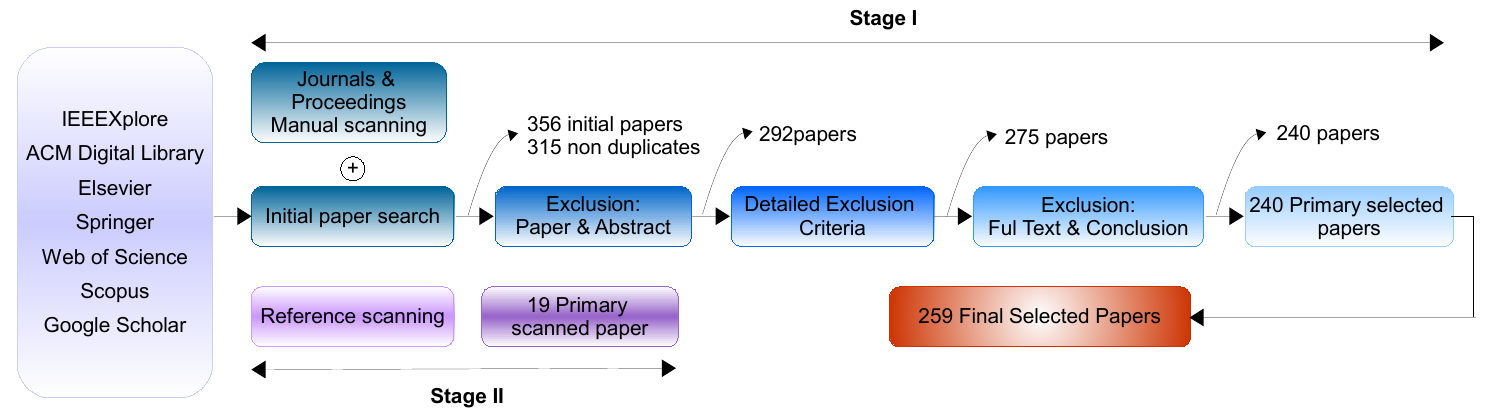} \\
(b)\\
\caption{\textcolor{black}{Summary of the approach used to search and select articles included in the review: (a) The adopted search procedure and (b) The selection criteria. }}
\label{fig:SearchProcess}
\end{figure}


The bibliometric research is performed in the context of a narrative review. The recent works related to the \ac{3DPC} involving \ac{DTL} techniques for performing their respective tasks have been searched. We searched the relevant keywords, like, "3D point cloud", "3D point cloud segmentation", "3D point cloud classification", "3D object detection", "deep transfer learning", and "domain adaptation", with different combinations on Scopus database in titles, abstracts, and keywords. The adopted search procedure is explained in Figs. \ref{fig:SearchProcess}(a) and   \ref{fig:SearchProcess}(b).
\textcolor{black}{Typically, in response to the relevant queries, 176 articles were retrieved, with 149 of them being non-duplicates. These articles underwent further scrutiny to determine their relevance to the theme of the present study. Ultimately, 108 papers were selected for inclusion in this study.}

\begin{figure}
    \centering
    \includegraphics[scale=0.55]{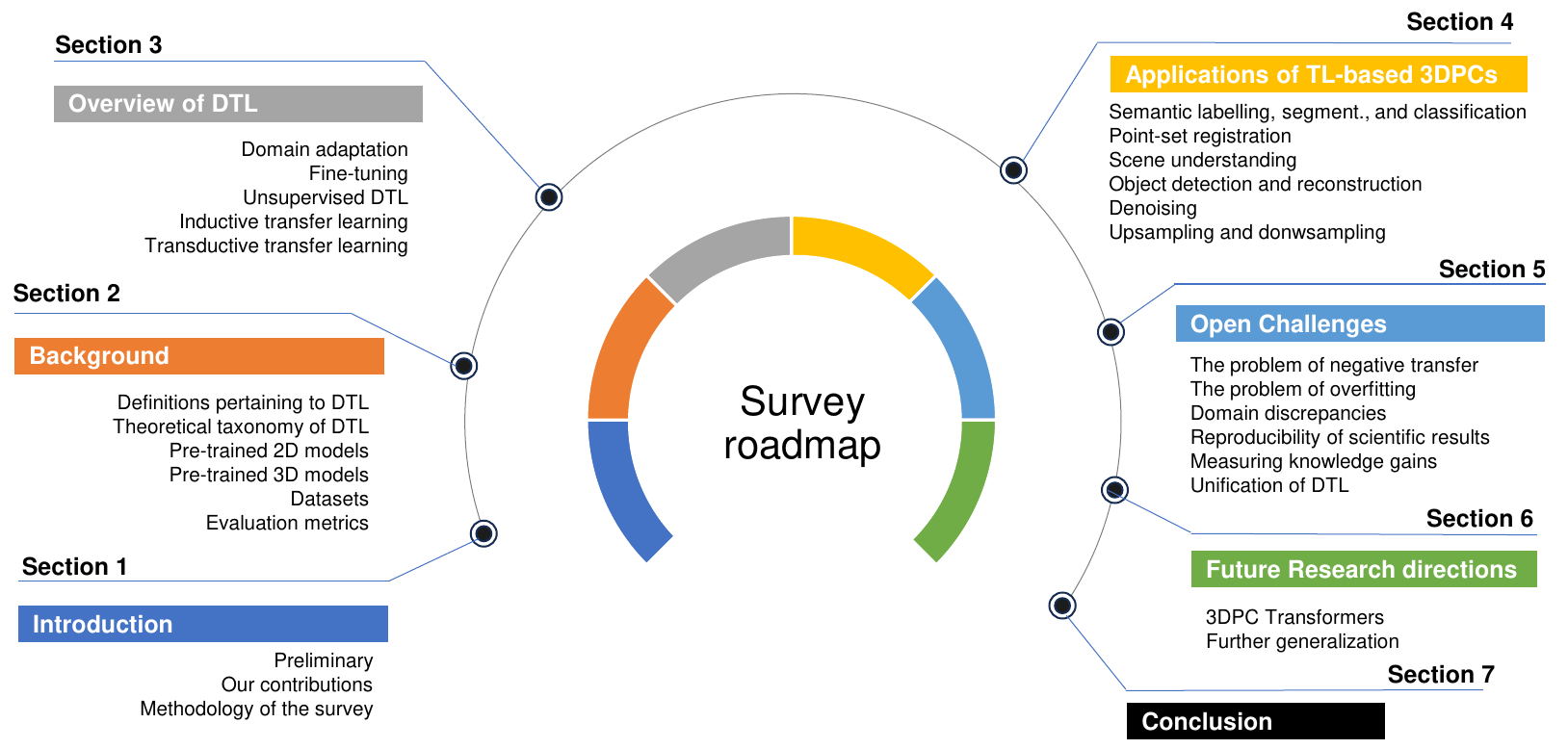}
    \caption{Survey structure with sections and sub-sections disctribution. }
    \label{fig:road}
\end{figure}

The paper is organized as follows: Section \ref{sec2} introduces background related to \ac{DTL} definitions and terms, pre-trained models, useful datasets, and metrics used in \ac{3DPC}. Section \ref{sec3} reviews relevant literature and work related to \ac{DTL} and \ac{DA}. Section \ref{sec4} outlines state-of-the-art proposed applications based on \ac{DTL}. Section \ref{sec5} discusses open challenges. Section \ref{sec6} suggests recent future research directions. Finally, Section \ref{sec7} concludes the survey. Fig. \ref{fig:road} illustrates the survey's structure, enhancing readability, and offering guidance to readers.

\section{Background} \label{sec2}
\subsection{3D Point Cloud}
A 3DPC is a collection of data points in a three-dimensional coordinate system. Each point in the point cloud is represented as:
\[ P = \{ p_1, p_2, ..., p_n \} \]
where each point \( p_i \) is a vector in \(\mathbb{R}^3\):
\[ p_i = (x_i, y_i, z_i) \]

\subsubsection{Operations and Transformations}

Common operations and transformations applied to 3DPCs include:

\begin{itemize}
  \item \textbf{Translation:}
    Moving the point cloud by adding a constant vector \( \mathbf{t} = (t_x, t_y, t_z) \) to each point:
    \[ p_i' = p_i + \mathbf{t} \]

  \item \textbf{Rotation:}
    Rotating the point cloud using a rotation matrix \( \mathbf{R} \). The rotation matrix is a 3x3 matrix:
    \[ p_i' = \mathbf{R} p_i \]
    Rotation matrices can be derived from Euler angles, axis-angle, or quaternion representations.

  \item \textbf{Scaling:}
    Changing the size by scaling each point by a scalar \( s \) or different scalars for each axis:
    \[ p_i' = s \cdot p_i \text{ or } p_i' = (s_x x_i, s_y y_i, s_z z_i) \]

  \item \textbf{Transformation:}
    A combination of translation, rotation, and scaling, represented as matrix multiplication in homogeneous coordinates:
    \[ p_i' = \mathbf{T} \begin{bmatrix} x_i \\ y_i \\ z_i \\ 1 \end{bmatrix} \]
    where \( \mathbf{T} \) is a 4x4 transformation matrix:
    \[
    \mathbf{T} = \begin{bmatrix}
    \mathbf{R} & \mathbf{t} \\
    0 & 1
    \end{bmatrix}
    \]
\end{itemize}

\subsection{Definitions pertaining to DTL}
\noindent This section presents the main definitions used in \ac{DTL}-based \ac{3DPC} applications.

\noindent \textbf{\textit{Def. 1 - Domain:}} Consider a dataset $X$ consisting of $n$ observations, $x_1, \cdots, x_n$, in a feature space $\chi$. The marginal probability distribution of $X$ is represented by $P(X)$. A domain, denoted as $\mathbb{D}$, is defined as the set containing $X$ and $P(X)$. In the field of \ac{DTL}, the domain containing the initial knowledge is referred to as the \ac{SD}, represented by $\mathbb{D}_S$, while the domain containing the unknown knowledge to be learned is called the \ac{TD} and represented by $\mathbb{D}_T$ .

\vskip2mm
\noindent \textbf{\textit{Def. 2 - Task:}} The dataset $X$ contains $n$ observations, $x_1, \cdots, x_n$, in a feature space $\chi$, and it is associated with a set of labels $Y$, $y_1, \cdots, y_n$, in a label space $\gamma$. A task can be defined as a set containing the labels $Y$ and a learning objective predictive function $\mathbb{F}(X)$, represented by $\mathbb{T}=\{Y, \mathbb{F}(X)\}$. This function is also denoted as the conditional distribution $P(Y|X)$. In accordance with this definition of task, the label spaces of the \ac{SD} and \ac{TD} are represented as $\gamma_S$ and $\gamma_T$ respectively \cite{ramirez2019learning}.

One way to classify \ac{DTL} methods is based on their approach to transferring knowledge, which can be broken down into \textit{what}, \textit{when}, and \textit{how} knowledge is transferred.

\noindent\textbf{(a) What knowledge is transferred:} The classification of \ac{DTL} methods based on "what knowledge is transferred" examines the specific characteristics of knowledge that can be transferred across domains or tasks. Some information is unique to a particular domain or task, while other knowledge is general and can improve the performance of the \ac{TD} or task. Based on this criterion, \ac{DTL} methods can be categorized as model-based, relation-based, instance-based, and feature-based \cite{morid2021scoping}.

\noindent\textbf{(b) How knowledge is transferred:} The classification of \ac{DTL} methods based on this question focuses on the specific algorithms or techniques used to transfer knowledge across domains or tasks.

\noindent\textbf{(c) When knowledge is transferred:} inquires as to when and under what circumstances knowledge should or should not be transferred.

\subsection{Theoretical taxonomy of DTL}
\begin{figure*}[t!]
\begin{center}
\includegraphics[width=1\textwidth]{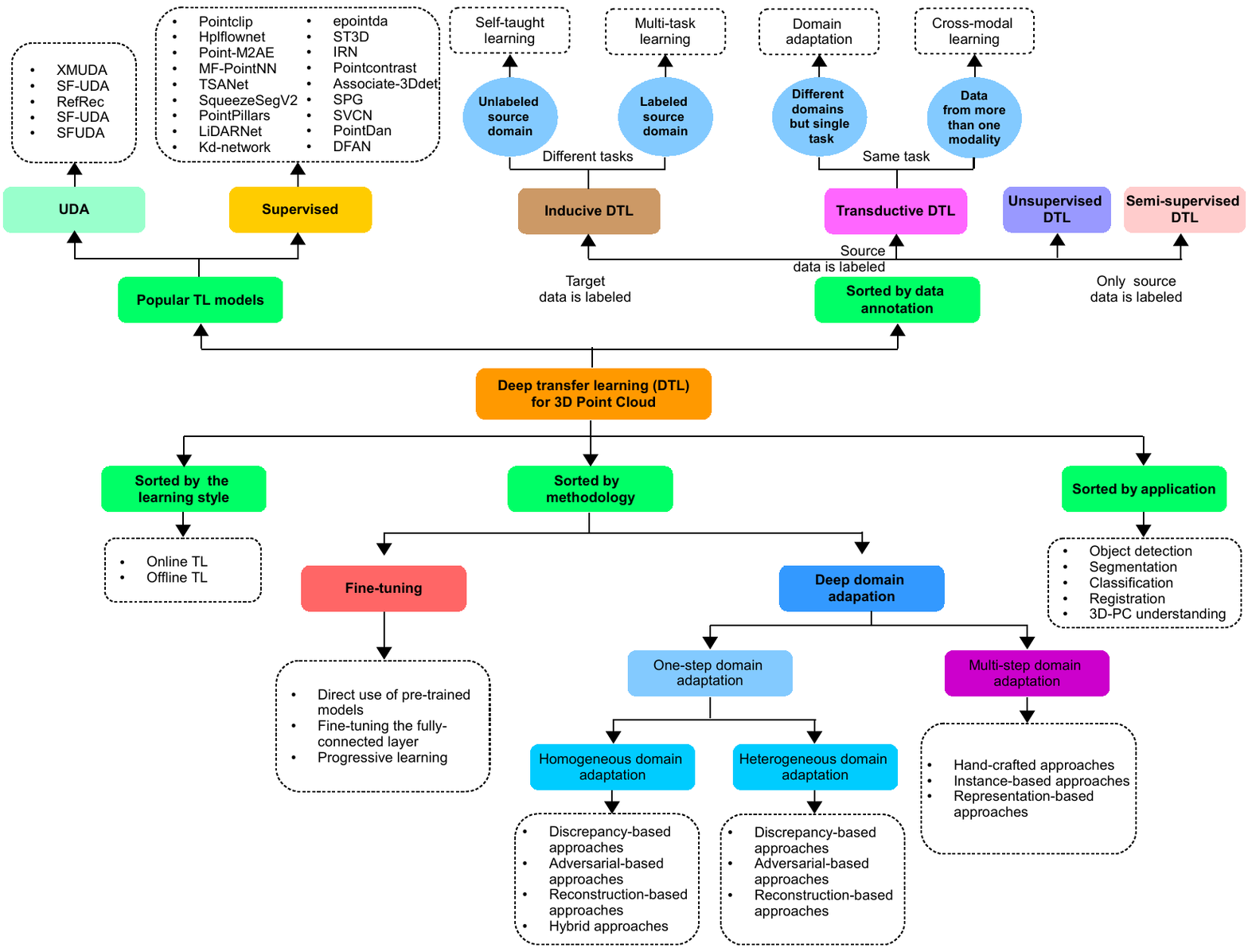}\\
\end{center}
\caption{\textcolor{black}{Proposed taxonomy of existing \ac{DTL} algorithms for \acp{3DPC}. }}
\label{TaxTL}
\end{figure*}

\begin{figure*}[t!]
\begin{center}
\includegraphics[width=1\textwidth]{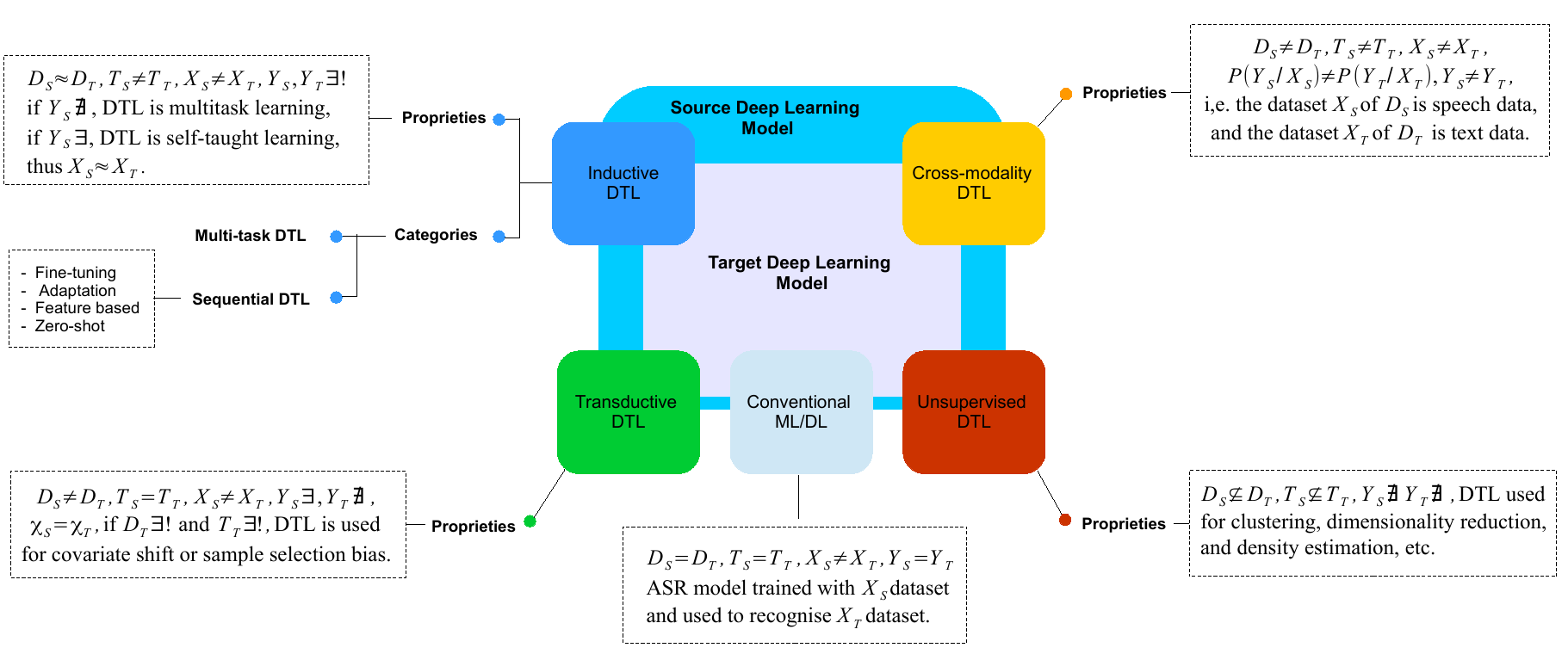}\\
\end{center}
\caption{The \ac{DTL} classification is determined by how similar the source and \acp{TD} and tasks are. The symbol ($\varsubsetneq$) is used to indicate that the domains/tasks are distinct yet related. The symbol ($\exists!$) indicates the presence of a single unique domain/task. While ($\cong$) denotes that the domains, tasks, or spaces do not match.}
\label{tab:TL-types}
\end{figure*}

In this section, a clear categorization of \ac{DTL} methodologies used for \acp{3DPC} is presented. The proposed classification, shown in Fig. \ref{TaxTL}, is arranged based on the following criteria: (i) learning style, (ii) methodology, (iii) DTL  type, (iv) data annotation, and (v) widely used \ac{DTL} models. \ac{DTL} techniques can generally be divided into different groups depending on whether the source and \acp{TD} and tasks are alike or not. \textcolor{black}{The mathematical descriptions of the different \ac{DTL} groups and their distinctions are presented in Fig. \ref{tab:TL-types}. While this information is discussed in the literature review, aiming to simplify it and present it in a more reader-friendly manner in this study. }

\subsubsection{Inductive DTL}
The goal of inductive \ac{DTL} is to improve the target prediction function $\mathbb{F}T$ in the \acp{TD} when compared to classical \ac{ML}. This is achieved even when the target tasks $\mathbb{T}{T}$ are different from the source tasks $\mathbb{T}{S}$. However, the \ac{SD}  $\mathbb{D}{S}$ and \ac{TD} $\mathbb{D}_{T}$ may not always be identical (as shown in Fig. \ref{tab:TL-types}). Inductive \ac{DTL} can take two forms, depending on the availability of labeled or unlabeled data:

\noindent \textbf{(a) Multi-task DTL:} This approach is used when the \acp{SD} has a large labeled dataset ($X_{S}$ labeled with $Y_{S}$). This is a specific form of multi-task learning where multiple tasks $(T_1, T_2,\dots, T_n)$ are learned simultaneously (in parallel), including both the source and target tasks\cite{li2021can}.

\begin{figure*}[t!]
\begin{center}
\includegraphics[width=1\textwidth]{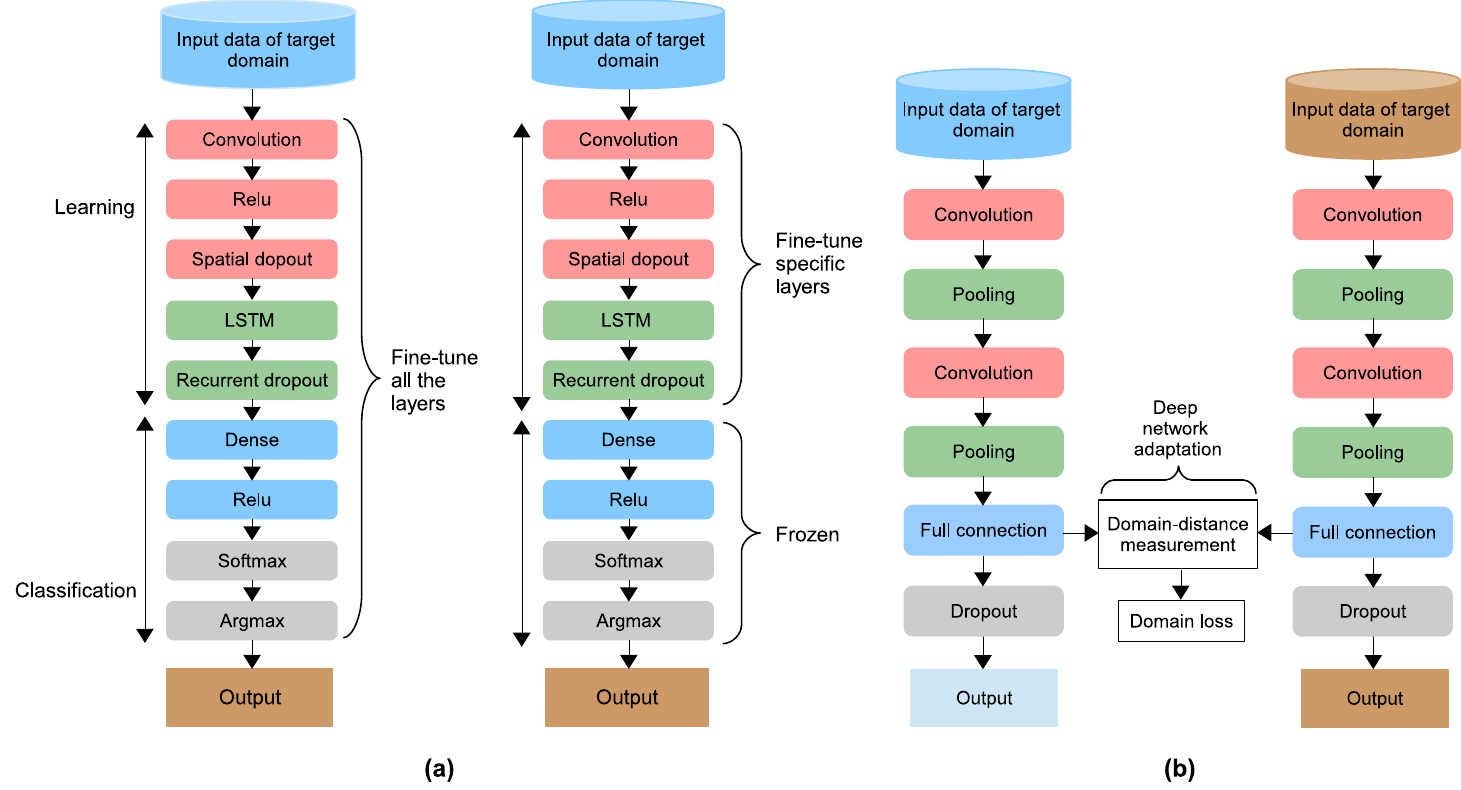}\\
\end{center}
\caption{Example of DTL models used in \acp{3DPC}: (a) fine-tuning, and (b) deep DA.} 
\label{finetuning_vs_DA}
\end{figure*}

\noindent \textbf{(b) Sequential learning:} also known as self-taught learning, is a method used when the dataset in the \ac{SD}  is unlabeled. It relies on (i) transferring the feature representation learned from a large collection of unlabeled datasets, and (ii) applying the learned representation to labeled data for classification tasks. This \ac{DTL} method involves sequentially learning multiple tasks where the gaps between the \acp{SD} and \acp{TD} may differ.
For instance, assume we have a pre-trained model (PTM) $M$ and we apply \ac{DTL} to several tasks $(T_1, T_2,\dots, T_n)$. In this approach, a specific task $\mathbb{T}_{T}$ is learned at each time step $t$ and it is slower than multi-task learning, however, when not all the tasks are present at the time of training, it may be advantageous. Sequential learning can be further classified into different types \cite{alyafeai2020survey}.
\begin{itemize}
\item [1-] \textbf{Fine-tuning:} involves training a new function, $\mathbb{F}_T$, that adapts the parameters of a PTM $M$ from a source task, $\mathbb{T}_S$, to a target task, $\mathbb{T}_T$, by translating the weights of the source task, $W_S$, to the weights of the target task, $W_T$. This can be done across all layers or just a subset of them, and the learning rate for each layer can be adjusted independently (known as discriminative fine-tuning). Additionally, new parameters, $K$, can be added to the model to improve its performance on the target task \cite{ribani2019survey}.
\begin{equation}
\label{adapt}
\mathbb{F}_T(W_T, K) = W_S \times K    
\end{equation}
    \item [2-] \textbf{Adapter modules:} they are designed to take a PTM, $M_S$, and adapt its weights, $W_S$, to a target task, $\mathbb{T}_{T}$. The adapter module achieves this by introducing a new set of parameters, $K$, that are smaller in size compared to $W_S$, i.e. $K\ll W_S$. Both $K$ and $W_S$ are decomposed into smaller, more compact modules, such that $W_S={w}_n$ and $K={k}_n$. This allows the adapter module to learn a new function, $\mathbb{F}_T$, which adapts the model to the target task.
\begin{equation}
\label{adapt}
 \mathbb{F}_T(K, W_S)= k_{1}'\times w_{1}\times \cdots  k_n'\times w_n    
\end{equation}
The equation (\ref{adapt}) illustrates the procedure of adapting a model to a new task by dynamic weight adjustment, original weights $W_S={w}_n$ remain unchanged, whereas the set of weights $K$ are updated to $K'={k'}_n$. This principle of Dynamic DA is illustrated in Fig. \ref{finetuning_vs_DA}.

\item [3-] \textbf{Feature based:} it focuses on learning concepts and representations at various levels of an image, such as corners or interest points, blobs or regions of interest points, ridges, or edges $E$. In this approach, the collection of $E$ obtained from a PTM $M$ is kept unchanged and only $W'$ is fine-tuned, such that the function $\mathbb{F}_T$ can be represented as $E \times W'$. The idea is that $E$ are the feature learned from PTM and fine-tuning only the last layer $W'$ to adapt to the new task.

\item [4-] \textbf{Zero-shot:}  is the simplest approach among all the others. It does not involve modifying or adding new parameters to a PTM by assuming that the existing parameters, denoted as $W_S$, cannot be changed. Essentially, zero-shot does not require any training to optimize or learn new parameters."
\end{itemize}

\subsubsection{Transductive DTL}
\label{subsecTTL}
In comparison to traditional \ac{ML}, which can be used as a benchmark for DA and \ac{DTL}, \ac{DTL} addresses the scenario where the \ac{TD} data, denoted as $\mathbb{D}{T}$, differs from the \acp{SD} data, denoted as $\mathbb{D}{S}$. While the \acp{SD} has annotated data ($X_{S}$ paired with $Y_{S}$), the \ac{TD} has no labeled data. The source and target tasks are similar as outlined in Fig. \ref{tab:TL-types}. Transductive \ac{DTL} aims at constructing a target prediction function $\mathbb{F}_T$ using the knowledge from both the \ac{SD} and \ac{TD}. Additionally, transductive \ac{DTL} can be further categorized into two groups based on the relationship between the \ac{SD} and \ac{TD}, as described in \cite{wan2021review}.

\noindent \textbf{(a) Deep domain adaptation (DDA):} refers to the situation where the feature spaces of the \ac{SD}, denoted as $\chi_{S}$, and the \acp{TD}, denoted as $\chi_{T}$, are the same. However, the probability distributions of the input data are different, with $P(Y_{S}/ X_{S})\neq P(Y_{T}/ X_{T})$ as described in \cite{liu2021optimal}. DDA is particularly useful when the target task has a unique distribution or limited labeled data, as highlighted in \cite{tan2018survey}. 

\noindent \textbf{(b) Cross-modality DTL:} most \ac{DTL} techniques require some form of relationship between feature spaces (or label spaces) of the source and \acp{TD}, i.e., $\mathbb{D}{S}$ and $\mathbb{D}{T}$. This means that \ac{DTL} can only be applied when the source and target have the same modality, such as text, speech, or video. In contrast, cross-modality \ac{DTL} is one of the most challenging areas of \ac{DTL}, as it assumes that the feature spaces of the source and \acp{TD} are completely different ($\chi_{S}\neq\chi_{T}$), such as speech-to-image, image-to-text, and text-to-speech. Additionally, the label spaces of the source $Y_S$ and target $Y_S$ domains may also differ ($Y_S\neq Y_T$) as described in \cite{niu2021cross}.

\noindent \textbf{(c) Unsupervised DTL:}  is a method of improving the learning of the target prediction function $\mathbb{F}T$ in the \ac{TD} $\mathbb{D}{T}$ by using knowledge from the \acp{SD} $\mathbb{D}{S}$ and the source task $\mathbb{T}{S}$, even when the labels $Y_S$ and $Y_T$ are not present. It is important to note that the source and target tasks, $\mathbb{T}{S}$ and $\mathbb{T}{T}$, are related but distinct. Such kind of approach is useful when labels are not available for \ac{TD} data, as stated in \cite{si2021unsupervised}.

\subsubsection{Adversarial DTL} 
Adversarial learning, as introduced in \cite{zhou2020adversarial}, is a method that helps to learn more transferable and discriminative representations. The first method using this approach, the domain-adversarial neural network (DANN) was presented in \cite{ganin2016domain}. Unlike traditional methods that use predefined distance functions, DANN utilizes a domain-adversarial loss within the network. This approach has shown to improve the network's ability to learn discriminative data and has been used in many visual surveillance studies \cite{shen2018crowd,georgescu2020background,choi2021unsupervised,soleimani2021cross}. However, prior works in DANN have not considered the different effects of marginal and conditional distributions. An alternative approach, dynamic distribution alignment, was proposed in \cite{wang2020transfer} that dynamically evaluates the importance of each distribution, thus providing a more nuanced approach."
Another approach called the adversarial scoring network (ASNet) was introduced in \cite{zou2021coarse} to bridge the gap between domains at different levels of granularity. This approach uses adversarial learning to align the \acp{SD} with the \ac{TD} in the global and local feature space during the coarse-grained stage. The transferability of source attributes is then evaluated in fine-grained stage by comparing the similarity between the source and target samples at multiple levels, utilizing the generative probability obtained in the coarse stage. The transferable elements are then selectively used to assist the \ac{DTL} adaptation process. This coarse-to-fine architecture effectively reduces the problem of domain disparity. Specifically, photographs are encoded into density maps by a generator, and then classified as \acp{SD} or \ac{TD} using a dual-discriminator. Adversarial training is employed between the dual-discriminator and generator to bring the domains' distributions closer together. The dual-discriminator also generates four different scores which are used as a signal to optimize the density of the \acp{SD} during adaptation for fine-grained transfer as stated in \cite{zou2021coarse}.

\begin{figure}[t!]
\begin{center}
\includegraphics[width=1\textwidth]{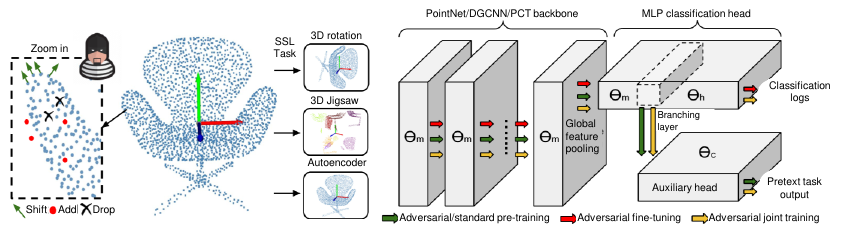}\\
\end{center}
\caption{Flowchart of the adversarial TL-based \ac{3DPC} classification approach proposed in \cite{sun2021adversarially}.}
\label{PCT_adv_exp}
\end{figure}

In \cite{sun2021adversarially}, the authors studied the adversarial robustness in \ac{3DPC} recognition using three different architectures: \ac{MLP} network (PointNet), convolutional network (DGCNN), and transformer-based network. They employed two methods: adversarial pretraining for fine-tuning, where \ac{SSL} tasks are utilized for pretraining, and adversarial joint training, where the self-supervised task is trained alongside the recognition task, as illustrated in Fig. \ref{PCT_adv_exp}.

\subsection{Popular DL-based PC models}
The development of DL-based 3DPC approaches is characterized by a rich diversity of deep learning models each aimed at overcoming specific challenges related to point cloud data processing. While these models offer significant improvements in processing speed, accuracy, and applicability, they also underscore the ongoing need for models that can generalize across different environments and handle the inherent complexities of 3D point data more effectively.

\subsubsection{Feature Extraction and Geometric Detail Enhancement Models}
Models such as PointNet \cite{qi2017pointnet} and PointNet++ \cite{qi2017pointnet++} spearheaded the direct processing of point clouds by respecting permutation invariance and enhancing the capture of hierarchical structures, respectively. They are pivotal in tasks like object classification and segmentation, though they struggle with local detail due to uniform point processing. PointVGG \cite{li2021pointvgg} and PointPAVGG \cite{shi2021pointpavgg} extend these capabilities by adapting image-based convolutional and attention mechanisms to point clouds, striving to bridge gaps in capturing intricate geometric details but facing challenges in computational efficiency. SpiderCNN \cite{xu2018spidercnn} and PointCNN \cite{li2018pointcnn} further innovate by adapting traditional CNN transformations to unordered point data, enhancing the models' ability to learn from complex datasets.

\subsubsection{Segmentation and Object Detection Models}
Models like SplatNet \cite{su2018splatnet}, SGPN \cite{wang2018sgpn}, and FoldingNet \cite{yang2018foldingnet} represent advances in segmentation and unsupervised learning. SplatNet leverages sparse bilateral convolutional layers for processing high-dimensional lattices in point clouds, suitable for segmentation but hampered by scaling issues. SGPN focuses on instance segmentation by predicting semantic classes and point groupings, which becomes complex in cluttered environments. FoldingNet explores unsupervised learning with a folding-based decoder that may not generalize across all point cloud types.

\subsubsection{Data Integration and Upsampling Models}
PU-Net \cite{yu2018pu} and PointGrid \cite{le2018pointgrid} focus on upsampling and integrating grid-based approaches with point processing. PU-Net enhances point cloud density using multi-level feature integration, while PointGrid blends point and grid-based methods for recognizing 3D models, though it struggles with sparse point clouds.

\subsubsection{Specialized Application Models}
Models such as Hand PointNet \cite{ge2018hand} and PointNetVLAD \cite{uy2018pointnetvlad} are tailored for specific applications like 3D hand pose estimation and global descriptor extraction, critical for place recognition. However, these models often face limitations when applied outside their intended scope, such as dynamic environments or varying object types.

\subsubsection{Innovative Approaches in Object Detection and Registration}
Emerging models like PREDATOR \cite{huang2021predator} and 3DIoUMatch \cite{wang20213dioumatch} showcase the potential of deep learning in object detection and registration. PREDATOR addresses low-overlap point cloud registration with a deep attention mechanism, suited for benchmark scenarios but limited in broader applications. 3DIoUMatch leverages a semi-supervised learning approach to enhance 3D object detection, showing dependency on initial label quality for performance efficacy. Table \ref{tab2} summarizes the most popular 3DPC models proposed in the literature.

\renewcommand{\arraystretch}{1.5}

\begin{center}
\color{black}
\scriptsize
\begin{longtable}[!t]{
m{0.3cm}
m{1.4cm}
m{4.5cm}
m{2.5cm}
m{2.5cm}
m{3cm}}
\caption{Summary of popular DL-based \ac{3DPC} understanding models. }
\label{tab2}\
\\ \hline
\textbf{Ref.} & \textbf{Model Name} & \textbf{Contribution Description} & \textbf{Dataset} & \textbf{Application} & \textbf{Limitation} \\ \hline
\endfirsthead
\textbf{Ref.} & \textbf{Model Name} & \textbf{Contribution Description} & \textbf{Dataset} & \textbf{Application} & \textbf{Limitation} \\ \hline
\endhead
\endfoot
\endlastfoot
\cite{qi2017pointnet} & PointNet & Introduced a neural network that processes point clouds directly, respecting permutation invariance. & Various 3D benchmarks & Object classification, part segmentation, semantic parsing & Does not capture local structural details. \\ \hline
\cite{dai2017scannet} & ScanNet & Developed an RGB-D video dataset with extensive annotations for deep learning applications. & ScanNet dataset & 3D object classification, semantic voxel labeling, CAD model retrieval & Limited diversity in scene views and semantic annotations. \\ \hline
\cite{riegler2017octnet} & OctNet & Proposed a sparse 3D data representation that allows for deep, high-resolution 3D convolutional networks. & Not specified & 3D object classification, orientation estimation, point cloud labeling & Focus on sparse data might not generalize to denser datasets. \\ \hline
\cite{qi2017pointnet++} & PointNet++ & Extended PointNet to capture local structures using a hierarchical neural network. & Challenging benchmarks of 3DPCs & Enhanced 3D recognition tasks & Performance drops with non-uniform point densities. \\ \hline
\cite{li2021pointvgg} & PointVGG & Introduced point convolution and pooling methods to adapt image-based techniques for point clouds. & Challenging benchmarks of 3DPCs & Object classification, part segmentation & May not fully capture intricate geometric details. \\ \hline
\cite{shi2021pointpavgg} & PointPAVGG & A VGG-based network that incorporates a point attention mechanism for feature extraction from point clouds. & ShapeNet, ModelNet & Point cloud classification, segmentation & Increased computational demand due to complex feature integration. \\ \hline
\cite{su2018splatnet} & SplatNet & Utilized sparse bilateral convolutional layers for processing point clouds in a high-dimensional lattice. & Not specified & 3D segmentation & Scaling issues with memory and computational cost in larger lattices. \\ \hline
\cite{yang2018foldingnet} & FoldingNet & Developed a deep auto-encoder with a folding-based decoder for unsupervised learning on point clouds. & Not specified & Unsupervised learning, 3D object reconstruction & Generic decoder structure may not work for all point cloud types. \\ \hline
\cite{yu2018pu} & PU-Net & Presented a method for upsampling 3DPCs using multi-level features. & Synthesis and scan data & Point cloud upsampling & Focus on upsampling may not improve other manipulations. \\ \hline
\cite{li2018so} & SO-Net & Built a permutation invariant architecture using Self-Organizing Maps for point cloud processing. & Not specified & Point cloud reconstruction, classification, segmentation, shape retrieval & Requires tuning of the network’s receptive field. \\ \hline
\cite{yang2018pixor} & PIXOR & Developed a real-time 3D object detection system from point clouds using BEV, optimized for autonomous driving. & KITTI, large-scale 3D vehicle detection benchmark & Real-time 3D object detection in autonomous driving & Primarily optimized for vehicle detection, may not generalize to other object types. \\ \hline
\cite{wang2018sgpn} & SGPN & Introduced a network for 3D instance segmentation by predicting point groupings and semantic classes. & Various 3D scenes & 3D instance segmentation, object detection, semantic segmentation & May struggle with highly cluttered or complex environments. \\ \hline
\cite{zhou2018voxelnet} & VoxelNet & Unified feature extraction and bounding box prediction for 3DPCs into a single deep network. & KITTI & 3D detection of cars, pedestrians, and cyclists & High computational cost due to dense voxelization of point clouds. \\ \hline
\cite{ge2018hand} & Hand PointNet & Developed a hand pose regression network using 3DPCs to capture complex hand structures. & Three challenging hand pose datasets & 3D hand pose estimation & Focuses on hand pose, limiting its applicability to other forms of point cloud processing. \\ \hline
\cite{uy2018pointnetvlad} & PointNetVLAD & Proposed a network for global descriptor extraction from point clouds for place recognition. & Created benchmark datasets for point cloud based retrieval & Place recognition from point clouds & May not be as effective in highly dynamic environments. \\ \hline
\cite{deng2018ppfnet} & PPFNet & Introduced a network for learning globally informed 3D local feature descriptors from point clouds. & Not specified & Finding correspondences in unorganized point clouds & Dependency on the quality of local features and global context understanding. \\ \hline
\cite{le2018pointgrid} & PointGrid & Combined point and grid-based approaches to recognize 3D models from point clouds. & Popular shape recognition benchmarks & 3D model recognition, classification, and segmentation & Might not handle extremely sparse or irregular point clouds effectively. \\ \hline
\cite{xu2018pointfusion} & PointFusion & Leverages image and point cloud data for 3D object detection without dataset-specific tuning. & KITTI, SUN-RGBD & Generic 3D object detection across diverse environments & The fusion process can be complex and computationally intensive. \\ \hline
\cite{qi2018frustum} & Frustum PointNets & Operates on raw point clouds for 3D object detection, combining 2D and 3D detection methods. & KITTI, SUN RGB-D & 3D object detection in both indoor and outdoor scenes & Might encounter difficulties with very sparse point clouds or heavy occlusions. \\ \hline
\cite{yew20183dfeat} & 3DFeat-Net & Learns 3D feature detectors and descriptors for point cloud matching using weak supervision. & Outdoor Lidar datasets & Point cloud matching for localization and mapping & Performance highly dependent on the effectiveness of weak supervision learning mechanisms. \\ \hline
\cite{xu2018spidercnn} & SpiderCNN & Developed SpiderConv units to handle 3DPCs & ModelNet40 & 3DPC classification and segmentation & May struggle with very large or noisy datasets \\ \hline
\cite{li2018pointcnn} & PointCNN & Introduced a X-transformation to handle unordered point clouds for CNNs & Multiple challenging benchmark datasets & Feature learning from point clouds & Dependent on the effectiveness of the X-transformation \\ \hline
\cite{wu2018squeezeseg} & SqueezeSeg & Created a CNN pipeline for semantic segmentation of LiDAR data & KITTI, GTA-V (simulated) & Semantic segmentation in autonomous driving & Reliance on synthetic data for improved accuracy \\ \hline
\cite{yan20222dpass} & 2DPASS & Boosted point cloud learning via 2D image fusion during training & SemanticKITTI, NuScenes & Semantic segmentation of point clouds & Requires multi-modal data during training, complex implementation \\ \hline
\cite{huang2021predator} & PREDATOR & Focused on low-overlap point cloud registration with deep attention & 3DMatch benchmark & Point cloud registration & Limited to pairwise registration, may not generalize beyond benchmark scenarios \\ \hline
\cite{wang20213dioumatch} & 3DIoUMatch & Implemented a semi-supervised 3D object detection with teacher-student learning & ScanNet, SUN-RGBD, KITTI & 3D object detection & High task complexity and dependency on initial label quality \\ \hline
\cite{chen2020pointposenet} & PointPoseNet & Developed a pipeline for 6D object pose estimation from point clouds & LINEMOD, Occlusion LINEMOD & 6D pose estimation & Limited to known objects in complex scenes, sensitivity to occlusions \\ \hline
\cite{cao2020asap} & ASAP-Net & Enhanced spatio-temporal modeling of point clouds & Synthia, SemanticKITTI & Point cloud sequence segmentation & Requires specific attention and structure-aware algorithms for optimal performance \\ \hline
\cite{biswas2020muscle} & MUSCLE & Proposed a compression algorithm for LiDAR data using spatio-temporal relationships & UrbanCity, SemanticKITTI & LiDAR data compression & Efficiency depends on the variability of the input data \\ \hline
\cite{zhang2021pc} & PC-RGNN & Addressed sparse and partial point clouds for 3D detection using GNNs & KITTI & 3D object detection & Highly dependent on the quality of initial point cloud data \\ \hline
\end{longtable}
\end{center}

\color{black}
\subsection{Pre-trained 2D models}

Xu et al. \cite{xu2021image2point} demonstrate how pretrained 2D image models can be adapted for 3DPC understanding with minimal modification. By extending 2D ConvNets and vision transformers to handle 3D data, the method involves inflating 2D filters to 3D and only finetuning specific layers like the input, output, and normalization layers. This approach leverages the deep feature representations learned from large-scale 2D image datasets, enabling the models to perform competitively on 3DPC tasks such as classification and segmentation, while significantly reducing the training time and data requirements compared to training from scratch. The transferability is facilitated by the similarity in feature representation between the 2D and 3D tasks, despite their differences in data modality.

Besides, shape classification and part segmentation pose significant challenges in \ac{CV}. While trained \acp{CNN} have shown impressive performance on regular grid data like images, accurately capturing shape information and geometric representation from irregular and disordered point clouds is problematic. 
To address this, \cite{li2021pointvgg} proposes point convolution (Pconv) and point pooling (Ppool) techniques inspired by convolution and pooling in image processing, specifically designed for point clouds to learn high-level features. Pconv gradually magnifies receptive fields to capture local geometric information, while Ppool tackles the disorder of point clouds using a symmetric function that aggregates points progressively for a more detailed local geometric representation. Our novel network, named PointVGG, incorporates Pconv, Ppool, and a graph structure for feature learning in point clouds, and is applied to object classification and part segmentation tasks. Experimental results demonstrate that PointVGG achieves state-of-the-art performance on challenging benchmarks of \ac{3DPC}.
Moreover, extracting high-level features from disordered point cloud data using pre-trained 2D \ac{CNN} remains challenging. To address this, \cite{shi2021pointpavgg} proposes a VGG-based network called point positional attention VGG (PointPAVGG), inspired by the classical VGG network. Our approach combines global and local features by extracting local geometric information from every sphere domain and analyzing the global position score using our point attention (PA) module. PointPAVGG, with its graph structure point cloud feature extraction and PA, is applied to point cloud classification and segmentation tasks. Through comprehensive experiments on ShapeNet and ModelNet, our method demonstrates superior performance, achieving state-of-the-art results in classification and segmentation tasks.

\color{black}
The work in \cite{mattheuwsen2020manhole} presents a method for automatic detection of manhole covers from mobile mapping point cloud data, which are large-scale spatial databases used for various purposes. The method uses a fully \ac{CNN} with \ac{DTL} and a simplified class activation mapping (CAM) location algorithm to accurately determine the position of manhole covers. Different source model architectures, such as AlexNet, VGG-16, Inception-v3, and ResNet-101, are assessed. Results showed that  VGG-16 achieved the best detection performance among others, with recall, precision, and F2-score of 0.973 each. The approach also achieves a horizontal 95\% confidence interval of 16.5 cm for location performance using VGG-16 architecture. The study highlights the importance of incorporating geometric information channels in the ground image for improved detection and location accuracy.
Aerial imaging using drones, an efficient timely data collection after natural hazards for post-event management, provides detailed site characterization with minimal ground support, but results in large amounts of 2D orthomosaic images and \ac{3DPC}. Effective data processing workflows are needed to identify structural damage states. Liao et al. \cite{liao2020deep} introduce two \ac{DL} models, based on 2D and 3D \acp{CNN}, for post-windstorm classification. \ac{DTL} from AlexNet and VGGNet is used for the 2D \acp{CNN}, while a 3D fully convolutional network (3DFCN) with skip connections is developed and trained for point cloud data. The models are compared using quantitative performance measures, and the 3DFCN shows greater robustness in detecting different damage classes. This highlights the importance of 3D datasets, particularly depth information, in distinguishing between different damage states in structures.

\color{black}
In \cite{bourbia2022blind}, a \ac{DL}-based approach is proposed, which involves projecting \ac{3DPC} into 2D rendering views and then feeding them into a \ac{CNN} for quality score prediction. \ac{DTL} is employed to leverage the capabilities of VGG-16 trained on the ImageNet database. The performance of the proposed model is evaluated on two benchmark databases, ICIP2020 and SJTU, and the results show a strong correlation between the predicted and subjective quality scores, outperforming state-of-the-art point cloud quality assessment models.
\cite{leroy2021pix2point} addresses the problem of learning outdoor \ac{3DPC} from monocular data using a sparse ground-truth dataset. A \ac{DL}-based approach called Pix2Point is proposed, which uses a 2D-3D hybrid neural network architecture and a supervised end-to-end minimization of an optimal transport divergence between point clouds. The proposed approach outperformed efficient monocular depth methods when trained on sparse point clouds. The paper highlights the potential of \ac{DL} for monocular \ac{3DPC} prediction and its ability to handle complete and challenging outdoor scenes. The encoding block, in the target model, consists of convolution, pooling, and normalization layers to extract feature descriptions from the RGB image. These features are then processed by a fully connected layer to obtain a preliminary set of 3D point coordinates. Several source models, based on VGG, DenseNet, and ResNet architectures, are explored and compared, which are referred to as backbones.

\color{black}
Balado et al. \cite{balado2020transfer} propose in 2020, a method that minimizes the use of point cloud samples for training \acp{CNN} by converting point clouds to images (pc-images). This enables the generation of multiple samples per object through multi-view, and the combination of pc-images with images from online datasets such as ImageNet and Google Images. Results suggest keeping some point cloud images in training; even 10\% can lead to high classification accuracy.
The work presented in \cite{stojanovic2019service} showcases a prototypical implementation of a service-oriented architecture for classifying indoor point cloud scenes in office environments. This approach utilizes multi-view techniques for semantic enrichment of captured scans and subsequent classification. The approach is tested using a pretrained \ac{CNN} model, Inception V3, to classify common office furniture objects such as chairs, sofas, and desks in \ac{3DPC} scans. The results show that the approach can achieve acceptable accuracy in classifying common office furniture, based on RGB cubemap images of the octree partitioned areas of the \ac{3DPC} scan. Additional methods for web-based 3D visualization, editing, and annotation of point clouds are also discussed.
%


\color{black}
The authors in \cite{dongyu2018object} propose a visual recognition and location method for object detection in soft robotic manipulation using RGB-D information fusion. The method involves scanning and reconstructing the environment using ORB-SLAM2, constructing an object feature database, matching point clouds using the iterative closest point (ICP) algorithm, identifying regions of interest, and using the inception-v3 model and \ac{DTL} for object recognition. The position of the object relative to the camera is obtained through correspondence between color information and point cloud data. The method showed that objects belonging to the same object have much lower matching error compared to those not belonging to the same object, and successful object identification and location were achieved through color recognition. Moving on, the authors in \cite{zhao2020als} adopt the ResNet50 \cite{he2016deep} pretrained on the ImageNet data set to extract deep features from each feature image and obtain five multi-scale and multi-view (MSMV) deep features per point.

Table \ref{tab3} presents a summary of DTL-based \ac{3DPC} models.

\begin{center}
\color{black}
\scriptsize
\begin{longtable}[!t]{
m{1cm}
m{2cm}
m{6cm}
m{6cm}}
\caption{Summary of DTL-based \ac{3DPC} models. }
\label{tab3}\\
\hline
Category & Model & Highlights & Limitations    \\ \hline
\endfirsthead
\multicolumn{4}{c}{Table \thetable\ (Continue)} \\ \hline
Category & Model & Highlights & Limitations   \\ \hline 
\endhead
\hline
\endfoot
\hline
\endlastfoot

\multirow{7}{1cm}{Point-based methods} & AltasNet \cite{yu2022part}     & In AtlasNet, a surface representation is inferred by regarding a 3D shape as a collection of parametric surface elements. & The reiterated many times largely determines the reconstruction.     \\ 

& MSN \cite{liu2020morphing}    & A sampling algorithm combines a set of parametric surface elements, which MSN predicts, with the partial input.  &  Fine-grained details of object shape aren't generated successfully.           \\ 

& ASHF-Net \cite{zong2021ashf}   & A hierarchical folding decoder with the gated skip-attention and multi-resolution completion target to exploit the local structure details of the incomplete inputs is proposed by ASHF-Net.  & The decoder [113] obtains unstructured predictions, and the surface of the results doesn't remain smooth.      \\ 
         
\multirow{4}{1cm}{Point-based methods} & PCN \cite{yuan2018pcn}  & The coarse-to-fine completion is performed by PCN, which combines the fully connected network and FoldingNet.  & The synthesis of shape details is not achievable.      \\ 

& SA-Net \cite{wen2020point}   & Hierarchical folding in the multi-stage points generation decoder is proposed by SA-Net.  & The implicit representation of the target shape from the intermediate layer, which helps refine the shape in the local region, is difficult to interpret and constrain.      \\ 

& FoldingNet \cite{yang2018foldingnet}   & It is commonly assumed in a two-stage generation process that a 2D-manifold can recover 3D objects.  & Explicitly constraining the implicit intermediate is challenging.      \\ 

& SK-PCN \cite{nie2020skeleton}   & The global structure is acquired by predicting the 3D skeleton with SK-PCN, and the surface completion is achieved by learning the displacements of skeletal points.  & The overall shapes are the sole focus of the meso-skeleton.      \\ 

\multirow{5}{1cm}{Point-based methods} & GRNet \cite{li2020grnet}     & Unordered point clouds are regularized by GRNet, which introduces 3D grids as intermediate representations.  & The resolution still governs it. A significant computational cost is incurred when a higher resolution is used.     \\ 

& VE-PCN \cite{wang2021voxel}    & The structure information is incorporated into the shape completion by VE-PCN through the utilization of edge generation.  &  High frequency components are the sole focus of the edges.          \\ 
                              
& Point-PEFT \cite{tang2024point} & Introduces a Parameter-Efficient Fine-Tuning method for 3D models, minimizing adaptation costs for downstream tasks with minimal trainable parameters. & While effective, the specialized method's broader applicability and long-term adaptability across diverse domains remain unproven. \\ 

& DAPT \cite{zhou2024dynamic} & Implement a Dynamic Adapter for point cloud analysis, offering efficient parameter use and reducing training resources significantly. & Although reducing trainable parameters, the adaptability and effectiveness in extremely diverse environments is yet to be fully assessed. \\ 

& AgileGAN3D \cite{song2024agilegan3d} & A novel framework for 3D artistic portrait stylization using unpaired 2D exemplars and advanced 3D GAN models. & The dependency on the quality and diversity of the 2D exemplars might limit the model's versatility in less controlled scenarios. \\ 

& 3D-TRAM \cite{shoukat20243d} & Combines transfer learning with a memory component to enhance 3D reconstruction from 2D images, leveraging CAD models for better accuracy. & The method's performance can vary significantly with the complexity of the scene and the quality of the available CAD models. \\ 

\hline

\end{longtable}

\end{center}

\color{black}
\subsection{Pre-trained 3D models}
Implementing a standardized approach to \ac{3DPC} neural network design has the potential to yield comparable advancements seen in the extensive research conducted on pre-training visual models, especially in the image domain. However, compared to the 2D domain, the design of neural networks for point cloud data is less mature, as evidenced by the numerous new architectures proposed recently. This is due to several factors, including the challenge of processing unordered sets \cite{zaheer2017deep}, the choice of neighborhood aggregation mechanism, which could be hierarchical \cite{qi2017pointnet++,klokov2017escape,zeng20183dcontextnet}, spatial CNN-like \cite{hua2018pointwise,xu2018spidercnn,li2018pointcnn,zhang2019shellnet}, spectral \cite{te2018rgcnn,wang2018local}, or graph-based \cite{xie2018attentional,verma2018feastnet}, and the fact that points are discrete samples of an underlying surface, which has led to the consideration of continuous convolutions \cite{boulch2020convpoint,yang2021continuous}.

\color{black}
While pre-training image models have been successful in achieving high levels of prosperity, pre-training 3D models is still in the developmental stage. To address this issue, numerous researchers have explored various \ac{SSL} mechanisms that utilize different pretext tasks, such as solving jigsaw puzzles \cite{sauder2019self}, estimating orientation \cite{poursaeed2020self}, and reconstructing deformations \cite{achituve2021self}. Drawing inspiration from pre-training strategies in the image domain, several approaches have been proposed in the 3D domain, including point contrast \cite{xie2020pointcontrast}, which uses a contrastive learning principle, and OcCo \cite{wang2021unsupervised}, Point-BERT \cite{yu2022point}, and Point-M2AE \cite{zhang2022point}, which introduce reconstruction pretext tasks to facilitate better representation learning. However, the lack of available data in the 3D domain remains a significant obstacle in developing more effective pre-training strategies.

\color{black}
In this respect, Choy et al. proposed the Minkowski Engine \cite{choy20194d}, which is an extension of sub-manifold sparse convolutional networks \cite{graham20183d} to higher dimensions. By facilitating the adoption of common deep architectures from 2D vision, sparse convolutional networks could help standardize \ac{DL} for point cloud. In \cite{xie2020pointcontrast}, the authors use a unified UNet \cite{ronneberger2015u} architecture built with Minkowski Engine as the backbone network in all our experiments and show that it can seamlessly transfer between tasks and datasets. \cite{zhang2022point} The paper proposes Point-M2AE scheme, a \ac{MAE} pre-training framework for \ac{SSL} of \ac{3DPC}. Unlike standard \ac{AE}-based transformers, Point-M2AE modifies the encoder and decoder into pyramid architectures to model spatial geometries and capture fine-grained and high-level semantics of 3D shapes. The encoder uses a multi-scale masking strategy for consistent visible regions across scales and a local spatial self-attention \cite{kheddar2024automatic} mechanism during fine-tuning. The lightweight decoder gradually upsamples point tokens with skip connections from the encoder, promoting reconstruction from a global-to-local perspective. Point-M2AE achieves 92.9\% accuracy on ModelNet40 using a linear SVM. Fine-tuning enhances performance to 86.43\% on ScanObjectNN and provides benefits in various tasks, such as few-shot classification, part segmentation, and 3D object detection.

\color{black}
The reference \cite{zhang2022pointclip} introduces the PointCLIP method, which encodes point clouds by projecting them into multi-view depth maps and aggregates view-wise zero-shot predictions for knowledge transfer from 2D to 3D. An inter-view adapter is designed to extract global features and adaptively fuse few-shot knowledge from 3D into CLIP pre-trained in 2D. Fine-tuning the lightweight adapter in few-shot settings significantly improves PointCLIP's performance. PointCLIP also exhibits complementary properties with classical 3D-supervised networks, and ensembling with baseline models further boosts performance, surpassing state-of-the-art models. PointCLIP is a promising alternative for effective \ac{3DPC} understanding with low resource cost and data regime, as demonstrated through experiments on ModelNet10, ModelNet40, and ScanObjectNN datasets.
%
Huang et al. \cite{huang2022transfer} proposes a method called MF-PointNN for surrogate modeling of melt pool in metallic additive manufacturing. Melt pool modeling is important for uncertainty quantification and quality control in metal additive manufacturing, but finite element simulation for thermal modeling can be time-consuming. MF-PointNN is a multi-fidelity approach that combines low-fidelity analytical models and high-fidelity  finite element simulation data using \ac{DTL}. A basic PointNN is first trained with low-fidelity data to establish correlation between inputs and thermal field of analytical models. Then, the basic PointNN is updated and fine-tuned using a small amount of high-fidelity data to build the MF-PointNN. This latter efficiently maps input variables and spatial positions to thermal histories, allowing for efficient prediction of the three-dimensional melt pool. Results of melt pool modeling for Ti-6Al-4V in electron beam additive manufacturing under uncertainty show the effectiveness of the proposed approach.

\subsection{Evaluation metrics}

\label{sec:EM}

Several metrics have been reported to evaluate the performance of different \ac{DTL} approaches for various \ac{3DPC} tasks. F1 measures and \ac{OA}, are the most frequently used measures to assess the performance of algorithms over benchmark datasets. These metrics are used for many purposes, including segmentation, classification, and registration. F1 measures give an idea about the behavior of the precision and recall curve, whereas \ac{OA} conveys the mean accuracy for instances of the test. Besides, \ac{IoU} and mean \ac{IoU} have been extensively used, especially for object detection and segmentation purpose. For instance, in \cite{zhao2018semantic}, authors have used mean \ac{IoU} between point-wise ground truth and prediction. In \cite{zhang2020unsupervised}, authors have computed \ac{IoU} for different shapes. It is calculated as the average of \acp{IoU} of all parts in a shape. Furthermore, for some particular categories, mean IoUs (mIoUs) are obtained by finding out the average of all  \acp{IoU} shapes. Other metrics, instance mIoU (Ins. mIoU) and category mIoU (Cat. mIoU) are also introduced and computed by calculating the average of all shapes and the average of mIoUs for all categories, respectively.

Metrics such as OA and mIoU provide a holistic view of the model's performance across different categories. Additionally, the frame per second (FPS) metric evaluates the efficiency and speed of real-time applications, while root mean squared error (RMSE) and mean absolute percentage error (MAPE) are used for precision in measurement tasks. 
Moving on, consistency rate (CR), consistency proportion (CP), and weight coverage (WCov) are advanced evaluation metrics used to assess the quality of 3D point cloud reconstructions. CR measures the rate at which a reconstructed scene maintains geometric consistency across different views or instances. CP evaluates the proportion of consistent data points within the entire dataset, providing a sense of how uniformly the model performs. WCov assesses the coverage of the weighted areas in the point cloud, indicating how well the model captures the essential features and structures of the scene. 


To evaluate the performance or semantic segmentation, authors in \cite{murtiyoso2021semantic} have used errors of commission and errors of omission metrics over the CMP façade dataset.  Frame per second (FPS) metric is used by \cite{imad2021transfer} on the ouster LiDAR-64 dataset. In addition to this, Zong et al. \cite{zong2022improved} have used average precision (mPrec) and average recall (mRec) to test their segmentation method. \textcolor{black}{Moreover, the best performance for \ac{OA} on the Shapenet dataset is 94.5\% \cite{bazazian2020dcg}. However, the highest precision (85.99\%) is achieved on CMP facade dataset \cite{murtiyoso2021semantic}. The F1 score has been calculated on different datasets, viz. CMP facade, ISPRS, Semantic 3D dataset, ModelNet4, and other. The best performance has been reported by \cite{arnold2021automatic} on ModelNet40.} \textcolor{black}{The aforementioned metrics are not only applicable to \ac{3DPC} tasks but also to other artificial intelligence-related tasks such as classification, segmentation, and other. However, subsequent subsections thoroughly elaborate on specific metrics relevant to \ac{3DPC} tasks.}

\color{black}
\subsubsection{Distance metrics}
Distance metrics play a pivotal role in the processing of point clouds for tasks such as nearest-neighbor search, clustering, and segmentation. They provide a measure of similarity or dissimilarity between points in space, influencing the outcomes of many algorithms.

\subsubsection{Euclidean Distance}
Defined as the square root of the sum of the squared differences between the coordinates of two points:
\[
d(\mathbf{p}, \mathbf{q}) = \sqrt{(p_1 - q_1)^2 + (p_2 - q_2)^2 + \cdots + (p_n - q_n)^2}
\]
where \( \mathbf{p} = (p_1, p_2, \ldots, p_n) \) and \( \mathbf{q} = (q_1, q_2, \ldots, q_n) \) represent the coordinates of two points in an \( n \)-dimensional space. The Euclidean distance \( d(\mathbf{p}, \mathbf{q}) \) calculates the "as-the-crow-flies" distance between the two points, effectively measuring the length of the straight line segment that connects them.

\subsubsection{Manhattan Distance}
Also known as taxicab or city block distance, it is the sum of the absolute differences of their coordinates:
\[
d(\mathbf{p}, \mathbf{q}) = |p_1 - q_1| + |p_2 - q_2| + \cdots + |p_n - q_n|
\]
This metric is ideal for grid-based and urban environments where travel paths are constrained to grid layouts.

\subsubsection{Mahalanobis Distance}
Takes into account the correlations of the dataset and is scale-invariant, defined as:
\[
d(\mathbf{x}, \mathbf{\mu}) = \sqrt{(\mathbf{x} - \mathbf{\mu})^T \mathbf{S}^{-1} (\mathbf{x} - \mathbf{\mu})}
\]

where:
\begin{itemize}
    \item \(\mathbf{x}\) is the vector representing the point whose distance from the distribution is being measured.
    \item \(\mathbf{\mu}\) is the mean vector of the distribution, representing the central tendency.
    \item \(\mathbf{S}^{-1}\) is the inverse of the covariance matrix \(\mathbf{S}\) of the distribution, which adjusts the distance measure to account for the spread and orientation of the data points.
\end{itemize}

This formulation accounts for the shape of the data distribution, correcting distances based on how data spreads and correlates across dimensions, thus providing a more nuanced measure of distance.

\subsubsection{Hausdorff distance}
Let us assume that there are two point clouds $S_1$ and $S_2$ in $R^3$ with $N_1$ and $N_2$ points in each cloud, respectively. In order to compare these point sets, one of the earliest methods that was proposed is the Hausdorff distance ($\mathcal{D}_{H}$), which is based on finding the minimum distance from a point to a set \cite{urbach2020dpdist}:
\begin{equation}
d(x,y)=\left\Vert x-y\right\Vert _{2}
\end{equation}
\begin{equation}
D(x,S)=\underset{y\in S}{\text{min }  } d(x,y)
\end{equation}
and calculates a symmetric max min distance \cite{huttenlocher1993comparing}:
\begin{equation}
\mathcal{D}_{H}(S_{1},S_{2})=\max \left\{ \underset{a\in S_{1}}{\max }%
D(a,S_{2}),\underset{b\in S_{2}}{\max } D(b,S_{1})\right\} 
\end{equation}

\subsubsection{Chamfer distance}
The \ac{CD} is a metric used to evaluate the similarity between two sets of points in space. It considers the distance between each point in both sets and finds the nearest point in the other set, then sums the square of these distances.\ac{CD} is commonly used in the ShapeNet's shape reconstruction challenge. The \ac{CD} between two point clouds, S1 and S2, is defined as \cite{nguyen2021point}:
\begin{equation}
\mathcal{D}_{CD}\left( S_{1},S_{2}\right) =\frac{1}{\left\vert S_{1}\right\vert }%
\sum_{x\in S_{1}}\underset{y\in S_{2}}{\min }\left\Vert x-y\right\Vert
_{2}^{2}+\frac{1}{\left\vert S_{2}\right\vert }\sum_{y\in S_{2}}\underset{%
x\in S_{1}}{\min }\left\Vert x-y\right\Vert _{2}^{2}
\end{equation}

\subsubsection{Earth mover's distance}
In contrast to the Hausdorff distance and its derivative methods that involve identifying the closest neighboring point for each point, the Earth mover's distance (EMD) or the Wasserstein distance operates by establishing a one-to-one correspondence (i.e., bijection represented by $\zeta$) between the two sets of points. The objective is to minimize the total distance between the corresponding points in the two sets \cite{urbach2020dpdist}.
\begin{equation}
\mathcal{D}_{EMD}(S_{1},S_{2})=\underset{\zeta :S_{1}\longrightarrow S_{2}}{\min }%
\sum_{a\in S_{1}}\left\Vert a-\zeta (a)\right\Vert _{2}
\end{equation}

\color{black}

\subsection{Datasets}   \label{sec:dataset}

There is a considerable number of datasets available for performing various \ac{3DPC} tasks using state-of-the-art techniques. A detail about datasets on which \ac{DL} has been implemented is reported in \cite{guo2020deep}. However, in our study, we have considered the datasets for which \ac{DTL} has been incorporated to carry out \ac{3DPC} tasks. These include widely used benchmarks like, PointNet \cite{xie2021automatic}, ShapeNet \cite{xie2020pointcontrast}, ModelNet \cite{zhang2020unsupervised}, ScanNet \cite{xie2020pointcontrast},   ISPRS \cite{lei2020point}, KITTI 3D object detection \cite{imad2021transfer}, \textcolor{black}{Campus3D \cite{li2020campus3d}} and semantic 3D dataset \cite{kim2022deep}. Apart from these well-known benchmarks, in this study, newly introduced datasets are also included. These comprises of Scaffolds dataset \cite{kim2022deep}, CMP façade dataset \cite{murtiyoso2021semantic}, DFC \cite{lei2020point}, Ouster LiDAR-64 \cite{imad2021transfer}, Santa Monica point cloud data \cite{chen2021classification}, UTD-MHAD \cite{sidor2020recognition}, PointDA-10 \cite{tian2021vdm}, NYU \cite{huang2020superb}, SHREC2018 \cite{benhabiles2018transfer}, and EndoSLAM dataset \cite{ozyoruk2021endoslam}. 

These datasets are used for \ac{3DPC} segmentation (semantic, panoptic, and instance), classification, and object detection and tracking and contain features beneficial for carrying out related tasks. For instance, datasets like ModelNet 40, ShapeNet and  ScanNet, among others, are used for shape classification and contain columns like sample size, classes, training, testing data percentage, and other useful features. In addition to this, SHREC2018 and EndoSLAM are medical-related datasets. The latter is created through the recording of multiple endoscope cameras for six porcine organs, as well as synthetically generated records. However, the former, the SHREC2018 protein dataset, contains 2267 protein details. To record the details, the protein data bank (PDB) format has been used. The datasets, references, and available links are given in Table \ref{dataset_em}.

\begin{table}[!t]
\caption{Dataset and evaluation metrics used by \ac{DTL} methods for various \ac{3DPC} tasks. }
\label{dataset_em}

\scriptsize
\begin{tabular}{
m{0.5cm}
m{2.8cm}
m{5cm}
m{2.5cm}
m{2.8cm}
}
\hline
Ref. & Dataset & Dataset availability & Evaluation metric & Best performance  (\%) \\ \hline
\cite{xie2020pointcontrast} & ShapeNet, ScanNet & \url{http://www.scan-net.org/#code-and-data} & - &   - \\ 

\cite{xie2021automatic}& PointNet \cite{qi2017pointnet}, ShapeNet \cite{chang2015shapenet}, and ModelNet-40 & \url{http://stanford.edu/rqi/pointnet/} \newline \url{https://shapenet.org/} \newline \url{https://modelnet.cs.princeton.edu/} &  - &  -  \\ 

\cite{eckart2021self} & UTD-MHAD & \url{http://www.utdallas.edu/kehtar/Kinect2Dataset.zip} & Recognition rate &  92.50  \\ 

\cite{arnold2021automatic} &  ModelNet40 &  \url{https://modelnet.cs.princeton.edu/} & F1, precision, recall  &  F1 (91)  \\ 

\cite{lee2021progressive} & ModelNet & \url{https://modelnet.cs.princeton.edu/} & MAE-T, MAE-F &  CA (97.6)  \\ 

\cite{zhang2020unsupervised} & ShapeNet, ModelNet & \url{https://shapenet.org/} \newline  \url{https://modelnet.cs.princeton.edu/} & OA & 90.40  \\ 

\cite{lei2020point}&  ISPRS, DFC &  \url{https://www.isprs.org/data/} &  F1, \ac{OA}  &  F1 (83.62), OA (89.84)  \\ 

\cite{imad2021transfer} & KITTI 3D object detection, Ouster LiDAR-64 & \url{https://ouster.com/resources/lidar-sample-data/} & FPS (frame per second) & FPS (30.6)    \\ 

\cite{dai2018connecting} &  ImageNet, KITTI 2D, VLP-16 &  \url{http://www.image-net.org/about-stats} &  IoU, F1, mAP & mAP(81.27), IoU(65.11), F1 (0.82)   \\ 

\cite{kim2022deep} &  Semantic3D dataset [22], Scaffolds dataset&  \url{http://www.semantic3d.net/view_dbase.php?chl=1} & F1, precision, recall & F1 (90.84)   \\

\textcolor{black}{\cite{li2020campus3d}} & \textcolor{black}{Campus3D} & \textcolor{black}{\url{https://3d.nus.app/}} & \textcolor{black}{OA, IoU, mIoU, CR, CP and WCov}  & \textcolor{black}{OA(90.9), mIoU(61.5)} \\    

\cite{murtiyoso2021semantic} & CMP façade dataset &  \url{https://cmp.felk.cvut.cz/tylecr1/facade/}&  Precision, recall, F1, Errors of commission, Errors of omission &  Precision (85.97), recall (89.80), F1 (87.85)  \\ 

\cite{chen2021classification} & Santa Monica point cloud data, KITTI 2D, VLP-16 & \url{http://www.cvlibs.net/datasets/kitti/} &  F1, precision, recall &  -  \\ 

\cite{tian2021vdm} & PointDA-10 & - & Accuracy & 49.70  \\ 

\cite{benhabiles2018transfer} & SHREC2018 & - & precision, recall, NN, T1, T2, EM, DCG  &  -  \\ 

\cite{ozyoruk2021endoslam} &  EndoSLAM dataset & \url{https://github.com/CapsuleEndoscope/EndoSLAM} & RMSE &  -  \\

\cite{zong2022improved}&  \ac{3DPC} tunnel dataset &  - & Average precision (mPrec), average recall (mRec) &  mPrec (71.2)  \\ 

\cite{diraco2021remaining} & 2014 scans & - & MAPE, RMSE &  MAPE (0.416), RMSE (0.112)  \\ 
 
\cite{kang2018building} & NYU & - & RMSE, Accuracy &   (81.8), \\ 

\cite{bazazian2020dcg}& ShapeNetPart & \url{https://shapenet.org/} & OA, mIoU, Cat. mIoU, Ins. mIoU & OA (94.5)    \\

\hline



\end{tabular}

\end{table}

\section{Overview of DTL} \label{sec3}
Arguably one of the top success stories of \ac{DL} is \ac{DTL}. Many applications in language and vision have benefited from the discovery that pretraining a network on a rich source set (e.g., ImageNet) can help boost performance once fine-tuned on a typically much smaller target set. Yet, very little is known about its usefulness in \ac{3DPC} understanding. We see this as an opportunity considering the effort required for annotating data in 3D.

\textcolor{black}{
\subsection{Domain adaptation}
\Acf{DA} has shown significant improvements in various \ac{ML} and \ac{CV} tasks, such as classification, detection, and segmentation. However, there are limited methods that have achieved \ac{DA} directly on \ac{3DPC} data, to the best of our knowledge. The challenge of point cloud data lies in its rich spatial geometric information, where the semantics of the entire object are contributed by regional geometric structures. Most general-purpose DA methods that focus on global feature alignment and disregard local geometric information may not be suitable for 3D domain alignment.
Wu and their colleagues \cite{wu2023sim2real} discusse the challenges of generating and annotating large amounts of real-world data for \ac{DL}-based approaches in robotics \ac{CV} tasks. To overcome this, the authors propose using simulation-to-reality (sim2real) \ac{DTL} for point cloud data in an industrial application case. They provide insights on generating and processing synthetic point cloud data to improve model performance when transferred to real-world data. The issue of imbalanced learning is also investigated, and the authors propose a novel patch-based attention network as a strategy to address this problem. Another work in \cite{zhou2023sampling}, presents a new method called \ac{DTL}-based sampling-attention network (TSANet) for semantic segmentation of 3D urban point clouds to facilitate the development of smart cities. The method includes a segmentation model with point downsampling–upsampling structure, embedding method, attention mechanism, and focal loss for feature processing and learning. The \ac{DTL} technique aims to reduce data requirements and labeling efforts by leveraging prior knowledge. The method is evaluated on a realistic point cloud dataset of Cambridge and Birmingham cities, demonstrating promising performance, surpassing other state-of-the-art models in terms of accuracy and mean \ac{IoU}. 
\cite{achituve2021self} introduces \ac{SSL}  for \ac{DA} in 3D perception problems, specifically on point clouds. It describes a new family of pretext tasks called deformation reconstruction, inspired by sim-to-real transformations, and proposes a novel training procedure called point cloud mixup (PCM) motivated by the MixUp method for labeled point cloud data. Evaluations on \ac{DA} datasets for classification and segmentation show significant improvement over existing and baseline methods, demonstrating the effectiveness of SSL-PCM-DA on point clouds. Similarly,  \cite{wu2019squeezesegv2} introduces a new model called SqueezeSegV2 for point cloud segmentation that is more robust to dropout noise in \ac{LiDAR} point clouds. The improved model structure, training loss, batch normalization, and additional input channel result in significant accuracy improvement when trained on real data. To overcome the challenge of limited labeled point-cloud data, the proposed scheme employs \ac{DA} training pipeline, consisting of learned intensity rendering, geodesic correlation alignment, and progressive domain calibration. When trained on real data, the new model exhibits significant segmentation accuracy improvements over the original SqueezeSeg. Moreover, when trained on synthetic data using the proposed \ac{DA} pipeline, the test accuracy on real-world data nearly doubles.} A similar scheme for segmentation is proposed in \cite{jiang2021lidarnet}, introducing LiDARNet, a boundary-aware \ac{DA} model tailored for semantic segmentation of \ac{LiDAR} point cloud data. The model uses a two-branch structure to extract domain private and shared features, and incorporates Gated-SCNN to learn boundary information during segmentation. The domain gap is further reduced by learning a mapping between domains using shared and private features. They also introduce a new dataset, SemanticUSL, for \ac{DA} in \ac{LiDAR} semantic segmentation, which has the same format and ontology as Semantic KITTI. Experiments on real-world datasets show that LiDARNet achieves comparable performance on the \ac{SD} and significant performance improvement (8-22\% mIoU) on the \acp{TD} after adaptation.

\textcolor{black}{
Other researchers have advanced unsupervised \ac{DA} approaches, demonstrated by the method outlined in \cite{yi2021complete}, aimed at enhancing semantic labeling accuracy for \ac{3DPC} in autonomous driving scenarios. The proposed approach uses a complete and label approach, leveraging a \ac{SVCN} to recover underlying surfaces of sparse point clouds and transfer semantic labels across different \ac{LiDAR} sensors. The approach does not require manual labeling for training pairs and introduces local adversarial learning to the model surface prior. Experimental results on a new benchmark dataset show significant performance improvements ranging from 8.2\% to 36.6\% compared to previous \ac{DA} methods. \textcolor{black}{Likewise, \cite{qin2019pointdan} introduces PointDAN, a 3D \ac{DA} network tailored for point cloud data, aligning global and local features at multiple levels to achieve domain alignment. It introduces a Self-adaptive node module for local alignment, which models discriminative local structures, and a node-attention module for hierarchical feature representation. For global alignment, an adversarial-training strategy is employed. PointDAN outperforms state-of-the-art \ac{DA} methods in terms of adapting \ac{3DPC} data across domains, as demonstrated on the benchmark dataset PointDA-10, created by the authors.}
Zhao et al. \cite{zhao2021epointda} proposes an end-to-end framework called ePointDA for simulation-to-real \ac{DA} (SRDA) in \ac{LiDAR} point cloud segmentation. ePointDA consists of three modules: self-supervised dropout noise rendering, statistics-invariant and spatially-adaptive feature alignment, and transferable segmentation learning. The framework bridges the domain shift at pixel-level and feature-level, without requiring real-world statistics. Experimental results on adapting from synthetic to real datasets demonstrate the superiority of ePointDA in \ac{LiDAR} point cloud segmentation. Xu and their colleagues \cite{xu2021spg} propose semantic point generation to enhance the reliability of LiDAR-based object detectors against domain shifts in autonomous driving. The scheme generates semantic points to recover missing parts of foreground objects caused by occlusions, low reflectance, or weather interference. By merging the semantic points with the original points, an augmented point cloud is obtained, which significantly improves the performance of modern LiDAR-based detectors in unsupervised \ac{DA} tasks. The method also benefits object detection in the original domain, surpassing KITTI when combined with PV-RCNN.}

\textcolor{black}{
Lang et al. \cite{lang2019pointpillars} utilize PointNets to represent point clouds organized in vertical columns. The scheme, called PointPillars, achieves superior speed and accuracy, surpassing KITTI benchmarks while running at a significantly higher frame rate of 62 Hz. A faster version achieves state-of-the-art performance at 105 Hz, making it a suitable encoding approach for object detection in point clouds in robotics applications like autonomous driving.  For object detection from \ac{3DPC}, the authors \cite{wang2022ssda3d} propose a Semi-Supervised \ac{DA} method for 3D object detection that leverages a small amount of labeled target data to improve adaptation performance. The method  consists of an inter-\ac{DA} stage, which uses a Point-CutMix module to align point cloud distribution across domains, and an intra-domain generalization stage, which employs intra-domain Point-MixUp in semi-supervised learning to enhance model generalization on the unlabeled target set. Experimental results show that the proposed scheme, with only 10\% labeled target data, outperforms a fully-supervised oracle model with 100\% target labels on the Waymo to nuScenes domain shift.} However, Du et al. \cite{du2020associate} tackle the challenge of enhancing feature representation robustness in \ac{3DPC} by employing the \ac{DA} approach. The severe spatial occlusion and point density variance in point cloud data make designing robust features crucial. The proposed approach bridges the gap between the perceptual domain (real scene) and the conceptual domain (augmented scene with non-occluded point clouds), mimicking the functionality of human perception. Experimental results show that this simple yet effective approach significantly improves the performance of \ac{3DPC} object detection, achieving state-of-the-art results.


%

\subsection{Fine-tuning}

Fine-tuning stands as a ubiquitous \ac{DTL} approach, aiding pretrained models to adapt to new tasks through iterative training, often supplemented with additional layers known as the target model. \cite{guo2020adafilter}.  The recent research studies, such as \cite{kumar2018co, li2019transferable, ge2017borrowing}, have argued that even if the target task and source task are different, transferring features via \ac{DTL} approaches outperforms random features selections. This characteristic has led to the great success of \ac{DTL} in different spheres of real-life applications such as medical imaging \cite{maqsood2019transfer, habchi2023ai}, recognition tasks like speech recognition \cite{KheddarASR2023}, hand-gesture \cite{cote2019deep}, face recognition \cite{ren2014transfer}, and rcently, in the processing of \ac{3DPC} data \cite{imad2021transfer, matrone2021transfer}. \textcolor{black}{One of the major concerns raised in fine-tuning a pretrained model is to identify which layer to fine-tune. The authors in \cite{xuhong2018explicit} have explicitly added regularization terms to the loss function for obtaining the parameters of the fine-tuned model close to the original pretrained model. AdaFilter \cite{guo2020adafilter}, is an adaptive fine-tuning approach, which considers criterion for optimization by selecting only a part of the convolutional filters in the pretrained model. Furthermore, they have exploited a recurrent gated network and considered activation of the previous layer for carefully fine-tuning the desired convolutional filters only. The adaptive fine-tuning scheme considers the similarity between the source and target tasks and datasets to reuse more pretrained filters.}

Many works that involve processing \ac{3DPC} data have exploited \ac{DTL} for many tasks, include 3D reconstructions of scaffolds \cite{kim2022deep}, \ac{3DPC} instance segmentation \cite{zong2022improved}, classification of soft-story buildings using \ac{3DPC} data \cite{chen2021classification}, \ac{3DPC} understanding \cite{xie2020pointcontrast}, 3D orientation recognition \cite{lee2021progressive}, \ac{SSL} on \acp{3DPC} \cite{eckart2021self}, and remaining useful life prediction \cite{diraco2021remaining}, among others. Furthermore, some fine-tuning strategies involve all the pretrained parameters, whereas others fine-tuning the last few layers \cite{guo2020adafilter}, as illusted in Fig. \ref{finetuning_vs_DA} (a).  G Diraco et al. \cite{diraco2021remaining} have performed an extensive experiment for a particular scenario in which the amount of data in the \ac{TD} is small and explored how it behaves if only part of the network is fine-tuned on the given target dataset. They have updated the decoder of their model, which has been trained on \acp{SD} and evaluated on the target dataset.


One factor contributing to the success of \ac{DTL} is its ability to perform well with smaller datasets. In support of this, the authors \cite{chen2021classification} have employed \ac{DTL} to address the challenge of training large parameters in deep convolutional networks. Their experimental results demonstrate the transferability of \ac{DTL} from \acp{SD} to \ac{TD}, as the dataset under consideration lacked sufficient data to train large parameters. Additionally, compared to traditional training methods, \ac{DTL} consumes less time, making it a more efficient approach.
\textcolor{black}{To support this claim, the authors noted that models such as VGGNet, Inception and ResNet, which were trained using fine-tuning techniques, exhibited shorter training times by triggering the early stopping mechanism. This improvement in training time has enabled the workflow to quickly identify soft-story buildings on a city scale. Furthermore, the authors have shown that \ac{DTL} can effectively reduce the risk of overfitting that can occur with \ac{DL} techniques. By leveraging a pretrained network, \ac{DTL} can transfer knowledge and prevent the model from memorizing the training data, leading to more robust and generalizable performance.} 
Moving on, Xiu et al. \cite{xiu2020collapsed} propose using airborne \ac{LiDAR} to detect collapsed buildings during earthquake emergency response, as \ac{DTL}-based damage detection with aerial images has limitations in detecting collapsed buildings with undamaged roofs. The authors develop a \ac{3DPC}-based dataset for building damage detection and propose a general extension framework and a visual explanation method to validate model decisions. The results conducted using PointNet \cite{qi2017pointnet}, PointNet++ \cite{qi2017pointnet++} and DGCNN \cite{wang2019dynamic} show that \ac{3DPC}-based methods can achieve high accuracy and are robust even with reduced training data. The model also achieves moderate accuracy on another dataset with different architectural styles without additional training.

\subsection{Unsupervised DTL}
Unsupervised \ac{DTL} is mainly based on \ac{UDA} to remove the need or addressing the issue of lack for labeled data and allows any image to be used as a datapoint for any \ac{CV} tasks \cite{jeon2022named, cheng2023image}.  
\textcolor{black}{
Numerous methods have been proposed for performing \ac{UDA} on 2D images, which can be divided into two main categories: methods based on domain-invariant feature learning and methods for learning domain mapping. The former \cite{kang2019contrastive,geyer2020a2d2,wang2022cross,xie2023collaborative,ren2022multi} aim to minimize the discrepancy between two distributions in the feature space, while the latter \cite{fernando2013unsupervised,bronstein2017geometric,lee2019sliced} use neural networks, such as CycleGAN \cite{qi2017pointnet++}, to directly learn the translation from the \ac{SD} to the \acp{TD}. In \cite{ledig2017photorealistic}, 2D translation is extended to depth images using a differential contrastive learning strategy for preserving underlying geometries. Despite their differences, these methods widely exploit domain adversarial training. Additionally, several useful techniques, such as pseudo-labeling \cite{huang2019texturenet} and batch normalization tailored for \ac{DA} \cite{gong2013reshaping}, have also been proposed.}

\textcolor{black}{
Although there have been significant efforts made in \ac{UDA} for 2D images and depth, \ac{UDA} on \ac{3DPC} is still in its early stages. \ac{UDA} on point clouds involves extending domain adversarial training from 2D images to \ac{3DPC} to align features on both local and global levels \cite{shen2022domain}. Nevertheless, adversarial methods on \ac{3DPC} struggle to balance local geometry alignment and global semantic alignment. CycleGAN \cite{zhu2017unpaired} is utilized by both \cite{zhao2021epointda} and \cite{saleh2019domain} to generate more realistic \ac{LiDAR} point clouds from synthetic data. This Sim2Real approach is used to minimize feature distances between the \ac{SD} and \ac{TD}. The work in  \cite{yi2021complete}, on the other hand, leverages segmentation on completed surface reconstructed from sparse point cloud for better adaptation.}

\textcolor{black}{
For object-level tasks, \cite{zhou2018unsupervised,qin2019pointdan} align global and local features, while \cite{wu2019squeezesegv2} and \cite{saleh2019domain} project point clouds to 2D and birds-eye view, respectively, to reduce sparsity. \cite{du2020associate} creates a car model set and adapts their features for detection object features, but only targets general car 3D detection on a single point cloud domain. Recently, \cite{wang2020train} published the first study targeting \ac{UDA} for 3D \ac{LiDAR} detection. They identify the vehicle size as the domain gap between KITTI \cite{geiger2013vision} and other datasets and resize the vehicles in the data. In contrast, \cite{xu2021spg} identifies point cloud quality as the major domain gap between Waymo's two datasets \cite{sun2020scalability} and proposes a learning-based approach to close the gap.}

\textcolor{black}{
Recent works on \ac{UDA} on point clouds mainly focus on designing suitable self-supervised tasks on point clouds to facilitate learning domain invariant features, which are discussed in detail in the following subsection.
In addition to \ac{UDA} on object point clouds, several methods have been proposed to address specific domain gaps on \ac{LiDAR} point clouds. These methods commonly address depth missing and sampling difference between sensors.
ST3D \cite{yang2021st3d} presents a task-specific self-training pipeline with curriculum data augmentation to further improve the adaptation process}

Alternatively, most existing \ac{UDA} approaches focus on uni-modal data, despite the availability of multi-modal datasets. 
In \cite{jaritz2020xmuda}, the authors propose a cross-modal \ac{UDA} (xMUDA) approach for 3D semantic segmentation, where both 2D images and \ac{3DPC} are utilized. This is challenging as the two input modalities are heterogeneous and can be affected differently by domain shift. In xMUDA, the modalities learn from each other through mutual mimicking, separate from the segmentation objective, to prevent the stronger modality from adopting false predictions from the weaker one. The proposed xMUDA approach is evaluated on various \ac{UDA} scenarios, such as day-to-night, country-to-country, and dataset-to-dataset shifts, using recent autonomous driving datasets. Results show that xMUDA significantly improves over uni-modal \ac{UDA} in all tested scenarios and complements state-of-the-art \ac{UDA} techniques. Saltori et al. \cite{saltori2020sf} introduces  a novel source-free \ac{UDA} (SF-UDA) framework for 3D object detection using \ac{LiDAR} point clouds, which does not require any annotations from the \ac{SD} or images/annotations from the \ac{TD}. It addresses domain shift in \ac{LiDAR} data, which is not only due to changes in environment and object appearances, but also to geometry (e.g., point density variations). SF-UDA$^{3D}$ utilizes pseudo-annotations, reversible scale-transformations, and motion coherency for \ac{DA}. Experimental results on large-scale datasets, KITTI and nuScenes,  show that SF-UDA$^{3D}$ outperforms previous methods based on feature alignment and state-of-the-art 3D object detection methods that use few-shot target annotations or target annotation statistics.
Also, \cite{cardace2021refrec} introduces RefRec, an approach that  investigate pseudo-labels and self-training for \ac{UDA} in point cloud classification. Instead of relying on multi-task learning, RefRec proposes two innovations for effective self-training on 3D data. First, it refines noisy pseudo-labels by matching shape descriptors learned from the unsupervised task of shape reconstruction on both domains. Second, it proposes a novel self-training protocol that learns domain-specific decision boundaries and mitigates the negative impact of mislabelled target samples and in-domain intra-class variability. RefRec achieves state-of-the-art performance on standard benchmarks for \ac{UDA} in point cloud classification, demonstrating the effectiveness of self-training for this emerging research problem.
Additionally, \cite{shi2022dfan} focuses on \ac{UDA} in 3D \ac{CV} tasks, specifically point cloud visual tasks. The proposed approach introduces a dual-branch feature alignment network (DFAN) architecture that leverages the characteristics of local and global features in point clouds. The approach utilizes different strategies for feature extraction and alignment in each branch, complementing each other. Hierarchical alignment for local features and distribution alignment for global features are also introduced. Experimental results on benchmark datasets demonstrate that the proposed approach achieves state-of-the-art performance in point cloud classification and segmentation tasks.

Previous studies on unsupervised 3D learning have principally concentrated on ShapeNet \cite{chang2015shapenet}, which is a repository of single-object computer aided design (CAD) models.
Typically, the idea is to use ShapeNet as the ImageNet counterpart in 3D so that the characteristics learned on single synthetic objects can be transferred to other real-world applications.
In this regard, numerous works have then been proposed. For instance, \cite{kang2022unsupervised} introduces a point-level DA by searched transformations. Typically, transformations of \acp{3DPC} are learned via the search of the best combination of operations on \acp{3DPC}, which transfers data from the \ac{SD} to the \ac{TD} while keeping the \ac{TD} unlabeled. Besides, most \ac{3DPC}-based \ac{UDA} techniques focus on extracting domain-invariant features in different domains  for feature alignment. In this regard, \ac{UDA} is advocated for the detection of 3D objects in the context of semantic point generation, as proposed in \cite{xu2021spg}.

\begin{figure}[t!]
\begin{center}
\includegraphics[width=1\textwidth]{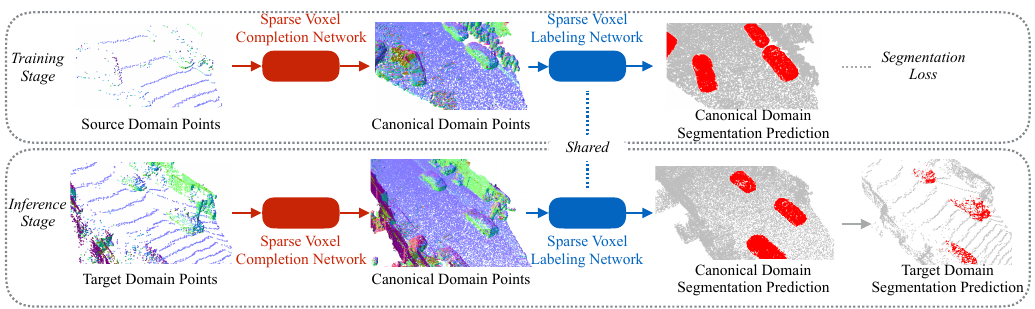}\\
\end{center}
\caption{Flowchart of the \ac{UDA} approach proposed in \cite{yi2021complete} for semantic segmentation
of \ac{LiDAR} \acp{3DPC} }
\label{UDDA}
\end{figure}

Moving on, in \cite{yi2021complete}, a \ac{UDA} problem approach for
the semantic labeling of \acp{3DPC} is proposed, focusing on domain gap reduction of \ac{LiDAR} sensors. Based on the observation that sparse  \acp{3DPC} are sampled from 3D surfaces, a  \ac{SVCN} was designed to
complete the 3D surfaces of a sparse \ac{3DPC}. Unlike
semantic labels, obtaining training pairs for \ac{SVCN} requires
no manual labeling. A local adversarial
learning to model the surface prior was also introduced. The recovered 3D surfaces serve as a canonical domain from which semantic
labels can transfer across different \ac{LiDAR} sensors. Experiments and ablation studies with  benchmark for
cross-domain semantic labeling of \ac{LiDAR} data show that
this approach provides 6.3-37.6\% better performance
than previous DA methods. Fig. \ref{UDDA} portrays the block diagram of the the \ac{UDA} scheme introduced in \cite{yi2021complete} for semantic segmentation of \ac{LiDAR} \acp{3DPC}.

\color{black}
\subsection{Semi-supervised DTL}
The integration of semi-supervised DTL into 3DPC understanding for tasks like object detection and segmentation is rapidly transforming the field, particularly in contexts where labeled data are scarce. This approach leverages both labeled and unlabeled data, enhancing learning efficiency and model performance across various applications, from autonomous driving to tunnel monitoring.
Typically, Tang et al. \cite{tang2019transferable} develope a semi-supervised 3D object detection model that utilizes a novel network to transfer knowledge from well-labeled object classes to weakly labeled classes, improving detection in classes with only 2D labels.
Moving on, Ji et al. \cite{ji2023semi} propose the Semi-supervised Learning-based Point Cloud Network (SPCNet) which integrates various learning modules to enhance tunnel scene segmentation from 3DPCs, significantly reducing the reliance on extensive labeled datasets.
Imad et al. \cite{imad2021transfer}: Focused on utilizing 3D LiDAR data for autonomous driving perception, enhancing object detection through a semi-supervised learning framework that minimizes the need for large-scale annotated datasets.

In this same direction, Huang et al. \cite{huang2020feature} present a point cloud registration framework that minimizes a feature-metric projection error without needing correspondences, using a semi-supervised approach. This method is particularly robust to noise and density differences in point clouds.
Horache et al. \cite{horache20213d} propose a method for generalizing deep learning for 3DPC registration on entirely different datasets using a combination of Multi-Scale Sparse Voxel Convolution and an unsupervised transfer algorithm, UDGE, enhancing adaptability across varied real-world datasets.
Li et al. \cite{li2021semi} introduce a semi-supervised point cloud segmentation method that utilizes both labeled and unlabeled data. By employing adversarial architecture for confidence discrimination of label predictions on unlabeled point clouds, their approach enhances segmentation performance.

Besides, Chen et al. \cite{chen2021multimodal} propose a multimodal semi-supervised learning framework that utilizes instance-level consistency and a novel multimodal contrastive prototype loss to enforce consistent representations across different 3D data modalities of the same object.
Similarly, Xiao et al. \cite{xiao2022transfer} tackle the synthetic-to-real data gap in 3DPCs segmentation by developing a large-scale synthetic dataset and a novel translation method that decomposes and addresses the differences in appearance and sparsity between synthetic and real datasets.
Moving on, Mei et al. \cite{mei2019semantic} develop a system for the semantic segmentation of 3D LiDAR data in dynamic scenes, using a semi-supervised learning strategy that combines limited manual annotations with large amounts of constraint data to enhance scene adaptability.
Additionally, Huang et al. \cite{huang2021spatio} introduce a spatio-temporal representation learning framework for learning from unlabeled 3DPCs in a self-supervised manner, facilitating the generalization of pre-trained models to a variety of downstream tasks.
Moreover, Qin et al. \cite{qin2020weakly} propose a weakly supervised framework for 3D object detection that leverages unsupervised 3D proposal generation and cross-modal knowledge distillation, reducing the dependency on annotated 3D bounding boxes.

On the other hand, Yu et al. \cite{yu2022data} tackle the data scarcity challenge in 3D tasks by transferring knowledge from robust 2D models to augment RGB-D images with pseudo-labels, significantly improving the pre-training of 3D models with limited labeled data.
Xu et al. \cite{xu2023hierarchical} develop a hierarchical point-based active learning strategy for 3DPC segmentation, which measures uncertainty at multiple levels and selects important points for manual labeling, effectively utilizing limited annotations.
Zhang et al. \cite{zhang2023simple} study weakly semi-supervised 3D object detection with point annotations to generate high-quality pseudo-bounding boxes, enabling 3D detectors to perform comparably to fully-supervised models with substantially fewer labeled data.
Lastly, Wang et al. \cite{wang2023ssda3d} introduce a Semi-Supervised Domain Adaptation method for 3D object detection (SSDA3D) that employs an Inter-domain Point-CutMix module to align point cloud distributions across domains and an Intra-domain Point-MixUp for enhancing model generalization on unlabeled target data.
Table \ref{table:semi-supervised} summarizes and compares the above-discussed studies based on several aspects.  This comparison underscores the dynamic evolution of semi-supervised DTL techniques in handling 3DPCs, which is pivotal for applications ranging from autonomous vehicles to environmental scanning and medical imaging. These advancements promise to drive significant improvements in how 3D data is processed and utilized across various fields.

\begin{table}[t!]
\caption{Comparison of Studies on Semi-Supervised DTL in 3DPCs}
\label{table:semi-supervised}
\centering
\scriptsize
\begin{tabular}{ m{0.6cm}  m{2.5cm}  m{1.8cm}  m{2.5cm}  m{3.4cm}  m{3.4cm} }
\hline
\textbf{Ref.} & \textbf{ML Model} & \textbf{Dataset} & \textbf{Application / Task} & \textbf{Advantage} & \textbf{Limitation} \\
\hline
\cite{tang2019transferable} & Semi-supervised 3D Object Detection & SUN-RGBD, KITTI & 3D object detection from 2D and 3D labels & Efficiently transfers 3D info from strong to weak classes & Requires sufficient data in strong classes for effective transfer \\
\cite{chen2021multimodal} & Multimodal Semi-supervised Learning & ModelNet10, ModelNet40 & 3D classification and retrieval & Improves data efficiency using multimodal data consistency & Performance depends on the quality of multimodal data integration \\
\cite{ji2023semi} & SPCNet & Real tunnel point clouds & Multi-class object segmentation in tunnel scenes & Reduces reliance on labeled data, enhances segmentation performance & Specific to tunnel environments, may not generalize \\
\cite{imad2021transfer} & DTL based Semantic Segmentation & KITTI, Ouster LiDAR-64 & 3D object detection in autonomous driving & Reduces need for large-scale datasets, fast processing & Limited by the initial data quality and transfer efficiency \\
\cite{wang2023ssda3d} & Semi-Supervised Domain Adaptation (SSDA3D) & Waymo, nuScenes & 3D object detection in diverse conditions & Adapts to new domains with minimal labeled data & Performance may degrade with severe domain shifts \\
\cite{xiao2022transfer} & DTL with PCT & SynLiDAR & 3DPC segmentation & Bridges the gap between synthetic and real data, extensive dataset & Focuses on segmentation; may not directly apply to other 3D tasks \\ 

\cite{huang2020feature} & Semi-supervised or unsupervised feature-metric registration & N/A & Point cloud registration & Robust to noise and does not require correspondences; fast optimization & Limited details on specific dataset adaptability and real-world implementation \\
\cite{horache20213d} & MS-SVConv and UDGE & 3DMatch, ETH, TUM & 3DPC registration & Generalizes across different datasets using unsupervised DTL & May require substantial computational resources; specific adaptation challenges not addressed \\
\cite{mei2019semantic} & CNN-based classifier & Custom LiDAR dataset & Semantic segmentation of dynamic scenes & Combines few annotations with large constraint data for improved adaptability & Primarily tailored to dynamic scenes, may not generalize to static or varied environments \\
\cite{huang2021spatio} & Spatio-Temporal Representation Learning (STRL) & Synthetic, indoor, outdoor datasets & 3D scene understanding & Learns from unlabeled data; generalizes to multiple 3D tasks & Dependence on temporal correlation which may not be present in all datasets \\
\cite{yu2022data} & DTL from 2D to 3D models & ScanNet & Semantic segmentation & Uses pseudo-labels for pre-training 3D models; enhances data efficiency & Relies on the quality and relevance of 2D model training to 3D tasks \\
\cite{qin2020weakly} & Weakly supervised 3D object detection & KITTI & 3D object detection & Reduces need for detailed annotations; uses unsupervised proposal generation & Performance may lag behind fully-supervised methods; adaptation to other datasets not detailed \\
\cite{li2021semi} & Semi-supervised point cloud segmentation & N/A & Point cloud segmentation & Utilizes both labeled and unlabeled data; improves with self-training & Specifics on dataset and environmental adaptability are not detailed \\
\cite{xu2023hierarchical} & Hierarchical point-based active learning & S3DIS, ScanNetV2 & Point cloud semantic segmentation & Efficient use of very few labeled data; incorporates active learning & May require intricate setup for uncertainty measurement and point selection \\
\cite{zhang2023simple} & Vision Transformer-based WSS3D & SUN RGBD, KITTI & 3D object detection & Low reliance on fully labeled data; uses point annotations effectively & Challenges with varying detector compatibility and scene diversity \\
\hline

\end{tabular}
\end{table}

\color{black}
\subsection{Inductive transfer learning}
\label{sec:ITL}

In \ac{ITL}, usually \ac{SD} and \ac{TD} remain the same, however target task differs from the source task.  In \ac{ITL} setting, knowledge is transferred from source task to attain high performance in the target task. As suggested in \cite{zhuang2020comprehensive, pan2009survey}, the target is labeled, whereas source may be labeled, unlabeled, or both. However, in the field of \ac{3DPC}, it is observed that only labeled source has been explored. 


\textcolor{black}{For instance, Chen et al. \cite{chen2021classification} focus on classifying soft-story buildings using \acp{CNN} and density features extracted from \acp{3DPC}. More specifically, Once the \ac{3DPC} data of Santa Monica city is collected, density features are extracted and converted into 2D imagery data. The task of identifying a soft-story building is then approached as a binary classification problem, which has effectively been addressed using \ac{CNN} models, including VGG, Inception, ResNet and Naive \ac{CNN}.} They have trained the \ac{SD} with labeled settings exploiting more than
1.2 million images and 1,000 labeled categories, and surprisingly VGGNet has proved to be the best in 2014, with its different versions namely, VGG19 and VGG16. In addition to this, 138 million parameters are trained in the network. VGGNet uses 3 x 3 filters, contrary to this, Inception uses 1 × 1 filters, limiting the number of input channels. Hence, the trainable parameters for Inception is low, approximately 6.4 million trainable
parameters. In an another work, Murtiyoso et al. \cite{murtiyoso2021semantic}, have taken advantage of DeepLabv3+ network which they trained on labeled dataset of building façade images. The trained network is deployed on \ac{TD}, 2D orthoimages received from photogrammetry. Similarly, a \ac{DL}-based encoder-decoder lightweight neural architecture, RandLA-Net, for the semantic segmentation of \ac{3DPC} data, has been utilized in \cite{kim2022deep} for performing per-point segmentation of large-scale \acp{3DPC}, as detailed in \cite{hu2020randla}, and demonstrated a good semantic segmentation performance in the Semantic 3D benchmark  \cite{kim2022deep, hackel2017semantic3d}.

In \cite{zhao2019point}, authors have exploited \ac{ITL} for \ac{3DPC} classification using an airborne laser scanning   method. They have extracted deep features using deep residual network, and \ac{CNN} was dedicated to classification task. To exploit \ac{CNN} further, the unevenly distributed \ac{3DPC} is transformed into voxels \cite{liu2019point}. However, this transformation can lead to information redundancy as well as some feature loss.PointNet and later PointNet++ were proposed in \cite{qi2017pointnet} and \cite{qi2017pointnet++}, respectively. They successfully directly classify the original \ac{3DPC}, leading to the proliferation of PointNet-like approaches for \ac{3DPC} classification. However, classification accuracy is affected when small range of multi-view projection influence the detailed description on each 3D point. In addition, there is a need of huge computing time for generating feature maps.

Lie and his team \cite{lei2020point} have used DensNet201 for obtaining deep features, and make use of an improved fully \ac{CNN} by incorporating it into the \ac{ITL}. The result was very promising and superior to state-of-the-art on ISPRS dataset for \ac{OA}. In many cases weight transfer has found to be instrumental in classification performance. For instance, Kamil and Marian \cite{sidor2020recognition} has suggested weights transfer while they use \ac{ITL} to improve the classification accuracy of \ac{BiLSTM}-based approach for identifying human activities. Similarly, Sun et al. \cite{sun2018deep} has incorporated deep \ac{ITL} to explore the options for solving the issues that are encountered with SAE networks in the prediction of remaining useful life by utilizing weight and feature transfer.

For semantic labeling of heritage, Arnold et al. \cite{arnold2021automatic} has utilized \ac{ITL} by considering geometric shape characteristic for the purpose of segmentation of memorial objects  from the scene. Further, they pretrained \ac{CNN} on a model from ModelNet's labeled dataset. Similarly, for body measurement of cattle, Huang et al. \cite{huang2019body} have pretrained the Kd-network by obtaining 3D deep model's initial parameters from ShapeNet. The authors extracted the \ac{3DPC} spatial feature information which are transferred for identifying cattle body silhouette. In addition, their \ac{ITL} method is applicable even if the data distribution is different for source and target data.

\subsection{Transductive transfer learning }
The \ac{TTL} was introduced by  Arnold et al. \cite{arnold2007comparative} to solve a task in which domains are different however the source and target tasks remain same . The authors advocate that at the training time, all unlabeled data must be available in the \ac{TD}, contrary to this, Pan and Yang \cite{pan2009survey} urges that for finding out marginal probability, it would be enough to seeing part of the unlabeled target data  at the training time.



Applying \ac{DTL} for different domain encounters many issues. One of these issues for \acp{SD} and \acp{TD} is distribution mismatching. There are few works based on \ac{UDA} methods that have been reported in the literature for the adaptation of a model to an unlabeled \ac{TD} from a labeled \ac{SD} for various different applications \cite{bousmalis2017unsupervised, duan2012domain, kang2019contrastive, long2017deep, zhang2018collaborative}. In addition to \ac{UDA}, authors in \cite{tian2021vdm} has suggested \ac{SFUDA}. They have used virtual domain modeling to address of one the most talked issue in \ac{SFUDA} without requiring original source, reducing source and task data mismatching. To fill the gap between \ac{SD} and \ac{TD} distributions, they have introduced an intermediate virtual domain, reducing the distribution mismatch into two steps, minimizing the domain gap between \ac{SD}  and virtual domains and then between the virtual and \acp{TD}. To achieve this goal, the authors have generated the virtual domain samples in the feature space using an approximated Gaussian mixture model (GMM) and the pretrained source model, so that the virtual domain maintains a similar distribution to the \ac{SD} without access to the original source data. On the other hand, they have proposed an effective distribution alignment method that gradually improves the compactness of the \ac{TD} distribution through model learning to reduce the aforementioned distribution gap. However, the first \ac{SFUDA} framework was proposed in \cite{saltori2020sf}, where the authors exploited the source model to fine-tune the \ac{TD} data without access to the source data for 3D object detection. The paper \cite{achituve2021self} discussed the use of \ac{SSL} to learn useful representations from unlabeled data in \ac{DA} for 3D perception problems. The authors proposed a new family of pretext tasks employing \ac{SSL} for \ac{DA} on point clouds, called deformation reconstruction, along with a novel training procedure called \ac{PCM} for labeled point cloud data. The results demonstrated that employing \ac{SSL} with this technique significantly improves classification and segmentation in \ac{DA} datasets.

As the research on DA for \ac{3DPC} is evolving, researchers are working to improve the method to reduce the \ac{SD}' differences from \ac{TD}, and to maximize the domain adaptability so that it can be generalized. However, most of the research addresses \ac{DA} through experimentation over different domain meant for different but related task.  Table \ref{UDA} summarizes some of the relevant studies, compares their characteristics and identifies their pros and cons.

\begin{figure}
\centering
\includegraphics[width=0.98\textwidth]{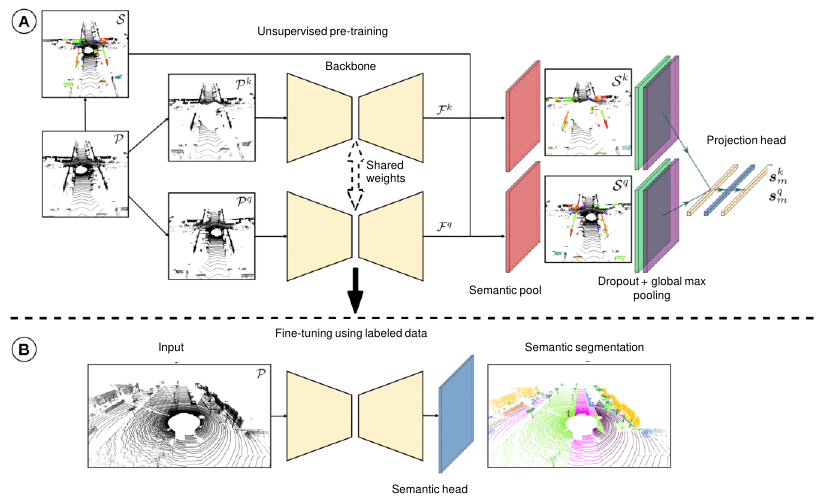}
\caption{\textcolor{black}{Example of using fine-tuning in \ac{3DPC} segmentation: (A) Data augmentation is used to generate the augmented views $\mathcal{P}^{q}$ and $\mathcal{P}^{k}$ from a point cloud $\mathcal{P}$, and class-agnostic segments $S$ are extracted from $\mathcal{P}$. The augmented segments $\mathcal{S}^{q}$ and $\mathcal{S}^{k}$ are determined with their point-wise features using the point indexes of $\mathcal{S}$ extracted from $\mathcal{P}$. Point-wise features $\mathcal{F}^{q}$ and $\mathcal{F}^{k}$ are computed, followed by dropout and global max pooling over each segment. The segment feature vectors are projected using the projection head to obtain the final features $\textbf{s}^{q}_{m}$ and $\textbf{s}^{k}_{m}$ from the $M$ segments, and the contrastive loss is computed, and (B) The pre-trained backbone is fine-tuned for the downstream task, i.e., semantic segmentation \cite{nunes2022segcontrast}.}}

\label{fig:fn2}
\end{figure}

The works \cite{rist2019cross, wang2019range} require handling preprocessing tasks and additional data, which in turn increases complexity. Rist et al. \cite{rist2019cross} exploit cross-sensor \ac{DA} via 3D voxels, while Wang et al. \cite{wang2019range} utilize adversarial networks for global adaptation by exploiting cross-range adaptation. In this context, the authors in \cite{wang2021strong} present a potentially significant contribution. Their approach circumvents complex procedures and the need for additional data. Instead, they leverage the KITTI benchmark dataset to align strong-weak features and enhance feature representation, supplementing it with available data when necessary. Specifically, they focus on the 'car' example from the benchmark dataset, considering 'near-range' and 'far-range' objects as the \ac{SD} and \ac{TD}, respectively. In addition to the underlying issues in handling different domain for DA problems, distribution mismatch within a domain has also been raised by researchers. In this regard, to adapt the different scenario within a single domain, Zhang et al. \cite{zhang2021srdan} have taken the advantage of cross-dataset and considered several adaptation scenario like day-night adaptation and adaptation of different scenes (say, from different places) as in the case with nuScenes dataset consisting of driving scenes from different geographical locations, Boston and Singapore. Their extensive experiments enable understanding how researchers can leverage cross-datasets for \ac{UDA}. Moreover, they have introduced range-aware and scale-aware detection mechanism for \ac{3DPC} tasks. In the similar direction, Jaritz et al. \cite{jaritz2020xmuda} has considered the same scenario, i.e., day-night adaptation and Boston-Singapore scenario, however hey have primarily focused on cross-modality adaptation using their xMUDA model. By the virtue of their proposed method, they have shown a better performance can be achieved for cross-modality. Moreover, A2D2 and Semantic KITTI datasets were used as source and target respectively. 
Moving forward, Nunes et al. \cite{nunes2022segcontrast} present a new contrastive learning approach for representation learning of 3DPs point cloud data in the context of autonomous driving. The approach extracts class-agnostic segments and applies contrastive loss to discriminate between similar and dissimilar structures. The method is applied on data recorded with a 3D \ac{LiDAR} and achieves competitive performance in comparison to other self-supervised contrastive point cloud methods.
Fig. \ref{fig:fn2} presents an example of fine-tuning in \ac{3DPC} segmentation \cite{jaritz2020xmuda}.

A UDA-based semantic labeling scheme of \acp{3DPC} is proposed in \cite{yi2021complete}, where domain discrepancies induced by different \ac{LiDAR} sensors have been addressed. Typically, a complete and label technique that recovers 3D surfaces is developed before passing them to a segmentation network. More precisely, a \ac{SVCN} to
complete the 3D surfaces of a sparse \ac{3DPC} is designed.
By contrast to semantic labeling, obtaining training pairs for \ac{SVCN} does not require any manual labeling.
Moreover, local adversarial learning has been introduced for modeling the surface priors. The recovered 3D surfaces serve as a canonical domain, from which semantic
labels can transfer across different \ac{LiDAR} sensors.

\begin{center}
\scriptsize

\begin{longtable}[!t]{
m{0.5cm}
m{1.5cm}
m{1.5cm}
m{1.5cm}
m{0.5cm}
m{0.5cm}
m{7cm}}
\caption{A summary of unsupervised DA frameworks and their characteristics used in \ac{3DPC} understanding.} 
\label{UDA}\\
\hline
Work & SD & TD & \ac{ML} model & SoDTs  &  SoDDs   & Pros and cons\\ \hline
\endfirsthead

\multicolumn{6}{c}{{Table \thetable\ (Continue)}} \\
\hline
Work & SD & TD & \ac{ML} model & SoDTs &  SoDDs  & Pros and cons  \\\hline 
\endhead

\hline
\endfoot

\cite{achituve2021self}      &    PointSegDA    &     PointSegDA    &    DefRec    &    S    &    D    &   The first study of SSL for DA on \acp{3DPC}.  \\

\cite{sidor2020recognition}      &    UTD-MHAD    &    UTD-MHAD     &   N/A    &    S     &    S    &   Computationally efficient and enable human activity detection using depth maps   \\

\cite{tian2021vdm}      &    ModelNet40    &     ShapeNet    &   VDM-DA    &    S    &   D    &   Enable virtual domain modeling for source data-free DA.  \\

\cite{wu2019squeezesegv2}      &    GTA-LiDAR    &    KITTI    &    SqueezesegV2    &   S    &    D    &   Enhance model structure and UDA for road-object segmentation.  \\

\cite{jiang2021lidarnet}      &    Semantic (KITTI, POSS, USL)    &     Semantic (KITTI, POSS, USL)    &    LiDARNet   &    S    &    D    &  Preserve almost the same performance on the SD after adaptation and achieve 8\%-22\% mIoU performance increase in the TD.  \\

\cite{yi2021complete}      &    Waymo    &     KITTI; \newline nuScenes    &    \acs{SVCN}   &    S    &    D   &   Provide 8.2-36.6\% better performance than previous DA methods.  \\

\cite{qin2019pointdan}      &    pointDA-10    &     pointDA-10    &    pointDAN    &    S    &    S    &   3DPC representation using a multi-scale 3D DA network.  \\

\cite{yang2021st3d}      &    Waymo    &     KITTI, nuSenses, Lyft    &   ST3D    &   S    &   D    &  Exceed fully supervised results on KITTI 3D object detection benchmark.  \\

\cite{saltori2020sf}      &    nuScenes    &     KITTI    &    SF-UDA    &    S    &    D    &   Enable \ac{SFUDA} but requires to be tested beyond cars, to detect other types of objects.  \\

\cite{cardace2021refrec}      &    PointDA-10; ScanObjectNN    &    ShapeNet, ModelNet40, ScanNet, ScanObjectNN    &    RefRec   &    S    &    D    &  Refine pseudolabels, offline and online, by leveraging shape descriptors learned to solve shape reconstruction on both domains  \\

\cite{wang2019range}      &    KITTI    &     nuScenes    &   CrAF  &    S    &    D    &   Conduct  more challenging cross-device adaptation.  \\

\cite{zhang2021srdan}      &    PreSIL; nuScenes    &     KITTI; nuScenes    &    SRDAN    &    S   &    D    &  Demosntrate the significance of geometric characteristics for cross-dataset 3D object detection.  \\

\cite{liu2021adversarial}      &    labeled data, VirtualKITTi    &    Semantic KITTI    &    CLDA   &    D    &    D    &  Improve segmentation performance for rare classes but still needs more enhancement. \\
 
 \cite{tang2021bi}      &    PointDA-10    &     PointDA-10    &    BADM; \newline PointDA-10    &    S    &    S    &   Achieve state-of-the-art performance and outperform other 3D DA techniques but not appropriate for different tasks or different domains.   \\

 \cite{vesal2021adapt}      &    bSSFP-MRI; MM-WHS    &     LGE-MRI; MRI and CT    &    UDA-GAN    &    D     &    D    &   Provide promising performance, compared to the state-of-the-art.  \\

 \cite{alam2021palmar}      &    Video data    &     LiDAR    &    PALMAR    &    S    &    D    &   63\% improvement of multi-person tTracking than state-of-the-art frameworks while maintaining efficient computation on the edge devices.  \\

 \cite{wang2021unsupervisedKITTI}      &    KITTI object; near-range    &     KITTI object; far-range    &    SCNET    &    S    &    S    &   Achieve a remarkable improvement in the adaptation capabilities but without enable knowledge transfer to different tasks.  \\

\cite{qiao2021registration}      &    GTA-V    &     Oxford RobotCar    &    vLPD-Net R; \newline vLPD-Net V    &    S    &    D    &    Achieve state-of-the-art performance on the real-world Oxford RobotCar dataset but the investigation of the loop closure and re-localization in real-world is needed.   \\

\cite{zhu2021automatic}      &    CadData    &     CamData    &    DANN; \newline SSLPC; \newline PoinDAN    &    S    &    D    &  Less time-consuming for implementation in production.  \\

\cite{wang2021cross}      &    PointDA-10    &     PointDA-10    &    DSDAN    &    S    &    S    &  Exceed the state-of-the-art performance of cross-dataset 3DPC recognition tasks.  \\

\cite{alam2020lamar}      &    High resolution LiDar data    &     Low resolution LiDar data    &    LAMAR    &    S    &    D    &   94\% Human activity recognition performance in multiple-inhabitant scenario. \\

\cite{jaritz2022cross}      &    A2D2    &     Semantic KITTI    &    xMUDA    &    S    &    D    &   Enable \ac{CLDA} in 3D semantic segmentation; however, knowledge cannot be transferred across different ttasks. \\

\cite{saleh2019cyclist}      &    KITTI    &     MDLS    &    CycleGAN; \newline YOLOv3   &    S    &   D    &   Outperformed other compared baseline approaches with more than 39\% improvement in F 1 -Measure score.  \\

\end{longtable}
\begin{flushleft}
Abbreviations: Same or different tasks (SoDTs); Same or different domains (SoDDs); Different (D); Same(S).
\end{flushleft}


\end{center}


\section{Applications of TL-based 3DPCs} \label{sec4}
TL-based \ac{3DPC} can be applied in various applications, including robotics and autonomous systems, augmented reality and virtual reality, medical imaging, geo-spatial data analysis, industrial inspection, building information modeling, etc. In this section, we focus on some of the main applications that attract increasing research effort. \textcolor{black}{Table \ref{3dpc_tasks} summarizes some of the pertinent \ac{DTL}-based \ac{3DPC} frameworks proposed to perform different tasks.
}


\subsection{Semantic labeling and segmentation} 
Semantic labeling of \ac{3DPC} is crucial for performing segmentation tasks to represent scans accurately, using manual as well as automatic labeling. For instance, recently, Xie et al. \cite{xie2020pointcontrast} have used the Stanford large-scale 3D indoor spaces (S3DIS) dataset for transferring knowledge to the target dataset. However, it contains a much smaller repository as compared to source dataset (ScanNet), and manual semantic labeling has been done with 13 categories. Arnold et al. \cite{arnold2021automatic} used simple \ac{MLP}  to learn the diversity of the feature space and acquire  knowledge for semantic labeling. 

\textcolor{black}{One of the prime concerns is to identify similar labels and different data from the same labels separately. To address this, authors \cite{pham2016semantic} have suggested a domain-independent approach for semantic labeling. Their proposed model considers similarity metrics as features to infer the correct semantic labels and learns a similarity function to identify if attributes have the same labels. Because the matching function is unrelated to specific labels, their model does not depend upon the label and thus are independent of domain ontologies. A similar approach for bridge \ac{3DPC} has been suggested in \cite{xia2022automated}. The authors have received an improved IoU with 94.29\%.}

\textcolor{black}{ The ability of \ac{DTL} to reduce the need for large-scale dataset can be economical and hence can be exploited in \ac{3DPC} tasks for performing accurately without requiring more extensive data. For example, with the help of \ac{DTL}, authors in \cite{sun2019not} have explored the analogy between real and synthetic data for semantic segmentation and tried reducing their gap. Based on deep \ac{CNN} (DCNN), authors in  \cite{hong2016learning} have suggested a weakly-supervised semantic segmentation technique. Unlike others, they have considered auxiliary segmentation annotations for distinct categories to guide segmentation on pictures with only image-level class labels. They have used a decoupled encoder-decoder architecture with an attention model allowing segmentation knowledge to be transferable across categories.}

\begin{figure}[t!]
\begin{center}
\includegraphics[width=1\textwidth]{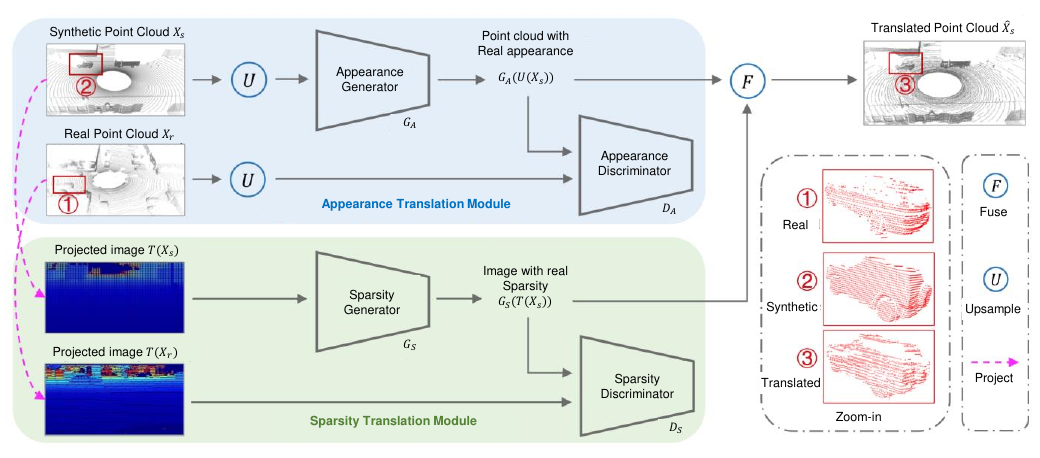}\\
\end{center}
\caption{The PCT method separates the task of translating \ac{3DPC} into two parts: appearance translation and sparsity translation. Using synthetic \ac{3DPC} as input, the appearance translation learns to create dense \ac{3DPC} that look similar to real ones. The sparsity translation then learns the typical sparsity patterns found in real 2D point clouds, and combines these patterns with the dense \ac{3DPC} in 3D space. The end result is a \ac{3DPC} that looks and is sparse similarly to real \ac{3DPC}.}
\label{PCT_exp}
\end{figure}

Semantic segmentation is conducted in \cite{xiao2022transfer} by transferring knowledge from synthetic to real \ac{LiDAR} \acp{3DPC}. Typically, a large-scale synthetic \ac{LiDAR} dataset,  is first collected from multiple virtual environments with rich layouts and scenes. It includes point-wise labeled \acp{3DPC} with accurate geometric shapes and comprehensive semantic classes. 
Moving on, a \ac{3DPC} translator (PCT) is designed to mitigate the discrepancy between real and synthetic \acp{3DPC}. Typically, the synthetic-to-real discrepancy is decomposed into a sparsity and an appearance component before separately handling them. Fig. \ref{PCT_exp} explains the PCT approach adopted in \cite{xiao2022transfer}, disentangling \ac{3DPC} translation into appearance and sparsity translation tasks. Accordingly, dense \acp{3DPC} having similar appearances are first learned within the appearance translation.
The sparsity translation then learns real sparsity distribution in 2D space and fuses it with the reconstructed
\ac{3DPC} in 3D space. Then, real sparsity distributions in 2D space are learned within the sparsity translation before being fused with the reconstructed \ac{3DPC} in 3D space.

\textcolor{black}{Moving forward, A. Murtiyoso et al. \cite{murtiyoso2021semantic} has used \ac{DTL} on a photogrammetric orthoimage by exploiting labeled and rectified images of building facades for training neural networks on them. Next, they allow the transition from 2D orthoimage to \ac{3DPC} by another program. With their promising results for photogrammetric data, their proposed work is a potential option to assist in automatic \ac{3DPC} semantic segmentation. Similarly, authors in \cite{matrone2021transfer} have claimed that semantic segmentation can perform equally competently to the latest \ac{ML} approaches in modeling cultural heritage.}

\textcolor{black}{On the other hand, \ac{3DPC} instance segmentation is equally employed for many applications. For example, in \cite{zong2022improved}, authors have taken a problem of height detection of a catenary conductor. For \ac{3DPC} instance segmentation, they have incorporated \ac{DTL} and 3D-BoNet model with a multiscale grouping (MSG) structure on a smaller dataset, with loss curve converging better for their model.}
In addition to its applications in building, tunneling, and construction, \ac{DTL} for \ac{3DPC} segmentation has found utility in robotics. Specifically, it is utilized for automatic toolpath generation to enable tasks such as masking, deburring, and polishing in various industrial applications. Z. Xie and his team \cite{xie2021automatic} have utilized \ac{3DPC} segmentation and classification to extract the object features. Furthermore, they have exploited \ac{DTL} to enhance performance and avoid investing much time in training data. Their toolpath recommendation automatically suggests patterns to choose from and does not require manual efforts. To identify human activities, authors \cite{sidor2020recognition} have carried out segmentation to separate the human figure from the background so that the descriptor can only be determined for a human figure.

\subsection{Classification} 
\begin{figure}[t!]
\begin{center}
\includegraphics[width=1\textwidth]{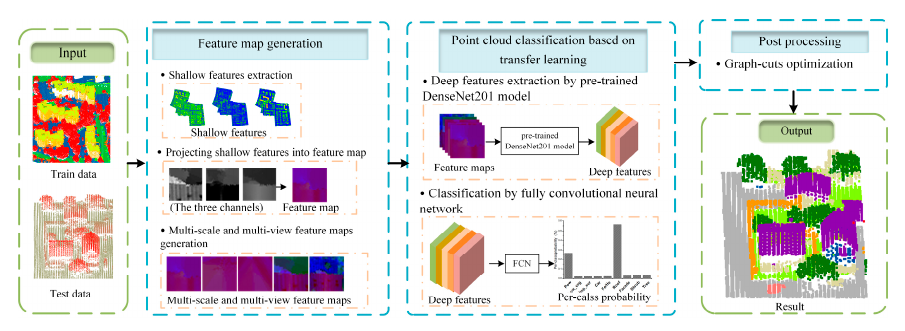}\\
\end{center}
\caption{The flow chart of the proposed classification method.}
\label{PC_classification}
\end{figure}

\ac{3DPC} classification is the process of classifying the features of 3D objects and categorizing the classes to which they belong. The \ac{3DPC} segmentation types and classification are illustrated in Fig. \ref{PC_classification}. In the diagram, it is shown how a building, with the help of \ac{CNN}, can be classified based on different classifications e.g., concerning time (period), region, and structure. Recently, researchers have explored different ways of directly classifying \ac{3DPC}, mainly to avoid information loss during the conversion of \ac{3DPC} into non-\ac{3DPC}.  Automated classification of point cloud data has great potential for various applications, but a limited number of labelled points can lead to overfitting and poor generalization in \ac{ML} models.

One of the pioneered works in the field is done by Qi et al. \cite{qi2017pointnet}, reported being the first \ac{DL} model to classify the original \ac{3DPC} directly. Their PointNet model is based on \ac{CNN}, and its success has attracted many works to follow their proposed architecture. \ac{CNN} has been widely used for classification tasks. For example, authors \cite{murtiyoso2021semantic} have used \ac{CNN} with \ac{DTL} to classify improved fisher vectors used to represent \ac{3DPC}. The work \cite{arief2019addressing}  introduces a method called Atrous XCRF that induces controlled noise by invoking conditional random field similarity penalties using nearby features to improve generalization. The method achieves high \ac{OA} and F1 score in a benchmark study using the ISPRS 3D labeling dataset and is on par with the current best model. Using \ac{DTL} with Bergen 2018 dataset  shows improvement in accuracy, but more work is needed to address generalization issues in \ac{DA} and \ac{DTL}. \textcolor{black}{To overcome the requirement of large dataset, Lei et al. \cite{lei2020point} have proposed a \ac{3DPC} classification method that integrates an improved \ac{CNN} into \ac{DTL}. They have used the DensNet201 model as a pretrained model for finding deep features able to accurately characterize the object. In addition, The authors \cite{zhao2019point} have proposed an ALS \ac{3DPC} classification method based on \ac{DTL} using \ac{CNN}. } In \cite{lei2020point}, an ALS \ac{3DPC} classification approach for integrating an enhanced fully-\ac{CNN} into \ac{DTL} with multi-view and multiscale and deep features is proposed. Typically, the shallow features of the ALS \ac{3DPC}, such as height, intensity, and curvature change, are extracted to generate feature maps by multiscale voxel and multi-view projection. 
\textcolor{black}{Second, these feature maps are fed into the pretrained DenseNet201 model to derive deep features. Experimental results show that \ac{OA} and the average F1 scores obtained by the proposed method are 89.84\% and 83.62\%, respectively.}

\subsection{Point-set registration}
With the increasing number of 3D sensors and 3D data production, consolidating overlapping \acp{3DPC}, which is called registration, has become a significant issue in many applications \cite{yang2020teaser}. Registration can be used online (e.g., in a \ac{LiDAR} SLAM pipeline for loop closure detection) or offline (e.g., in 3D reconstructions of RGB-D indoor scenes \cite{choi2015robust} or for outdoor \ac{LiDAR} map building for autonomous vehicles).
However, \ac{3DPC} registration can be challenging with real-world scans because of noise, incomplete 3D scans, outliers, etc. There are numerous existing approaches to this issue; however, recently, \ac{DL} approaches have become very popular, especially for learning descriptors and computing transformations. These data-driven approaches, especially those that involve learning on large datasets with the ground truth pose as supervision, have effectively solved registration problems \cite{wang2019deep,qiao2021registration}. Specifically, in \cite{qiao2021registration}, a registration-aided DA model for \ac{3DPC}-based place recognition is developed, \textcolor{black}{called vLPD-Net. The method  utilizes virtual large-scale point cloud descriptors and a structure-aware registration network, along with adversarial training, to minimize the gap between synthetic and real-world domains.}

\subsection{Scene understanding}

The method proposed by Xu et al. \cite{xu2021image2point} uses  \ac{DTL} technique by inflating weights to transform an image-pretrained model into a \ac{3DPC} model. Authors in \cite{chen2021shape} have proposed a model 'Shape Self-Correction' for \ac{3DPC} analysis where they have explored how unlabeled data can be utilized for a better \ac{3DPC} understanding. In addition, their approach can identify distorted points and automatically correct them by producing state-of-the-art results. Moreover, this model can be used as a pretrained model to be fine-tuned for related tasks.  Going forward, Zhang et al. \cite{zhang2021pointclip} also tried to exploit a pretrained large-scale 2D dataset to generalize it to \ac{3DPC} understanding using a \ac{CLIP}. Their model, PointCLIP, projects \ac{3DPC} into multi-view depth maps by encoding it, and successfully, view-wise zeroshot prediction is aggregated to transfer knowledge from 2D to 3D. Furthermore, many works exploit unsupervised approaches for \ac{3DPC} understanding. For example, Xie et al. \cite{xie2020pointcontrast} have suggested an unsupervised pretraining framework. To avoid the domain gap and utilize their approach across the different domains in the real world, they have directly made use of a complex network to pretrain the model. The major contribution they have is the examination of the transferability of learned representations generalization across the domain for \ac{3DPC} understanding. In addition, their architecture can also represent point-level information by capturing local features. To further exploit the geometric structure of the \ac{3DPC} and relationships of local features, Liu et al. \cite{liu2022fg} have suggested a convolution-based geometric-aware approach. Moreover, they proposed a filtering mechanism that identifies noise and outliers for a high-level understanding.

\subsection{Object detection and denoising}
The exploitation of \ac{DTL} for various applications of \ac{3DPC} in 3D object detection and recognition has recently attracted researchers, ranging from medical science to biological research and robotics. For instance, authors in \cite{benhabiles2018transfer} have used \ac{DTL} for protein structure indexing using \ac{3DPC}. The suggested approach saves training time by fine-tuning the  pretrained network, using Euclidean distance to extract shape-matching,  with generic 3D objects. Furthermore, authors in \cite{zong2022improved} have proposed a detection method incorporating instance segmentation and \ac{DTL} for overhead catenary height detection of a tunnel using a 3D \ac{3DPC} tunnel dataset. To recognize human activities, Sidor et al. \cite{sidor2020recognition} have suggested an approach by transforming depth maps into \ac{3DPC}. The authors have used \ac{DTL} combined with multiple networks to improve the classification mechanism based on BiLSTM. In \cite{tian2021vdm}, \ac{3DPC} semantic segmentation based on \ac{DTL} is used for 3D object detection. The advantage of this method is that it reduces large-scale training datasets requirement, and \textcolor{black}{ consequently, the training duration is reduced as \ac{DTL} exploits knowledge from prior classification tasks, aiding object detection after preprocessing, thereby promoting cross-task generalization.}

Denoising is an important \ac{3DPC} task to remove noise usually found in acquiring images from scanning devices. These noises can influence many downstream tasks such as segmentation, detection, reconstruction and classification, etc. To remove these noises at different scales of \ac{3DPC} data, several algorithms have been suggested. The first \ac{DTL}-based \ac{3DPC} denoising scheme was proposed in \cite{roveri2018pointpronets}. It involved denoising patches of points by projecting them onto a learned local frame and repositioning the points onto the surface using a \ac{CNN}. Some interesting \ac{3DPC} denoising approaches include score-based \ac{3DPC} denoising \cite{luo2021score}, graph Laplacian regularization \cite{zeng20193d}, bipartite graph approximation and total variation \cite{dinesh2018fast}, weighted multi-projection \cite{duan2018weighted}, and feature graph learning \cite{hu2020feature}, etc. 




\textcolor{black}{
\subsection{Upsampling and donwsampling}
%
%
Point clouds obtained through 3D scanning are often incomplete, unevenly distributed, and corrupted by noise. Upsampling, on the other hand, involves generating a dense set of points that not only restores uniformity and proximity to the surface but can also address small gaps in the data, all through a single network. For instance, authors have proposed Segnet \cite{badrinarayanan2017segnet}, and they designed decoder to upsamples their input feature map(s) having lower resolution enabling non-linear upsampling to produce dense feature maps. In a similar way, Eltner et al. \cite{eltner2021using} have used an encoder that extracts a low-resolution activation map while the decoder upsamples it to obtain a pixel-wise classification. More recently, an upsampling method for \ac{3DPC} to reduce human annotation cost is presented in \cite{zhang2022upsampling}. They relied solely on the upsampling operation to perform effective feature learning of \acp{3DPC}. 
Simialrly, Imad et al. \cite{imad2021transfer} have added more convolutional layers to the decoder stage, which has been coupled with upsampling layers to increase the size of the spatial tensor and generate high-resolution segmentation outputs.}

\textcolor{black}{
Down-sampling point clouds refers to the process of reducing the number of points contained within them. This is commonly done to decrease the processing time required, or to select a specific number of points for use in training, among other purposes.Moreover, downsampling is also used to transform \ac{3DPC} into 2D representations. Typically, in \cite{garrote20173d}, a 2D scene modelling method is described that effectively and meaningfully transforms the 3D data to a 2.5D representation and then to a 2D grid map. In addition to this, downsampling has been used in 3D data processing to remove the noise. For example, \cite{orts2013point} have used  growing neural gas (GNG) based downsampling technique for noise removal. However, downsampling necessitates careful consideration of whether the points are significant for the output, so that all important points are passed to the next layer and not removed. To this end, critical points layer (CPL) has been proposed in \cite{nezhadarya2020adaptive}, an adaptive downsampling layer that learns to downsize an unordered \ac{3DPC} while keeping the crucial (critical) points. Downsampling has also been used in building and automation, for instance, a downsampling technique to optimize Terrestrial laser scanning (TLS) datset is suggested in \cite{suchocki2019down}.}

\begin{center}
\scriptsize
\begin{longtable}[!t]{
m{0.5cm}
m{1.5cm}
m{1.5cm}
m{0.5cm}
m{5cm}
m{5cm}}
\caption{Summary of \ac{DTL}-based \ac{3DPC} frameworks describing different tasks.}
\label{3dpc_tasks}\\
\hline
Ref. & MoN & Task/application & DTL  & Limitations & Contribution/advantage/key point   \\ \hline
\endfirsthead

\multicolumn{6}{c}{{Table \thetable\ (Continue)}} \\
\hline
Ref. & MoN & Task/application & DTL  & Limitations & Contribution/advantage/key point  \\\hline 
\endhead

\hline
\endfoot

\cite{xie2020pointcontrast}  & ResNet-34, Pointcontrast & \ac{3DPC} understanding & \textcolor{black}{FT} & \textcolor{black}{Testing for generalizability across multiple datasets and verifying computational costs after applying DTL were not conducted.} &  transferability of learned representation in \acp{3DPC} to high-level scene understanding.  \\

\cite{zhang2020unsupervised} & UFF  & Segmentation and classification  & UDTL & they optimize the classifiers in a single stage, a multi-stage classifier that can reduce the redefined cost function could enhance the result. &  Propose a learning system with an unsupervised feedforward feature (UFF) by incorporating encoder-decoder architecture.  \\

\cite{murtiyoso2021semantic} & DeepLabv3+ & Semantic segmentation & ITL & Due to the terrestrial nature of the data acquisition, several parts of the orthophoto, notably blind spots were distorted by the orthorectification algorithm & An improved automated procedure for \ac{3DPC} semantic segmentation, especially for photogrammetric data.   \\

\cite{imad2021transfer} & \textcolor{black}{AE} &  \ac{3DPC} semantic segmentation for 3D object detection & ITL  & The method does not work for raw \acp{3DPC}. & A \ac{3DPC} is projected into a birds-eye-view by which the amount of annotated data and time required for training has been minimized  \\

\cite{zong2022improved} & 3D BoNet + MSG structure & \ac{3DPC} instance segmentation & ITL &  The proposed RKT method is costly &  New tunnel dataset is generated for \ac{3DPC} segmentation.  \\

\cite{arnold2021automatic} & ConvPoint network, MLP & \ac{3DPC} segmentation for classification & ITL & Semantic interpretation of objects with fewer visual features is missing &  automation of extraction and labeling of memorial objects from cultural heritage sites using \ac{3DPC} data.  \\

\cite{lei2020point} & DenseNet201 & \ac{3DPC} classification & TTL  & Algorithms are not integrated efficiently, which makes the feature maps generation complex & This algorithm can quickly and accurately extract the features and reduce the effect of ground object size on classification results.   \\

\cite{kim2022deep} & RandLA-Net & \ac{3DPC} semantic segmentation & ITL & The SLAM algorithm is not efficient for indoor spaces, \ac{LiDAR} can be located to a limited height only, which provides a minimal view for acquired data &  Robot dog has been exploited for data acquisition instead of human which saves time and minimize the monetary efforts.  \\
    
\cite{chen2021classification}  & \textcolor{black}{CNN} & Classification & ITL & The \ac{3DPC} data is used instead of street-view images, which needs open space. Buildings with irregularities and reinforcing shapes can misclassify as soft-story buildings. &   To accurately classify the soft-story building using CNN model pretrained with DTL  \\

\cite{sidor2020recognition} & \textcolor{black}{BiLSTM}  &  Segmentation and classification  & TTL & eigenvalue-based descriptors for human activity recognition and other state-of-the-art methods are not discussed &  With the help of \acp{3DPC} and VFH descriptor, human activities are recognized.  \\
     
\cite{tian2021vdm} & SFUDMA  & Object recognition  & UDA & \textcolor{black}{The proposed Virtual Domain Modeling needs to be deployed for a variety of tasks.} &  They successfully achieve the goal of distribution alignment between the SD and TD by training deep networks without accessing the SD data  \\

\cite{huang2020superb} & DenesNet-121  & Depth estimation  & \textcolor{black}{ITL} & \textcolor{black}{Testing for generalizability across multiple datasets was not performed. }&  Depth and surface estimation with a higher resolution  \\

\cite{benhabiles2018transfer}  & GLT4IP, pointNet & Protein structure indexing & \textcolor{black}{FT} & \textcolor{black}{The validation of real-time performance improvements through transfer learning was not conducted.} &  DTL based indexing of protein structures using \acp{3DPC}  \\

\cite{ozyoruk2021endoslam} & Endo-SfMLearner & Depth estimation & UTL & \textcolor{black}{Unresolved aspects: enhancing data adaptability, resolving issues, integrating with segmentation, abnormality detection, and classification tasks for improved performance.} & Presented a new dataset for endoscopy, “EndoSLAM”   \\
      
\cite{eckart2021self} & GMM  & Segmentation and classification  & ITL & \textcolor{black}{Further research into novel pretext tasks tailored specifically to the idiosyncrasies of 3D data has not been conducted.}&  For representation learning, the method does not require any procedure like transformation/data augmentation.  \\

\cite{lee2021progressive} & POCO   & 3D orientation, and classification  & ITL & Orientation accuracy can be improved further &   Experimentally shown how DTL can be used for a model to serve as a platform for 3D object’s orientation representation.  \\

\cite{dai2018connecting} & YOLO & Object detection & ITL & Method heavily relies on \ac{LiDAR} data only and may not be that beneficial for another domain of image dataset &  Suggest a method efficient enough and compatible for any \acp{LiDAR}, even containing a different number of channels. Moreover, they do not require a re-training step.  \\

\cite{diraco2021remaining} & GoogleNet, AlexNet, VGG  & Classification  & ITL & verification of the degree of dependence of the DNN architecture on the characteristics of the mechanical system subject to degradation &  representation of punch deformation with depth and normal vector maps (DNVMs) obtained from 3D scan \acp{3DPC}  \\


\cite{huang2019body} & Kd-network & Shape classification, feature recognition & ITL & This method not applicable for any cattle and may produce heavy mistakes and significant errors for adult Qinghai yaks &  Livestock can be observed without any physical involvement of human  \\


\cite{badrinarayanan2017segnet}& Segnet &  Upsampling  & UDTL & 
This method needs to separate the roles between the optimizer and the model to achieve the desired outcome. &  It is much smaller and can be trained end-to-end using stochastic gradient descent. It has been engineered to be effective for memory and processing time during inference.  \\

\cite{zhang2022upsampling} & UAE &  upsampling  & SSTL & \textcolor{black}{The complexity increased with the double-head  architecture. Additionally, mechanical systems with different geometry and material from the punch tool were not investigated.} &  It allows classification and segmentation tasks to benefit from pretrained models. And it reduces the human annotation cost.  \\

\cite{nezhadarya2020adaptive} & CPL &  downsampling  & \textcolor{black}{DA} &  \textcolor{black}{The complexity of CP-Net is high, potentially leading to increased computation costs.} &  an adaptive downsampling layer that learns to downsize an unordered \ac{3DPC} while keeping the crucial (critical) points  \\

\cite{roveri2018pointpronets} & PointProNets &  denoising  & S &  The \ac{3DPC} data utilized in the method was created by adding random noise; hence it is ineffective for \acp{3DPC} that were legitimately created using \ac{LiDAR}. &  The first TL based \ac{3DPC} denoising scheme.  \\


\end{longtable}
\begin{flushleft}
Abbreviations: Unsupervised TL (UTL); supervised DTL (STL); self-
supervised DTL (SSTL); supervised (S); upsampling auto-encoder (UAE); model or network (MoN); fine-tuning (FT).
\end{flushleft}


\end{center}

\subsection{Scene generation}
3D point cloud scene generation or reconstruction is of paramount importance in various fields, including robotics, autonomous driving, virtual reality, and architecture. By capturing the precise three-dimensional coordinates of a scene, 3D point clouds provide a detailed and accurate representation of the environment. This capability is crucial for tasks such as object recognition, navigation, and interaction within a space, enabling robots and autonomous vehicles to perceive and understand their surroundings effectively. In architecture and construction, 3D reconstruction facilitates accurate modeling and analysis of structures, aiding in design, inspection, and renovation processes. Furthermore, in virtual reality and gaming, 3D point cloud generation enhances the realism and immersion of digital environments, providing users with a more engaging and interactive experience. The ability to create detailed and accurate 3D models from point clouds revolutionizes various industries by improving precision, efficiency, and safety in numerous applications.

The SGFormer study \cite{lv2024sgformer} introduces a Semantic Graph Transformer for point cloud-based 3D scene graph generation, addressing the limitations of graph convolutional networks (GCNs) by using Transformer layers for global information passing. The model includes graph embedding and semantic injection layers to enhance object visual features with linguistic knowledge from large-scale language models. Benchmarked on the 3DSSG dataset, SGFormer demonstrates significant improvements in relationship prediction over state-of-the-art methods. The Sat2Scene study \cite{li2024sat2scene} proposes a novel architecture for direct 3D scene generation from satellite imagery using diffusion models and neural rendering techniques. This method generates texture colors at the point level and transforms them into a scene representation for rendering arbitrary views, showing proficiency in generating photo-realistic street-view image sequences and cross-view urban scenes, validated through experiments on city-scale datasets.

The ART3D study \cite{li2024art3d} introduces a framework combining diffusion models and 3D Gaussian splatting for 3D artistic scene generation, utilizing depth information and initial artistic images to generate point cloud maps and enhance 3D scene consistency. ART3D shows superior performance in content and structural consistency metrics compared to existing methods. Liu et al. \cite{liu20233d} propose a high-precision 3D building model generation method using multi-source 3D data fusion, achieving state-of-the-art performance in multi-source 3D data quality evaluation and large-scale, high-precision building model generation. Chung et al. \cite{chung2023luciddreamer} present LucidDreamer, a domain-free 3D scene generation pipeline leveraging large-scale diffusion-based generative models, producing highly detailed Gaussian splats and outperforming previous methods in reconstruction quality. The S2HGrasp study \cite{wang2024single} explores generating human grasps from single-view scene point clouds, addressing challenges of incomplete object point clouds and scene points, showcasing strong generalization capabilities and effective grasp generation. Lastly, the SGRec3D study \cite{koch2024sgrec3d} presents a self-supervised pre-training method for 3D scene graph prediction, achieving significant improvements in 3D scene graph prediction with reduced labeled data requirements during fine-tuning, demonstrating state-of-the-art performance.
\textcolor{black}{
In \cite{liu2023pyramid}, Liu et al. introduce the Pyramid Discrete Diffusion model (PDD), a framework employing scale-varied diffusion models to generate high-quality large-scale 3D scenes. Using a coarse-to-fine paradigm, PDD addresses the complexity and size challenges of 3D scenery data, particularly for outdoor scenes. The model demonstrates effective generation capabilities and data compatibility, allowing easy fine-tuning across different datasets.}

Table \ref{scene-generation} compares various studies on 3D scene generation and reconstruction, highlighting key aspects such as models used, datasets, contributions, performance values, and limitations. The models range from SGFormer, utilizing a Semantic Graph Transformer, to MS3DQE-Net for multi-source 3D data fusion, and Pyramid Discrete Diffusion (PDD) for large-scale scene generation. Datasets include 3DSSG, city-scale datasets from satellite imagery, artistic images with depth information, and MLS 3D point clouds.

Each study offers unique contributions: SGFormer excels in relationship prediction, Sat2Scene integrates diffusion models for urban scenes, and ART3D bridges artistic and realistic images. MS3DQE-Net focuses on high-precision building models, LucidDreamer achieves domain-free scene generation, S2HGrasp generates human grasps from incomplete point clouds, and SGRec3D enhances 3D scene graph prediction. Performance metrics show significant improvements, such as 40.94\% R@50 for SGFormer, FID of 71.98 for Sat2Scene, and PSNR of 34.24 for LucidDreamer. Limitations include over-smoothing in GCNs, handling significant view changes, and managing the complexity of 3D data. These challenges highlight areas for future research in 3D scene generation and reconstruction.

\begin{table*}[t]
\centering
\caption{Comparison of 3D Scene Generation/Reconstruction Studies}
\label{scene-generation}
\scriptsize
\begin{tabular}{p{0.5cm}|p{2cm}|p{2.5cm}|p{3cm}|p{3cm}|p{3cm}}
\hline
\textbf{Ref.} & \textbf{Model(s) Used} & \textbf{Dataset/Data Type} & \textbf{Main Contribution} & \textbf{Best Performance Value} & \textbf{Limitation} \\  \hline
\cite{lv2024sgformer} & SGFormer & 3DSSG & Semantic Graph TransFormer for 3D scene graph generation & 40.94\% R@50, 88.36\% boost in complex scenes & Over-smoothing in GCNs \\ 
\cite{li2024sat2scene} & Diffusion Models, Neural Rendering & City-scale datasets, satellite imagery & Direct 3D scene generation from satellite imagery & FID: 71.98, KID: 5.91 & Handling significant view changes and scene scale \\ 
\cite{li2024art3d} & Diffusion Models, 3D Gaussian Splatting & Artistic images, depth information & 3D artistic scene generation & PSNR: 24.041, SSIM: 0.863, LPIPS: 0.214 & Bridging artistic and realistic images \\ 
\cite{liu20233d} & MS3DQE-Net, Deep Learning & MLS 3D point clouds, 3D mesh data & High-precision 3D building model generation & State-of-the-art performance in data quality evaluation & Balancing local accuracy and overall integrity \\ 
\cite{chung2023luciddreamer} & Diffusion-based Generative Model & Various datasets, VR content & Domain-free 3D scene generation & PSNR: 34.24, SSIM: 0.9781, LPIPS: 0.0164 & Limited by training on 3D scan datasets \\ 
\cite{wang2024single} & S2HGrasp (Global Perception, DiffuGrasp) & S2HGD dataset, single-view scene point clouds & Generating human grasps from single-view point clouds & Contact Ratio: 99.41\%, Volume: 6.58cm\textsuperscript{3} & Handling incomplete object point clouds \\ 
\cite{koch2024sgrec3d} & SGRec3D & Various 3D scene understanding datasets & Self-supervised pre-training for 3D scene graph prediction & +10\% on object prediction, +4\% on relationship prediction & Requires object-level and relationship labels \\ 
\textcolor{black}{\cite{liu2023pyramid}} & \textcolor{black}{Pyramid Discrete Diffusion (PDD)} & \textcolor{black}{Large-scale 3D scenes, outdoor scenes} & \textcolor{black}{Coarse-to-fine 3D scene generation} & \textcolor{black}{mIoU: 68.0, MA: 85.7, F3D: 0.20} & \textcolor{black}{Complexity and size of 3D scenery data} \\ \hline
\end{tabular}
\end{table*}

\section{Open Challenges} \label{sec5}
\textcolor{black}{
\ac{DTL} and DA can be effective techniques for \ac{3DPC} understanding, but they also come with their own set of challenges. Typically, \ac{DL} models require large amounts of labeled data to achieve high accuracy. However, in the case of \ac{3DPC} understanding, it can be challenging to obtain large amounts of labeled data due to the time-consuming and expensive process of manually annotating point clouds \cite{mathes2023we}. Additionally, \ac{3DPC} data can be highly variable in terms of size, shape, and orientation. This variability can make it challenging to develop \ac{DTL} and DA models that can generalize well to new datasets \cite{zhang2023plot,himeur2022next}.
Moreover, DA is needed when the distribution of the source and \acp{TD} is different. In the context of \ac{3DPC} understanding, domain shift can occur due to changes in sensor modalities, lighting conditions, and other environmental factors. To that end, DA techniques must be used to adapt the model to the \ac{TD} and prevent performance degradation.
Moving on, \ac{3DPC} data can contain complex geometric structures, such as non-uniformly distributed points, non-planar surfaces, and varying levels of detail. These complexities can make it challenging to develop effective \ac{DL} models that can accurately capture the structure and features of the data \cite{mathes2023we}. 
\ac{DTL} and DA models can be computationally expensive, especially when dealing with large \ac{3DPC} datasets. This can limit their practical applicability and require specialized hardware to achieve acceptable performance \cite{romanengo2023recognising,himeur2023face}.}

DA methods have shown significant improvements in various \ac{ML} and \ac{CV} tasks, such as classification, detection, and segmentation. However, to date, there are only a few methods that have been successful in applying DA directly to \ac{3DPC} data. The unique challenge in working with \ac{3DPC} data is its large amount of spatial geometric information, and the object's semantics which depend on the regional geometric structures. This means that general-purpose DA methods, which focus on global feature alignment and neglect local geometric information, are not effective for aligning 3D domains, as stated in \cite{qin2019pointdan}.

\subsection{The problem of negative transfer}

Negative transfer refers to a situation when knowledge transfer can negatively influence the model's accuracy, possibly because the \ac{SD} and \ac{TD} are significantly different or they are supposed to perform different tasks \cite{pan2009survey}. Moreover, not taking adequate advantage of the relationship between the \ac{SD} and \ac{TD} can also lead to negative transfer                     \cite{diraco2021remaining}. In such situations, random and forced knowledge transfer without identifying the specific knowledge that can be useful and without considering "what", "when" and "how" to transfer can further propel negative transfer \cite{ribani2019survey}. Hence, it can be inferred that high affinity among source and target tasks is of great importance for a successful \ac{DTL} and to avoid the negative transfer.  To this end, researchers have suggested clustering methods for tasks and identifying similar tasks by keeping them in the same cluster. In addition to this, an approach to group learning tasks together is suggested in \cite{argyriou2008algorithm}. This approach enables what and when to transfer by identifying tasks within a group. Similarly, imbalanced training samples can increase negative transfer chances in a \ac{UDA} scenario. For Example, for \ac{3DPC} representation, Qin et al. \cite{qin2019pointdan} have suggested a 3D DA network and have argued that for domain alignment, unique parts can induce negative transfer and hence covering standard features in 3D space can help to avoid the negative transfer. Furthermore, to address negative transfer while dealing with different domains/modalities for \ac{3DPC}, Gong et al. \cite{gong2021mdalu} have suggested a multi-\ac{SD} adaptation and label unification approach. They have used attention-guided adversarial alignment for a strong distribution alignment between the \ac{SD} and \ac{TD}. Additionally, their proposed uncertainty
maximization module limits the self-assured predictions for unlabeled samples in the \acp{SD}. To make their approach more promising, they have introduced a fusion module based on pseudo-labeling. Especially for unlabeled samples, they perform pseudo-labeling for the \acp{SD} and all samples in the \ac{TD}.

\subsection{The problem of overfitting}

The established successful model, such as PointCNN and other DNN models, despite their widespread use and success for \ac{3DPC} tasks, still face some potential challenges, especially when the dataset available is relatively small. One possible reason for this is significant parameters, even in several million. It requires a large amount of data to fit this many parameters adequately. Since \ac{3DPC} data suffer from the scarcity of labeled data, this limitation makes \ac{3DPC} tasks prone to overfitting, and poor generalization \cite{arief2019addressing}.

Several mechanisms help in tackling the overfitting issue. For instance, \cite{huang2019body} have used \ac{DTL} on a Kd-network, they have frozen the weights of all network layers except the last fully connected layer, and only the fully connected layer is modified so that the gradients during backpropagation are not calculated, which can avoid the occurrence of overfitting and improve training
efficiency \cite{wang2019ridesharing}. Entropy minimization (EM), a regularization method, has also been suggested by \cite{vesal2021adapt} in addressing overfitting. EM helps improve generalization and, hence, robustness for preventing overfitting. Arief et al. proposed a solution to address the overfitting problem of PointCNN, a DNN-based model, by using the Atrous XCRF model, which is a combination of \ac{CRF} and \ac{RNN} \cite{arief2019addressing}. DNN-based models such as PointCNN are highly susceptible to overfitting when the available dataset is relatively small, this is because of the large number of parameters in such models which requires a large number of training data. To tackle this problem, they introduced controlled noise during the training of a DNN-classifier, the method works by retraining a validated model using unlabeled test data. The training process is guided using the hierarchical structure of the \ac{CRF} penalty procedure. With the proposed algorithm XCRF and addition of A-XCRF layer, they were able to improve the model's accuracy by utilizing the unlabelled data.

\subsection{Domain discrepancies}
Recently, \ac{UDA} has gained a good deal of attraction by the research community, especially for various complex \ac{3DPC} tasks like semantic segmentation \cite{kim2022deep}  and object recognition \cite{tian2021vdm}. However, domain discrepancy exists due to the \textit{domain shift} \cite{morerio2017minimal}; hence, it becomes difficult for a model to perform across domains accurately. Different solutions have been proposed to minimize the domain discrepancy for \ac{3DPC} tasks. To this end, \ac{DL} methods based on the adversarial network, such as domain adversarial neural networks [18], have improved performance for \ac{UDA} tasks. The primary job of bridging the domain gap is performed by introducing a generator that keeps fooling the discriminator until it identifies the discrepancies \cite{, zhu2021automatic}. Moving forward, for classification purposes, a model is presented in \cite{qin2019pointdan} that focuses on aligning the local and global features.
Similarly, for 3D segmentation on \ac{LiDAR} sensors, authors in \cite{wu2019squeezesegv2} have considered aligning activation correlation \cite{morerio2017minimal}. The out space alignment and entropy minimization has been integrated by the authors in \cite{vesal2021adapt} for addressing the above \ac{UDA} issues. Domain discrepancies that can arise due to \ac{LiDAR} sensors are handled by Yi et al. \cite{yi2021complete}. Accordingly, a \ac{SSL} framework to improve \ac{UDA} performance has been suggested in \cite{achituve2021self}.

\begin{figure}
    \centering
    \includegraphics[width=1\textwidth]{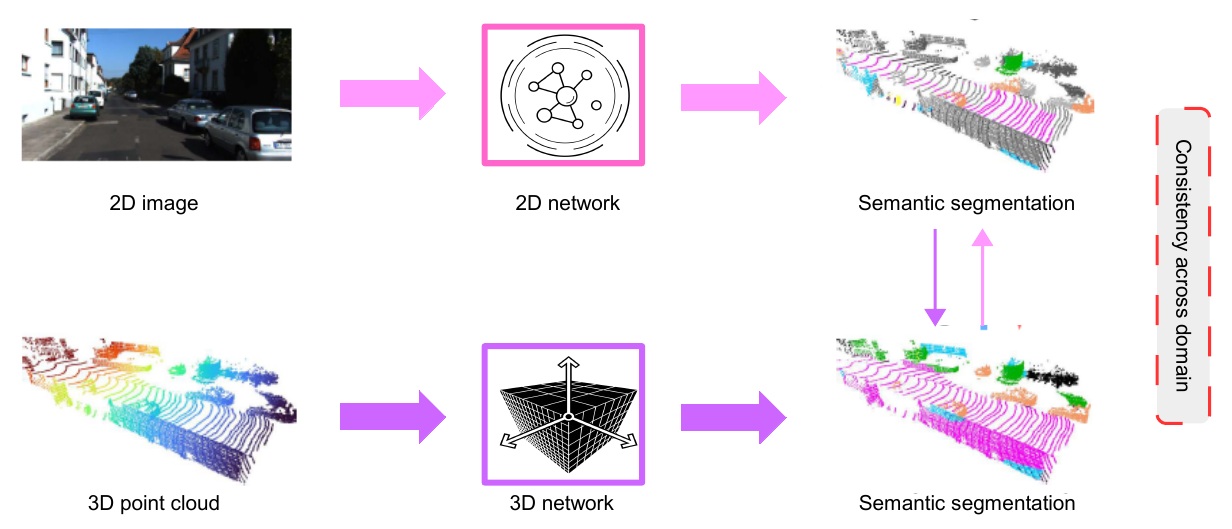} 
    \caption{The \ac{CLDA} proposed in \cite{jaritz2022cross}, where consistency has been ensured through mutual mimicking.}
    \label{fig:CDA}
\end{figure}

The proliferation in related research has paved the way for an increased interest in domain-invariant representations (DiR). The primary aim of the DiR is to facilitate a way that there should be insignificant discrepancies for features coming from different domains. To this end, Jaritz et al. \cite{jaritz2022cross} have leveraged the cross-modal discrepancies while preserving the best performance of each
sensor from self-driving data from cameras and \ac{LiDAR} \ac{3DPC}
sensors because domain gaps are different for different sensors. For example, comparing a \ac{LiDAR} and a camera, the former is more robust to lighting changes with respect to the latter. Whereas there are always dense images that come as output from a camera while \ac{LiDAR} sensing density is proportional to the sensor setup. Notably, 'cross-modality' differ from multi-modal fusion \cite{jaritz2022cross} as multi-modal implies training a single model in a supervised way for combining inputs like \ac{LiDAR} and camera \cite{liang2019multi}, RGB-D \cite{valada2020self}, etc.  This cross-modality DiR can help avoid the limitations of \ac{UDA} across modalities in which one modality influences the performance of the other modality in a negative way. In Fig. \ref{fig:CDA}, an overview of \ac{CLDA} is presented. Here, a prediction of 3D segmentation labels is performed through 2D and a 3D network by providing an image and a \ac{3DPC} as input to this network. 3D predictions are converted to 3D while consistency is ensured through mutual mimicking. This approach is proved to be advantageous for \ac{UDA}.

\subsection{Reproducibility of scientific results}
Despite the growing interest in using \ac{DTL} for various energy applications, there are still several factors that hinder the widespread adoption of \ac{DTL}-based models and impact reproducibility, and the ability to compare \ac{DTL}-based solutions. Firstly, it is difficult to evaluate the generality of \ac{DTL} models as most of the frameworks are evaluated on datasets collected under similar conditions, such as the same climate \cite{himeur2022deep}. Secondly, there is a lack of consistency in using the same datasets and benchmarks to validate new \ac{DTL} models, due to the limited availability of open-source, benchmarked datasets \cite{mateo2020transferring}. Finally, different metrics and parameter settings have been used to measure the distance between the source and \ac{TD}s and evaluate the performance of \ac{DTL}-based solutions on different datasets. This makes it challenging, and even impossible, to compare \ac{DTL} techniques uniformly \cite{ling2022graph}.  Despite that, there are some studies already uploaded on GitHub and other online platforms to facilitate the reproducibility and comparison tasks; the effort put in this direction still needs to consider the challenges introduced by \ac{DTL} for \ac{3DPC}s.

\subsection{Measuring knowledge gains}
Assessing the knowledge gained when using \ac{DTL} models for specific tasks is crucial, yet this challenge has not been fully addressed in the literature. Bengio et al. in \cite{glorot2011domain} proposed four measures, transfer error, transfer loss, transfer ratio, and in-domain ratio, to quantify the gain of knowledge in \ac{DTL}. However, it is unclear how these measures would perform with other \ac{DTL}-based methods, particularly in the energy sector, where the class sets are different between problems. Moreover, these measures can lead to non-definite performance evaluations if a perfect baseline model is achieved. To address these issues, simpler measures such as accuracy, F1 score, MSE, RMSE, MAE, performance improvement ratio (PIR) or other statistically-inspired coefficients have been widely used to evaluate \ac{DTL}-based solutions in the energy sector. These measures provide additional information, such as class agreement, and are better suited to the specific requirements of the energy sector.

\subsection{Unification of DTL}
The evolution of \ac{DTL} in energy applications has been rapid and innovative but also rather fragmented. This is due to the range of unique mathematical formulations utilized to delineate \ac{DTL} algorithms. Numerous researchers have developed and proposed various versions of \ac{DTL}, each bearing its unique label and implementation approach. Examples of these include "Heterogeneous \ac{DTL}" as proposed in \cite{hu2019heterogeneous}, "Statistical investigations" in \cite{fan2020statistical}, and "DA-DTL" in \cite{lin2021deep}. This multiplicity of terms and methods can often lead to confusion among scholars in the field.
It is clear that there is a pressing need to homogenize the terminologies and conceptual underpinnings of \ac{DTL} to foster better comprehension and collaboration within the research community. A seminal step towards this unification was taken in \cite{patricia2014learning}, laying a foundation upon which future studies can build. However, the exploration remains incomplete with respect to \ac{3DPC} applications in particular.

The unification of \ac{DTL} concepts entails collating, comparing, and reconciling the differing methodologies, with an ultimate aim to develop a more cohesive and comprehensive framework. A unified approach would facilitate the application of \ac{DTL} in \ac{3DPC} applications by standardizing algorithm descriptions, mitigating confusion, and making it easier for researchers to implement, adapt, and build upon existing algorithms. This, in turn, would expedite the progression of \ac{DTL}-based solutions, thereby accelerating advancements in the realm of energy applications.
Therefore, it is paramount to expand the unification efforts initiated in \cite{patricia2014learning} to include \ac{3DPC} applications. Future research endeavors should prioritize developing a harmonized approach that accommodates the diverse range of \ac{DTL} methodologies while also meeting the specific needs of \ac{3DPC} applications. This would not only streamline the use of \ac{DTL} in the field but also pave the way for more innovative developments in energy applications.

\section{Future Research directions} \label{sec6}

\subsection{3DPC Transformers}
\begin{figure}[t!]
\begin{center}
\includegraphics[width=1\textwidth]{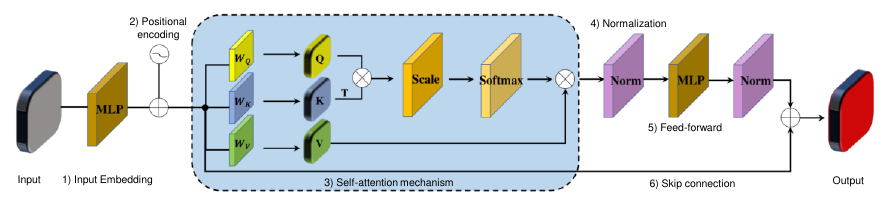}\\
\end{center}
\caption{Example of a Transformer's encoder architecture.}
\label{trans_3dpc}
\end{figure}

Transformer models have greatly improved NLP \cite{djeffal2023automatic} and CV, but their application in \ac{3DPC} processing remains uncertain. Questions remain about their ability to handle irregular and unordered data in 3D, their suitability for different 3D representations and their competence in various 3D processing tasks. Fig. \ref{trans_3dpc} illustrates an example of a Transformer encoder architecture. 


\begin{figure}[t!]
\begin{center}
\includegraphics[width=1\textwidth]{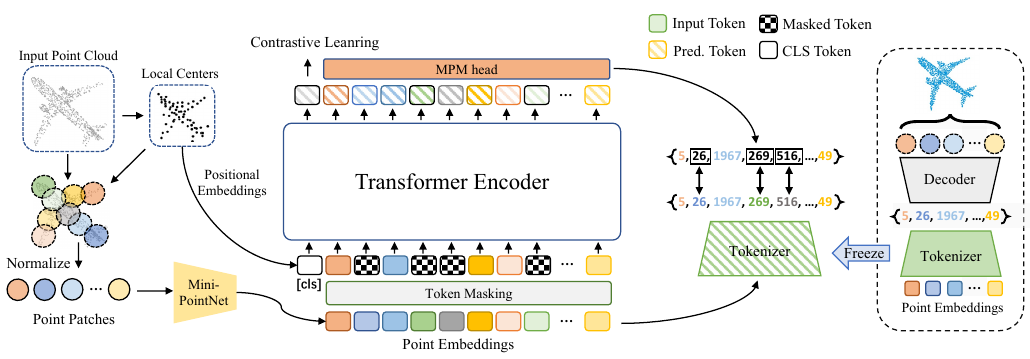}\\
\end{center}
\caption{An illustration of CNN based DTL for \ac{3DPC} segmentation and its types}
\label{Point_BERT}
\end{figure}

One important challenge when using transformers is that they require large amounts of training data. In the field of \ac{3DPC}, obtaining large datasets is a challenge and this makes training Transformers for 3D tasks difficult. \textcolor{black}{The authors in \cite{qian2022pix4point} have investigated the use of knowledge from multiple images for \ac{3DPC} understanding. They proposed a pipeline called \textit{Pix4Point} that allows for utilizing pretrained Transformers in the image domain to improve downstream \ac{3DPC} tasks.} This is achieved by using a modality-agnostic pure Transformer backbone with tokenizer and decoder layers specialized in the 3D domain. Another model named \enquote{Point-BERT} which is based on BERT's architecture \cite{devlin2018bert} to learn Transformers for generalizing the idea of BERT to \ac{3DPC} is proposed in \cite{yu2022point}. Point-BERT is trained using a masked point modeling task. The pipeline of Point-BERT is illustrated in Fig. \ref{Point_BERT}, where the input \ac{3DPC} is first partitioned into several point patches.


\subsection{Further generalization}

Enhancing the generalization capabilities of \ac{DTL} algorithms for \ac{3DPC} understanding involves some unique challenges and strategies due to the distinctive nature of 3D data. For instance, data augmentation techniques for 3D data can be more complex than for images or text. Techniques could include rotations, translations, adding Gaussian noise, or mirroring the point cloud along a plane. This would create a more diverse training dataset and can help the model generalize better. In order to reduce the domain shift between the source and target task, techniques such as feature alignment can be used. This involves minimizing the difference between the distributions of the source and target features. Moreover, 
utilizing architectures that maintain geometric invariance can be helpful as \ac{3DPC} data can be rotated or scaled. Networks, such as PointNet or PointNet++ which are invariant to permutations and transformations of the input points can be utilized. Additionally, methods such as PointNet++ use hierarchical neural networks to capture local structures induced by the metric space points live in, and also model global structures. This can help with generalization as it learns at different scales.
Moving on, regularization methods such as dropout or weight decay can be implemented to prevent overfitting and ensure the model generalizes better.
Lastly, designing loss functions that better capture the properties of 3D data can also help with generalization.

\subsection{Multi-task learning}

\subsubsection{Segmentation and Detection}
This section focuses on frameworks that handle multiple tasks like segmentation, detection, and sometimes additional tasks like classification or instance identification within point clouds. Typically, Pham et al. 2019 \cite{pham2019jsis3d} and Zou et al. 2022 \cite{zou2022multi} both explore multi-task learning frameworks that handle segmentation and classification simultaneously. Pham et al. focus on combining semantic and instance segmentation, while Zou et al. integrate classification and segmentation tasks to enhance each other, utilizing a Y-shaped graph neural network. Moving on, Chen et al. 2022 \cite{chen2022jspnet} propose a method for joint semantic and instance segmentation using a shared encoder and dual decoders, focusing on enhancing feature representation through cross-task interactions.
similarly, Ye et al. 2023 \cite{ye2023lidarmultinet} extend the multi-task approach to LiDAR-based perception tasks, unifying object detection, semantic segmentation, and panoptic segmentation in a single network to optimize performance across tasks. These studies collectively push the boundaries in efficient multi-task handling, demonstrating that integrating tasks can improve overall performance and reduce computational costs by sharing learned features across tasks.

\subsubsection{Advanced Architectures for Enhanced Feature Extraction}
MTL can be used for developing sophisticated neural network architectures to better handle the complexities of point cloud data. In this regard, Dubey et al. 2022 \cite{dubey2022haradnet} introduce an attention-based architecture for radar point cloud data, focusing on enhancing target localization and classification through an anchor-free network design. Similarly, Hassani et al. 2019 \cite{hassani2019unsupervised} leverage an unsupervised multi-task model to learn point and shape features effectively, employing a graph-based encoder that tackles tasks like clustering and reconstruction simultaneously.
Moving forward, Lin et al. 2022 \cite{lin2022multi} use a multi-task learning framework to improve semantic segmentation efficiency in Mobile Laser Scanning point clouds by integrating color prediction as an auxiliary task.

\subsubsection{Specific Enhancements}
Many studies have focused on applying multi-task learning to achieve specific enhancements in processing or understanding point cloud data. In this direction, Zhao et al. 2024 \cite{zhao2024robust} develop a robust network for preprocessing LiDAR data, tackling denoising, segmentation, and completion tasks within a shared framework to improve data quality. Similarly, Rios et al. 2021 \cite{rios2021multitask} employ a multi-task approach in an evolutionary optimization context, using a 3DPC autoencoder to unify different design representations in a common latent space.
Feng et al. 2021 \cite{feng2021simple} propose a multi-task network to handle various perception tasks for autonomous driving, demonstrating how a unified approach can enhance both detection and road understanding tasks.

Besides, the study in \cite{rebut2022raw} introduces FFT-RadNet, a novel HD radar sensing model that optimizes the computation of angular positions from radar data, facilitating vehicle detection and free driving space segmentation. Moving on, Shan et al. 2023 \cite{shan2023gpa} propose a no-reference point cloud quality assessment metric, GPA-Net, which utilizes a multi-task framework to enhance the accuracy of quality regression by also predicting distortion type and degree. Besides, the authors in  \cite{hatem2023point} present Point-TTA, a test-time adaptation framework for point cloud registration that improves model generalization and performance on unknown testing environments through self-supervised auxiliary tasks. On the other hand, Zhang et al. 2022 \cite{zhang2022improved} - The study proposes an improved multi-task network for roof plane segmentation from airborne laser scanning point clouds, which effectively segments and identifies roof planes.
Wei et al. 2021 \cite{wei2021multi} - This paper proposes a multi-task network for 3D keypoint detection, which efficiently captures local and global features for accurate keypoint localization and semantic labeling.
Lastly, the work in \cite{zhang2022machining} introduces a multi-task network for machining feature recognition from point cloud data, aiming to accurately segment and identify complex machining features.

\subsection{Cross-Modal Transfer Learning}

\subsubsection{Enhancing 3DPC Understanding Through 2D-3D Correspondences}
Many studies employ contrastive learning and knowledge distillation strategies to align and enhance features between 2D images and 3DPCs. These methods focus on leveraging the rich textual and visual information available in 2D data to augment the spatial understanding provided by 3DPCs, significantly boosting performance in tasks such as classification, segmentation, and dense captioning. For instance, Afham et al. \cite{afham2022crosspoint} introduce CrossPoint, utilizing self-supervised contrastive learning to map 2D image features to 3DPCs for better object classification and segmentation.
The PointCMT, proposed in Yan et al. \cite{yan2022let}, adopts a teacher-student framework for cross-modal training, using 2D images to enhance 3DPC classification through knowledge distillation.
Wu et al. \cite{wu2023self} propose CrossNet, a method that aligns 3DPC features with both colored and grayscale images to enhance classification and segmentation tasks across different domains.
Group 2: Cross-Modal and Cross-Domain Adaptation for Comprehensive 3D Understanding

\subsubsection{Comprehensive 3D Understanding}
CMTL can be employed in various works to tackle the challenge of transferring knowledge across not just modalities but also different domains, aiming to improve the performance of 3D perception tasks under varying conditions and datasets, without relying heavily on labeled data. Typically, Zhang et al. \cite{zhang2023cross} develop a novel adaptation strategy that aligns features between images and point clouds to enhance semantic segmentation without requiring 3D labels. The X4D-SceneFormer, introduced by Jing et al. \cite{jing2024x4d}, uses a dual-branch Transformer architecture to transfer dynamic textual and visual cues from 2D video sequences to 4D point cloud sequences, focusing on temporal and semantic coherence. Zhou et al. \cite{zhang2023pointmcd} introduce PointCMC, which utilizes a Local-to-Local (L2L) module and a Cross-Modal Local-Global Contrastive (CLGC) loss to enhance cross-modal knowledge transfer between image and point cloud data.

\subsubsection{Innovative Cross-Modal Applications in Robotics and Dynamic Environments}
This section discusses the group of studies that applies cross-modal transfer learning to specific, challenging scenarios such as robotic tactile recognition and dynamic 4D scene understanding, demonstrating the flexibility and potential of cross-modal learning in practical and dynamic applications.
Specifically, Falco et al. \cite{falco2019transfer} explore visuo-tactile object recognition, where a robot uses visual data to enhance its tactile recognition capabilities, effectively bridging the gap between seeing and touching.
Additionally, the authors in \cite{jing2024x4d} focus on enhancing 4D point cloud understanding by transferring knowledge from RGB video sequences, improving the dynamic scene comprehension significantly.
Murali et al. \cite{murali2022deep} propose xAVTNet, a visuo-tactile cross-modal framework for object recognition by autonomous robotic systems, utilizing active learning and perception strategies.

\subsubsection{Enhanced Sensory Perception and Retrieval}
CMTL can be used to improve sensory perception and retrieval tasks, enabling more effective and efficient recognition and classification across different sensory inputs.
For example, Shen et al. \cite{shen2022simcrosstrans} implement a simple cross-modal transfer from 2D to 3D sensors, using Vision Transformers to handle occlusions and improve performance in low-light scenarios.
Jing et al. \cite{jing2021cross} enhances cross-modal retrieval capabilities by training a network to minimize feature discrepancies across 2D images, 3DPCs, and mesh data, ensuring robust retrieval performance across modalities.

\subsubsection{Point Cloud Completion and Enhancement}
Zhu et al. \cite{zhu2023csdn} introduce a cross-modal shape-transfer dual-refinement network (CSDN) designed for point cloud completion, utilizing images to guide the geometry generation of missing regions and refine the output through dual refinement processes.
Li et al. \cite{li2022cross} propose a model integrating Cross-Domain and Cross-Modal Knowledge Distillation to mitigate domain shifts and enhance the interaction between LiDAR and camera data, improving adaptation in unsupervised settings.
Zhang et al. \cite{zhang2023pointmcd} explore PointMCD, a multi-view cross-modal distillation architecture, which aligns features between 2D visual and 3D geometric domains to boost the learning capacity of point cloud encoders.

\subsubsection{Domain Adaptation and Generalization in Cross-Modal Settings}
Peng et al. \cite{peng2021sparse} discuss leveraging 2D images for 3D domain adaptation through cross-modal learning, addressing the loss of useful 2D features and promoting high-level modal complementarity.
Nitsch et al. \cite{nitsch2020learning} propose a transductive transfer learning approach that uses a multi-modal adversarial autoencoder to transfer knowledge from images to point clouds, focusing on object detection.
Li et al. \cite{li2023bev} introduce a bird's-eye view approach for cross-modal learning under Domain Generalization (DG) for 3D semantic segmentation, optimizing domain-irrelevant representation modeling.

\subsubsection{Semantic Segmentation}
The ProtoTransfer, proposed in \cite{tang2023prototransfer}, explores knowledge transfer from multi-modal sources, such as LiDAR points and images, to enhance point cloud semantic segmentation, using a class-wise prototype bank to fully exploit image representations and transfer multi-modal knowledge to point cloud features. This approach effectively handles the challenge of unmatched point and pixel features by employing a pseudo-labeling scheme to integrate these features into the prototype bank, demonstrating superior performance on large-scale benchmarks.
Similarly, Jaritz et al. \cite{jaritz2020xmuda} explore xMUDA, a cross-modal UDA technique for 3D semantic segmentation, leveraging mutual mimicking between 2D images and 3DPCs.
Xing et al. \cite{xing2023cross} discuss enhancing domain adaptation for 3D semantic segmentation using cross-modal contrastive learning to improve interactions between 2D-pixel features and 3D point features.


Table \ref{tab:comparison2} presents a comprehensive comparison of various studies focused on cross-modal learning and point cloud analysis, highlighting significant contributions and challenges in the field. For example, CrossPoint utilizes a self-supervised approach to establish 3D-2D correspondence, improving 3D object classification and segmentation, although it faces complexities in cross-modal alignment. Similarly, PointCMT enhances the representation of point clouds by incorporating view images derived from 2D image quality, thereby boosting 3DPC classification capabilities. Other notable methodologies include CrossNet, which excels in both intra- and cross-modal learning across multiple benchmark datasets, and X-Trans2Cap, which demonstrates effective cross-modal knowledge transfer in 3D dense captioning, albeit at a high computational cost.

The studies also outline several limitations inherent in current cross-modal and point cloud analysis techniques. For instance, many models, like the Cross-modal adaptation and simCrossTrans, struggle with domain shifts or performance variations across different 3D tasks, respectively. Additionally, while technologies such as the Visuotactile object recognition and xAVTNet offer high accuracy in specific applications such as object recognition in robotics, they are limited by their dependency on particular datasets and high system complexities. Other common challenges include the sensitivity to hyperparameters, as seen in Dual-Cross for cross-modal UDA, and the difficulty in feature alignment and generalization across different domains, which affects models like PointMCD and BEV-DG.

\begin{longtable}{m{0.7cm}m{2cm}m{2cm}m{2.5cm}m{3.1cm}m{3.2cm}}
\caption{Comparison of Studies on Cross-Modal Learning and Point Cloud Analysis}\
\label{tab:comparison2} \\
\hline
\textbf{Ref.} & \textbf{ML Model} & \textbf{Dataset} & \textbf{Application / Task} & \textbf{Advantage} & \textbf{Limitation} \\ \hline
\endfirsthead
\multicolumn{6}{c}{{\bfseries Table \thetable\ continued from previous page}} \\
\hline
\textbf{Ref.} & \textbf{ML Model} & \textbf{Dataset} & \textbf{Application / Task} & \textbf{Advantage} & \textbf{Limitation} \\ \hline
\endhead

\hline \multicolumn{6}{r}{{Continued on next page}} \\ \hline
\endfoot

\hline \hline
\endlastfoot
\cite{afham2022crosspoint} & CrossPoint & Varied & 3D object classification, segmentation & Uses 3D-2D correspondence; self-supervised & May require complex cross-modal alignment \\
\cite{yan2022let} & PointCMT & ModelNet40, ScanObjectNN & 3DPC classification & Enhances point-only representation; uses view-images & Depends on 2D image quality \\
\cite{wu2023self} & CrossNet & Multiple benchmarks & 3DPC classification, segmentation & Intra- and cross-modal learning; versatile & Complex model structure for fine-tuning \\
\cite{yuan2022x} & X-Trans2Cap & ScanRefer, Nr3D & 3D dense captioning & Effective cross-modal knowledge transfer; high accuracy & High computational cost in training \\
\cite{zhang2023cross} & Cross-modal adaptation & SemanticKITTI, KITTI360, GTA5 & 3DPC semantic segmentation & Leverages 2D datasets; unsupervised & May struggle with substantial domain shifts \\
\cite{falco2019transfer} & Visuo-tactile object recognition & 15 objects dataset & Object recognition & High accuracy; cross-modal transfer & Limited to specific tactile and visual datasets \\
\cite{jing2024x4d} & X4D-SceneFormer & HOI4D & 4D point cloud tasks & Incorporates temporal dynamics; high performance & Complex model, heavy on resources \\
\cite{shen2022simcrosstrans} & simCrossTrans & SUN RGB-D & 3D sensor performance & Utilizes ViTs; robust to occlusions & Performance may vary across different 3D tasks \\
\cite{jing2021cross} & Cross-modal retrieval & ModelNet10, ModelNet40 & Cross-modal retrieval & Efficient feature learning; high retrieval accuracy & Depends on network's feature extraction ability \\
\cite{tang2023prototransfer} & ProtoTransfer & nuScenes, SemanticKITTI & Point cloud semantic segmentation & Exploits multi-modal fusion; high accuracy & May miss benefits for unmatched features \\
\cite{zhu2023csdn} & CSDN & Varied & Point cloud completion & Coarse-to-fine approach with dual-refinement; cross-modal data use & Complex model structure \\
\cite{murali2022deep} & xAVTNet & Robotics system & Visuo-tactile object recognition & Integrates visuo-tactile data; uses active learning & Specific to robotics; high system complexity \\
\cite{li2022cross} & Dual-Cross & Multi-modal & Cross-modal UDA & Integrates CDKD and CMKD for domain adaptation & Highly sensitive to hyperparameters \\
\cite{peng2021sparse} & DsCML, CMAL & Multi-modal & 3D semantic segmentation & Enhances cross-modal learning; diverse domain adaptation & Potential feature mismatch \\
\cite{nitsch2020learning} & Adversarial Auto Encoder & KITTI & Object detection & Transductive transfer from images to point clouds & Requires multi-modal data \\
\cite{zhou2023pointcmc} & PointCMC & Varied & 3D object classification, segmentation & Enhances fine-grained cross-modal knowledge transfer & Alignment challenges \\
\cite{jaritz2020xmuda} & xMUDA & Autonomous driving datasets & 3D semantic segmentation & Utilizes mutual mimicking in multi-modality & Complexity in heterogeneous input spaces \\
\cite{xing2023cross} & Cross-modal contrastive learning & Varied & 3D semantic segmentation & Improves adaptation effects; uses neighborhood feature aggregation & Depends on precise feature correspondence \\
\cite{zhang2023pointmcd} & PointMCD & 3D datasets & 3D shape classification, segmentation & Uses cross-modal distillation; aligns features across modalities & Complicated feature alignment process \\
\cite{li2023bev} & BEV-DG & 3D datasets & 3D semantic segmentation & Implements BEV-based cross-modal learning; robust to misalignment & Domain generalization complexity \\

\end{longtable}

\subsection{Diffusion models for 3DPC undertanding}
Diffusion models are increasingly used for 3DPC applications due to their ability to generate high-quality, detailed structures, which is crucial for tasks such as 3D modeling and virtual reality. These models excel in handling complex distributions characteristic of 3DPCs, which are typically unordered and involve intricate spatial relationships. Their gradual denoising process allows them to implicitly learn the nuances of geometric structures, making them suitable for a range of tasks including reconstruction, up-sampling, and completion. Additionally, diffusion models are robust to noisy data, a common challenge with real-world 3D sensor data, and can be integrated with specialized 3D neural network architectures like PointNet, enhancing their effectiveness in processing 3DPCs. This flexibility and robustness make diffusion models a promising approach for advanced 3D applications across various industries.

The studies on diffusion model-based 3DPCs offer innovative solutions across various domains. Zheng et al. \cite{zheng2024point} introduce PointDif, a pre-training method for 3D models that employs a conditional point generator to enhance feature aggregation and guide point-to-point recovery, showing notable improvements in classification and segmentation tasks. Kasten et al. \cite{kasten2024point} describe SDS-Complete, a text-to-image diffusion model for completing 3D objects from partial point clouds, demonstrating its effectiveness in handling Out-Of-Distribution objects. Liu's study \cite{liu20243d} focuses on a Diffusion Probabilistic Network for point cloud segmentation, utilizing a reverse diffusion process to refine topological structures. Jiang et al. \cite{jiang2024se} develop an SE(3) diffusion model-based framework for 3D registration, achieving precise object pose alignment by manipulating noise in the SE(3) manifold. Jin introduces \cite{jin2024multiway} a multiway point cloud mosaicking approach, employing a diffusion-based denoising process for enhanced registration accuracy. Feng's \cite{feng2024diffpoint} DiffPoint combines Vision Transformers with diffusion models for 2D-to-3D reconstruction, achieving superior results in both single and multi-view tasks. Sharma \cite{sharma2024generating} proposes a class-conditioned diffusion model to generate synthetic point cloud embeddings, significantly enhancing classification performance. Mo's \cite{mo2024dit} DiT-3D uses a Diffusion Transformer to generate high-quality 3D shapes from voxelized point clouds, integrating 3D window attention for computational efficiency. Yi's \cite{yi2024gaussiandreamer} GaussianDreamer leverages both 2D and 3D diffusion models for rapid, high-quality 3D generation from text prompts. Lastly, Ho \cite{ho2024diffusion} introduces Diffusion-SS3D, integrating diffusion models into a semi-supervised learning framework for 3D object detection, improving the quality of pseudo-labels and enhancing detection accuracy in diverse 3D spaces. These advancements collectively highlight the versatility and effectiveness of diffusion models in enhancing 3DPC processing and analysis.

Ohno et al. (2024) present a privacy-centric pedestrian tracking system using 3D LiDARs, capturing pedestrians as anonymous point clouds and leveraging a generative diffusion model to predict trajectories in unmonitored areas, achieving a high F-measure of 0.98 \cite{ohno2024privacy}. In a different approach, Bi et al. (2024) introduce DiffusionEMIS, a novel paradigm that models 3-D electromagnetic inverse scattering as a denoising diffusion process, significantly improving performance and noise resistance over conventional methods \cite{bi2024diffusionemis}. Dutt (2024) develops Diff3F, which distills diffusion features onto untextured shapes, demonstrating robustness and reliability across several benchmarks \cite{dutt2024diffusion}. Simultaneously, Li (2024) proposes a framework for generating stable crystal structures using a point cloud-based diffusion model, enhancing material design \cite{li2024generative}. Expanding the applications, Ze (2024) integrates 3D visual representations into diffusion models for robotic imitation learning with the 3D Diffusion Policy (DP3), which shows remarkable success rates and generalizability \cite{ze20243d}. Addressing point cloud registration, She (2024) utilizes graph neural PDEs and heat kernel signatures to enhance keypoint matching accuracy and robustness \cite{she2024pointdifformer}. Meanwhile, Li (2024) details a method for reconstructing 3D colored objects from images using a conditional diffusion model, achieving high fidelity in shape and color reconstruction \cite{li20243d}. Chen (2024) introduces V3D, adapting video diffusion models for 3D generation that produce high-quality outputs rapidly with impressive multi-view consistency \cite{chen2024v3d}. Further, Hu (2024) presents a novel generative model combining latent diffusion with topological features, enabling diverse and adaptable 3D shape generation \cite{hu2024topology}. Finally, Dong (2024) proposes a novel generative model for man-made shapes, incorporating diffusion processes with a quality checker to ensure geometric feasibility and physical stability, significantly outperforming traditional methods on ShapeNet-v2 \cite{dong2024gpld3d}.

Several future directions can be suggested to further improve the use of diffusion models in 3DPC. These include enhancing privacy while utilizing detailed data, improving computational efficiency in model adaptation with minimal trainable parameters, and increasing robustness against data quality variations. Additionally, there's a trend toward cross-domain applications and interdisciplinary approaches, such as integrating 2D image processing techniques. The generation of complex and detailed 3D structures, especially in dynamic environments, is also a focal point, along with the development of semi-supervised and fully automated learning systems. Innovations in incorporating topological and geometric features into diffusion processes highlight the potential for more precise and versatile models. Finally, there's a growing need for standardization in benchmarking methods to evaluate and compare the performance of diffusion models in handling 3DPCs. These insights indicate a move towards more efficient, robust, and application-diverse uses of diffusion models in the realm of 3D data analysis.

\section{Conclusion}  \label{sec7}
With the rapid advancement of 3D scanning technologies, such as \ac{LiDAR} and RGB-D cameras, capturing and processing \ac{3DPC} data has become increasingly popular in various fields, including robotics, \ac{CV}, virtual reality, and autonomous vehicles. On the other hand, deep \ac{DTL}, a cutting-edge technique in \ac{ML} and \ac{DL}, has emerged as a powerful approach to leverage pre-trained DNNs and transfer their knowledge to new \ac{3DPC} tasks overcome several challenges to using \ac{DL}.
To shed light on the latest innovations related to using \ac{DTL} in \acp{3DPC}, this article presented a comprehensive review of the
state-of-the-art techniques for \ac{3DPC} understanding using \ac{DTL} and DA, including \ac{3DPC} object detection, \ac{3DPC}  semantic labeling, segmentation and classification, \ac{3DPC} registration,  downsampling/upsampling and \ac{3DPC} denoising. In doing so, a well-defined taxonomy has been introduced, and detailed comparisons have been conducted, presented with reference to different aspects, such as the types of adopted \ac{DL} models, obtained performance, and knowledge transfer strategies (fine-tuning, DA, \ac{UDA}, etc.). Moving forward, the pros and cons of presented frameworks have been covered before identifying open challenges. Lastly, potential research directions have been listed.

Point cloud understanding has made significant progress through years of research, but as it becomes more widely used in real-world applications, it faces new challenges. One major challenge is partial overlap in sensor-acquired point clouds, which makes direct registration difficult. While some solutions have been developed for alignment under partial overlap, the rate of overlap is often limited. Finding a comprehensive solution to this problem, therefore, is a valuable and promising area of research.
Moreover, only simple objects can currently be solved for alignment, despite the fact that the \ac{3DPC} registration techniques that combine \ac{DTL} and conventional methods have made significant advancements. For instance, when dealing with complex scenarios and large-scale \acp{3DPC}, these approaches fall short of the desired outcomes and continue to use the conventional algorithms.
However, these algorithms are stochastic, and the number of iterations rise exponentially with outliers. By combining \ac{DTL} and conventional \ac{ML} methods, better results are obtained. Specifically, traditional \ac{ML} approaches are transparent, whereas \ac{DTL} schemes excel at fitting data. Hence, one of the trends for the future research is how to combine the benefits of both.

Besides, the need for generalizing \ac{3DPC} scene understanding algorithms arises from the fact that different application scenarios present the algorithms with different challenges. Nevertheless, given the state of the research, it is difficult to suggest a general algorithm. For instance, an aircraft's skin is very large, its surface is smooth, and it has few features with a curvature. When feature-based methods are used in the registration process, significant misalignment will consequently happen. The development of targeted, lightweight, and efficient algorithms for particular application scenarios is thus an attractive research hotspot in the near future.

\end{document}